%% file: main.tex
\documentclass{article}
\usepackage[final, nonatbib]{neurips_2024}

\usepackage[utf8]{inputenc} %
\usepackage[T1]{fontenc}    %
\usepackage[breaklinks,colorlinks]{hyperref}
\usepackage{url}            %
\usepackage{booktabs}       %
\usepackage{amsfonts}       %
\usepackage{amsmath}
\usepackage{nicefrac}       %
\usepackage{microtype}      %
\usepackage{subfiles}
\usepackage{xcolor}
\usepackage{multirow}
\usepackage{enumerate}
\usepackage{graphicx}
\usepackage{amssymb}        %
\usepackage{breqn}
\usepackage{listings}
\usepackage{bm}
\usepackage{float}
\usepackage{subcaption}
\usepackage{caption}
\usepackage{parskip}
\usepackage[export]{adjustbox}

\definecolor{codegreen}{rgb}{0,0.6,0}
\definecolor{codegray}{rgb}{0.5,0.5,0.5}
\definecolor{codepurple}{rgb}{0.58,0,0.82}
\definecolor{backcolour}{rgb}{0.95,0.95,0.92}

\lstdefinestyle{mystyle}{
    commentstyle=\color{codegreen},
    keywordstyle=\color{magenta},
    numberstyle=\tiny\color{codegray},
    stringstyle=\color{codepurple},
    basicstyle=\ttfamily\footnotesize,
    breakatwhitespace=false,
    breaklines=true,
    captionpos=b,
    keepspaces=true,
    numbers=left,
    numbersep=5pt,
    showspaces=false,
    showstringspaces=false,
    showtabs=false,
    tabsize=2
}

\usepackage[
    style=alphabetic,
    sorting=ynt,
    backref=true
]{biblatex}
\usepackage{threeparttable}
\usepackage{multicol}

\title{Self-Supervised Learning for Time Series\\
\vspace{.2cm}
\large A Review \& Critique of FITS}
\date{June 24, 2024}

\author{
  Andreas Løvendahl Eefsen\thanks{Listing is in alphabetical order. We are especially thankful to our supervisor, Tommy Sonne Alstrøm, for sparking the inspiration to do time series, his thought-provoking questions that sparked new research directions, and his invaluable guidance and feedback throughout our project.
  \\\\
  This document originated as a more comprehensive report, akin to a group bachelor thesis for a university project, which included sections on Stochastic Gradient Descent (SGD), choice of optimizer, Autoregressive Integrated Moving Average (ARIMA), and other topics. To adhere to the conventional format of papers posted on arXiv, the main content has been streamlined to focus primarily on novel or relevant information, with the additional sections cut or moved to the appendix.}\\
  DTU\\
  \texttt{s224223@dtu.dk}\\
  \And
  Nicholas Erup Larsen\\
  DTU\\
  \texttt{s224175@dtu.dk}\\
  \AND
  Oliver Glozmann Bork Hansen\\
  DTU\\
  \texttt{s194591@dtu.dk}\\  
  \And
  Thor Højhus Avenstrup\\
  DTU\\
  \texttt{s224233@dtu.dk}\\
}

\begin{document}

\maketitle

\begin{center}
June 24, 2024
\end{center}
\vspace{.5cm}

\begin{abstract}
Accurate time series forecasting is a highly valuable endeavour with applications across many industries. Despite recent deep learning advancements, increased model complexity, and larger model sizes, many state-of-the-art models often perform worse or on par with simpler models. One of those cases is a recently proposed model, FITS,  claiming competitive performance with significantly reduced parameter counts. By training a one-layer neural network in the complex frequency domain, we are able to replicate these results. Our experiments on a wide range of real-world datasets further reveal that FITS especially excels at capturing periodic and seasonal patterns, but struggles with trending, non-periodic, or random-resembling behavior. With our two novel hybrid approaches, where we attempt to remedy the weaknesses of FITS by combining it with DLinear, we achieve the best results of any known open-source model on multivariate regression and promising results in multiple/linear regression on price datasets, on top of vastly improving upon what FITS achieves as a standalone model.

Our code developed and used for the project is available in the following GitHub Repository: \href{https://github.com/thorhojhus/ssl\_fts/}{github.com/thorhojhus/ssl\_fts/}
\end{abstract}

\newpage
\section{Introduction}
\input{introduction}

\newpage
\section{Research Questions}
\input{researchsqs}

\newpage
\section{Methods}
\input{methods}

\newpage 
\section{Data}
\input{data}

\newpage
\section{Results} \label{sec:results}
\input{results}

\newpage
\section{Discussion}
\input{discussion}

\newpage
\section{Conclusion}
In this report, we have explored the performance of the model Frequency Interpolation for Time Series Analysis (FITS), testing its performance on common datasets as well as \$GD and \$MRO. The FITS model leverages the Fourier transform to project time-domain data into the frequency domain, capturing periodic patterns that can be highly informative in  accurate forecasting on certain types of datasets. Through various experiments and comparisons with other models, including the autoregressive integrated moving average (ARIMA) model and hybrid models like DLinear, we have explored the strengths and limitations of FITS.

We have found that FITS performs very well on datasets with clear periodicity, such as the ETT, Electricity and Weather datasets, where it often outperforms other larger models especially at longer forecast horizons. The frequency representation inherent in FITS allows it to easily decompose and reconstruct signals which results in it being able to easily capture patterns other models might miss. However, FITS struggles with datasets that exhibit random or non-periodic behavior such as stock prices, where simpler models or naive baselines perform better. Likewise if the base period of a signal is too long compared to the window used for the Fourier transform, FITS is unable to fully capture the underlying signal instead learning higher frequency representations.

We also experimented with enhancements to the FITS model, such as deep FITS with ModReLU or $\mathbb{C}$ReLU, deep FITS after upscaler, and linear layer bypass FITS. Some of these showed promise at shorter forecast horizons, but the improvements were often marginal and came at the cost of increased model complexity. Which indicates that the original FITS implementation is already close to optimal for its intended purpose and when operating only in the frequency domain. 

Additionally, we evaluated the use of hybrid models, combining the strengths of FITS and DLinear, which showed improved performance in some cases. These models effectively captured both periodic and non-periodic components of the data, making them robust for a wider range of time series forecasting tasks.

In conclusion, FITS stands out as an efficient model for time series which can be modelled appropriately using the frequency representation. It is particularly useful for longer forecasting horizons compared to previous models while being exceedingly simple and efficient.

\newpage

\printbibliography

\newpage
\section{Appendix}\label{appendix:A}
\input{appendix}

\end{document}

%% file: introduction.tex
In this project, we aim to explore time series forecasting with the use of state of the art machine learning models. Time series analysis and forecasting is an important tool in many fields of science and business, which leads to an interest in developing more effective and precise time series forecasting models.

In the financial sector, forecasting stock prices, exchange rates, and economic indicators can bring a competitive edge to investors by managing risks and optimizing their portfolio returns. Healthcare providers can employ forecasting to predict diseases and take preventive measures. Additionally, the integration of sensor data in healthcare enables continuous monitoring of patient health, allowing for real-time analysis and more accurate predictions of medical conditions. And in the energy sector, forecasting electricity demand and renewable energy production is essential for grid stability and resource planning. Accurate forecasting is critical because electrical grids require a delicate balance between supply and demand. If the grid receives too much electricity, it can lead to overloading and potential damage to transmission lines, transformers, and other equipment. On the other hand, if there is insufficient electricity supply to meet the demand, it can result in power outages, blackouts, and disruptions to businesses and households. This can have massive negative cascading economic and social effects, as modern society relies heavily on a stable and reliable power supply.

The primary assertion of the FITS paper is that with just 10k-50k parameters, their method reaches comparable performance with previous state-of-the-art models of much larger parameter size, namely TimesNet (300.6M) \cite{times}, Pyraformer (241.4M), FEDformer (20.68M), Autoformer (13.61M), and PatchTST (1M) \cite{xu2024fits} on datasets from the aforementioned sectors. This is a substantial claim and we aim to reproduce their results and afterwards use the model on other datasets.

\subsection{Time Series Data \& Forecasting}%

A \emph{time series} is a continuous or discrete series or sequence of observations over a time frame. A lot of real world data would conform to the format of a \emph{time series}, examples are rainfall volumes over a given day or month, temperature over a year, sales of products over months, continuous sensor data, as well as changes in concentrations in chemical substances. There are examples of time series data abound in many fields including economics and business, chemistry and natural sciences, engineering as well as social sciences.

A time series, denoted $Z_T$, will be sampled such that the the discrete time series will have equidistant time points with a total of $T$ observations, for which we denote the observed value at time $t \leq T$ as $z_t$.

For the forecasting task, the forecasted time series at time step $t+l > T$ is denoted as $\hat{z}_{t+h}$. The \emph{forecast horizon}, also known as the \emph{leading time}, is here denoted $l$. The goal is to minimize the error from the forecasted values to that of the real time series for example the mean squared error:

\begin{equation}
    \min\frac{\sum_{t=k}^{k+l} (\hat{z}_{t+l} - z_{t+l})^2}{l}
\end{equation}

The difficulty in forecasting varies greatly and depends on the type of time series one wishes to predict. Data with clear harmonics, trends and little volatility will be easier and more simple to forecast than more volatile and chaotic data such as stock prices and noisy data.

A time series can also express properties of \textit{stationarity} and \textit{seasonality}. These two properties are critical in helping to understand how to build a model for the problem.

\textbf{Seasonality in Time Series}

There exists an abundant amount of time series data, where patterns repeat in intervals. Populations of certain animals, the average temperature during a month and even sales of merchandise can show signs around holidays and other times of the year.

What they all share, is a \textit{repeating} pattern. In literature, the usual letter denoting the length of the pattern, or seasonality length, is $m$; a simple example would be a sine curve, whose seasonality would be $m = 2\pi$\\\\

\textbf{Stationarity in Time Series}

A \textit{weakly} stationary time series exhibits no different behaviour when shifted in time. In simple probability terms, we require that the probability distribution over the time series should not depend on the time i.e., $f_{z|T=t}(z|T = t) = f_{z}(z)$.

In other words, if we write the time series as:

\begin{align*}
    z_t = \mu + \epsilon_t + \psi_1\epsilon_{t-1} + \psi_2\epsilon_{t-2} + \cdots
\end{align*}

Then a requirement of stationarity is that the weights, $\psi$, have to be \textit{absolutely summable} meaning $\sum|\psi| < \infty$, and as such, the parameter $\mu$ has a meaning denoting the level the time series varies about. The covariance between $z_t$ and $z_{t-j}$ should also be equal, and only depend on the relative time difference.\cite{arma_book}\cite{Hamilton1994-cw}

\subsection{Evolution of Forecasting Methods}

Forecasting has undergone a significant transformation over time, evolving from simple statistical approaches to more sophisticated ML techniques and lately transformer-based models. This section provides an overview of the key milestones in the evolution of forecasting methods.

\subsubsection{Statistical Forecasting}
The late 19th and early 20th centuries saw the emergence of statistical forecasting methods. These techniques, such as moving averages, exponential smoothing, and linear regression, used historical data to identify patterns and make predictions. They provided a more systematic and data-driven approach compared to intuition-based methods. Specially the ARIMA model is of interest, and will be described in more detail in the appendix \ref{sec: ARIMA}. \label{goback0}

\subsubsection{Machine Learning-based Models}
With the advent of machine learning, forecasting methods have undergone a significant transformation. Machine learning algorithms, such as artificial neural networks, long short-term memory (LSTM) networks, random forests, and support vector regression, have been applied to forecasting tasks. These methods can handle complex patterns, nonlinearities, and large datasets, often outperforming traditional statistical methods.

\subsubsection{Transformer-based Models}
In recent years, Transformer-based models have emerged as a popular architecture surrounding ANNs, garnering significant attention in machine learning communities due to their unique ability to capture long-range dependencies and complex patterns in sequential data. The main breakthrough of Transformer-based models is the multi-headed attention mechanism which updates the weights of the model based on all input steps. This allows it to retain more information about how each step in the sequence affects each other. One of the major pitfalls of the traditional self-attention mechanism, however, is the $O(L^2)$ (L = input length) time and memory complexity when training models.

One of the first papers to use Transformer architecture on time series data was Informer \cite{zhou2021informer}. Informer's main focus is improving the speed of inference, training, and efficiently handling long inputs. Among other things, it introduces a new "ProbSparse" self-attention mechanism, which instead achieves $O(L\log L)$ and works by focusing on a subset of the most informative parts of the input sequence by selecting only the top $u$ queries that have the highest sparsity measurements. Informer achieved an average of $\sim 1.6$ MSE and $\sim 0.79$ MAE across all datasets.

Half a year later, Autoformer comes out \cite{wu2022autoformer}, claiming state-of-the-art accuracy with an average of $\sim 0.86$ MSE (48\% lower) and $\sim 0.53$ MAE (33\% lower) compared to Informer. Using a new technique dubbed Auto-Correlation, they instead operate at the series level while ProbSparse operates at the point/query level, and also uniquely uses self-attention distilling in the encoder by halving layer inputs. Furthermore, Autoformer introduces a series decomposition block which progressively extracts and aggregates the long-term stationary trend from predicted intermediate results, allowing it to also capture general patterns in long sequences.

FEDformer \cite{zhou2022fedformer} hopes to also take the overall trend into account by combining the Transformer, capturing more detailed structures, with their seasonal-trend decomposition method and Frequency enhanced blocks, both capturing a more global view of the time series. Applying a Fourier transform to map the time series into the frequency domain had already been done in the 80s but this is the first time we see it in combination with a transformer -- and yielding superior results. FEDformer claims to further reduce the prediction error from Autoformer by 14.8 \% and 22.6 \% for both multivariate and univariate forecasting.

\subsubsection{New frontiers}
Using just a basic one-layer linear model directly mapping the input sequence to the output sequence, DLinear \cite{DLinear} claims Transformers are ineffective for time series forecasting. The linear models in this paper are simply linear layers (i.e., fully connected layers without non-linear activations) that map the input sequence directly to the output sequence. Linear is a vanilla linear model, NLinear first normalizes the input by subtracting the last value before applying a linear layer and adding the subtracted value back to handle distribution shift, and DLinear decomposes the input into trend and remainder components which are processed separately by linear layers before being summed for the final prediction. 

However, in 2022, PatchTST \cite{PatchTST} challenges this claim by slightly outperforming DLinear on most datasets. PatchTST introduces: (1) segmentation of time series into subseries-level patches as input tokens to the Transformer, and (2) channel-independence where each univariate time series is processed separately but shares the same embedding and Transformer weights. These modifications make PatchTST quite superior to other transformers and slightly better than using simple linear layers.

FITS (\textbf{F}requency \textbf{I}nterpolation \textbf{T}ime \textbf{S}eries Analysis Baseline) is a recently (2023) proposed model that leverages Fourier transforms in conjunction with a single complex-valued linear layer for time series forecasting. Despite this simplicity, it achieves comparable performance to state of the art models with significantly fewer parameters. In another effort to reduce parameter size, FITS uses a low-pass filter in the frequency domain to capture relevant patterns without with minimal impact on test performance. By training in the complex frequency domain, the neural network layer efficiently learns amplitude scaling and phase shifting to interpolate time series. This is a very innovative approach compared to anything seen before and is the only one (of the aforementioned) to introduce any filters on the data. FITS reports to outperform DLinear by 8.80\% on average.

%% file: researchsqs.tex
\subsection{Deep FITS}
FITS uses only a single linear layer without any activation functions and can be considered a form of linear regression on the frequency-time data produced using FFT. This simplicity facilitates its small profile in terms of parameters with fast inference and training time but it might also be a potential area of improvement. To explore whether enhancing this model could improve performance, we experiment with multiple deep variant of FITS. These variants consists of a configurable number of linear layers, ReLU-like activation functions, and dropout layers either in the upscaling part of the network, after the reverse Fast Fourier Transform or parallel to the frequency part of the network.

\subsection{Combining FITS with DLinear}

In the evolution of forecasting methods, there is a clear tendency for model improvements to stem from processing the data in different ways before making predictions, rather than simply scaling up the models or complexifying the model architecture. DLinear decomposes its input into trend and seasonal components, patterns we notice after some preliminary testing that FITS fails to capture, and models them separately with linear layers before being summed for the final prediction. 

FITS takes a different approach by first transforming the time series data into the frequency domain before that modeling with a complex-valued linear layer, and achieves superior results on some datasets. 

We hypothesize an integrated approach of the two models will lead to improved performance by leveraging the strengths of both methods.

\subsection{Measure of randomness and unpredictability in datasets \label{sec:motivation_hurst} {[\textcolor{red}{\textrightarrow} \ref{goback4}]}}

While reviewing other literature, we were astounded to find that for one of the datasets in \cite{DLinear}, the Repeat column, i.e $x_t = x_{t-1}$, beat all other models in spite their complexity or simplicity on short forecasting horizons. This could imply that if some datasets are sufficiently random, models will incorrectly capture noise as patterns and have worse prediction power.

In \cite{PatchTST}, it is argued (with citation to \cite{ROSSI}) that the toughest benchmark for forecasting is a random walk without drift, which closely resembles and describes the exchange rate dataset which had the aforementioned quirk. Apparently, the prevailing consensus among statisticians and economists is that if a market is "efficient", meaning all public information is already priced in and new information that affects stock prices are assumed to be unpredictable and random, the asset price becomes impossible to predict, and the best prediction for $x_t$ on average will just be $x_{t-1}$ as published by the acclaimed Eugene Fama in \cite{FAMA}. Intuitively, this also makes sense given that the mean of $n$ random walks as $n \to \infty$ will be equal to the drift (a straight line) [ref. fig. \ref{fig:compare_random_walk}] \label{goback2} as it is equally likely to be up or down.

We want to investigate whether we can find a general numerical measure of this unpredictability in other datasets as well. For that, we introduce the Hurst exponent.

%% file: methods.tex
\subsection{Complex Frequency Domain}
In time-series data, the signal is a series of observations in a chronological order, often with equally spaced points representing an amplitude corresponding to some quantity at a fixed sampling rate. This domain focuses on understanding how the signal evolves over time, identifying trends, seasonal patterns, and any anomalies or outliers that may exist. However, while the time-domain analysis provides valuable insights into how the signal changes over time, it often fails to reveal the underlying periodic structures and frequency content of the data.

This is where frequency analysis becomes incredibly useful. Frequency analysis allows us to decompose a time-domain signal into its sinusoidal components, each characterized by a specific frequency, amplitude, and phase. By transforming the signal from the time domain to the frequency domain, we can uncover its inherent periodicities and identify dominant frequencies.

In frequency analysis the frequency component of signals can be represented as a complex number, which captures both the amplitude and phase of the component. The exponential notation for such a number is:
$$
z = Ae^{i\theta}
$$
Where $A$ is the amplitude and $\theta$ is the phase of the number. This form is useful in the context of Fourier transforms where signals are decomposed into their sinusoidal components expressed by the complex exponential. 

To calculate the real and imaginary components of the number the rectangular form can be used:
$$
z = A \cos(\theta) + i A \sin(\theta)
$$

Where $A \cos(\theta)$ represents the real component of the number and $i A \sin(\theta)$ the imaginary component. 

\subsubsection{Fourier Transform}
A fourier transform is a broadly used method for decomposing a function of time, such as time series data, into the frequencies the function consists of. The Fourier transform essentially transforms data in the time domain to the frequency domain. The inverse function, a \textit{synthesis}, for Fourier transformation also exists, and converts the frequencies back into time domain data, which is also referred to as the inverse Fourier transform. Because our data is in the time domain, Fourier transforms are a very powerful tool to disassemble the different attributes (frequencies) in our data, especially if there are any patterns that are prevalent through an extended period of time.

The formula for a continuous Fourier transform is 
$$
F(\omega) = \int_{-\infty}^{\infty} f(t) \cdot e^{-i\omega t} \, dt
$$
Where $F(\omega)$ is the frequency-domain representation, $f(t)$ is the time-domain signal, and $e^{-i\omega t}$ is the basis function for decomposing a signal into its frequency components. When you integrate the function over a specific amount of time you get a function that represents amplitude and phase of each frequency component in the signal provided. The inverse Fourier transformation from frequency components to time components is given by:
$$
x(t) = \int_{-\infty}^{\infty} F(\omega) e^{i\omega t} , dt
$$

A continuous signal is problematic for computers to effectively work with and as such they are sampled with various rates. The length of your sampled signal will allow for greater representations in the frequency domain, and as such a faster sampling rate will give a greater representation at the cost of memory and storage. A critical concept for determining the sampling rate for a signal is the Nyquist-Shannon sampling theorem, which states that to accurately reconstruct a continous signal from the sampled version the sampling rate must be at least twice the highest frequency of the signal:
$$
f_s \geq 2 f_{max}
$$
Where $f_s$ is the sampling rate and $f_{max}$ is the highest frequency present in the signal. 

In many applications related with computer science, it is natural to sample a continuous signal to discretize it. Naturally, as the time series data provided is already in a computer-readable format, we need to perform discrete Fourier transforms (DFT). The formula for a discrete fourier transform is seen in equation {\ref{eq:fft}}.

\begin{equation}
    X_k = \sum_{n=0}^{N-1} x_n \cdot e^{-i \frac{2\pi}{N} nk}, \quad k = 0, 1, \ldots, N-1
\label{eq:fft}
\end{equation}

Using the above definition directly to calculate the DFT of a signal would be costly procedure. The reason is clear from the formula - For each data point, for which there are $N$ of, a sum of length $N$ is performed. As such, the time complexity of the processing is proportional to the squared length of the signal we wish to transform i.e., in big O notation it has the complexity of $O(N^2)$.

\subsubsection{Fast Fourier Transform}

To address this inefficiency, we use the Fast Fourier Transform (FFT), which is an optimized algorithm for computing the Fourier transform. The FFT significantly reduces the computational cost by exploiting the symmetry and periodicity properties of the DFT. The FFT algorithm reduces the time complexity to $O(N \log N)$, making it much more feasible for practical applications.

The FFT often used is the Cooley-Tukey algorithm, which operates by recursively breaking down a DFT of any composite size $N = N_1N_2$ into many smaller DFTs of sizes $N_1$ and $N_2$, simplifying the computation. This divide-and-conquer approach leverages the efficiency of smaller DFTs to compute the larger one more quickly. The steps of the Cooley-Tukey algorithm are described in further detail in the appendix \ref{appendix:FFT}. \label{goback6}

\textbf{Real Valued Fast Fourier Transform}

A simple increase in computational efficiency is possible when working with real valued data, $X \in \mathbb{R}$. In such case, only approximately half the computations are required to perform the FFT, as the output will be complex conjugates of one-another:

\begin{equation*}
    X_k = \overline{X_{N-k}}, \quad \forall k \in \left[1, \left\lfloor\frac{N}{2}\right\rfloor\right]
\end{equation*}

\subsection{FITS -- Frequency Interpolation (for) Time Series}

In the project, the FITS model is written and reimplemented by the team, in order to assess and evaluate it on the original paper's datasets as well as some additional ones. The FITS model as described in the original paper does a series of simple data manipulations and transformations. The pipeline of the FITS model can be seen in the figure \ref{fig:recon_pipeline} below.

\begin{figure}[H]
    \centering
    \includegraphics[width=\textwidth]{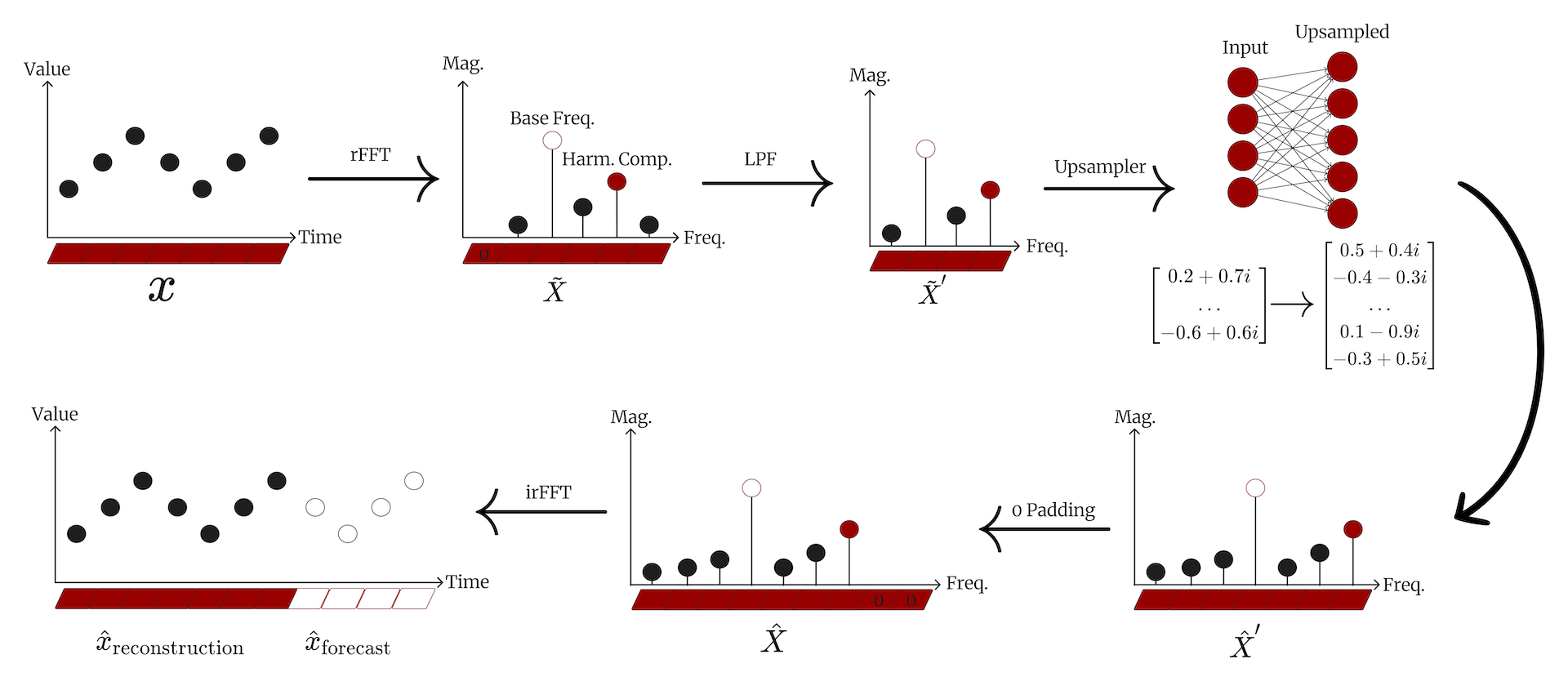}
    \caption{FITS pipeline for reconstruction task}
    \label{fig:recon_pipeline}
\end{figure}

During training and a forward pass through the model, the time series data batch is normalized to zero-mean and unit variance $\hat{x} = \frac{x - \mu}{\sigma}$ followed by a discrete Fourier transform calculated using the real valued Fast Fourier transform variation of equation {\ref{eq:fft}}.

One of the methods realized by the authors of the paper, is that for a time series projected into the equivalent frequency domain, the high frequencies in the model usually only represent the noisy terms in the time series, which are not important to the general trends in the series. By using a low-pass filter, the effective size of the model from that point onward is reduced with a marginal error increase. The challenge is choosing a cut-off frequency such that the frequencies that remain still represent and maintain the time series' trends without cutting away important information that could be important for the forecasting or reconstruction tasks.

\begin{figure}[H]
    \centering
    \includegraphics[width=1\linewidth]{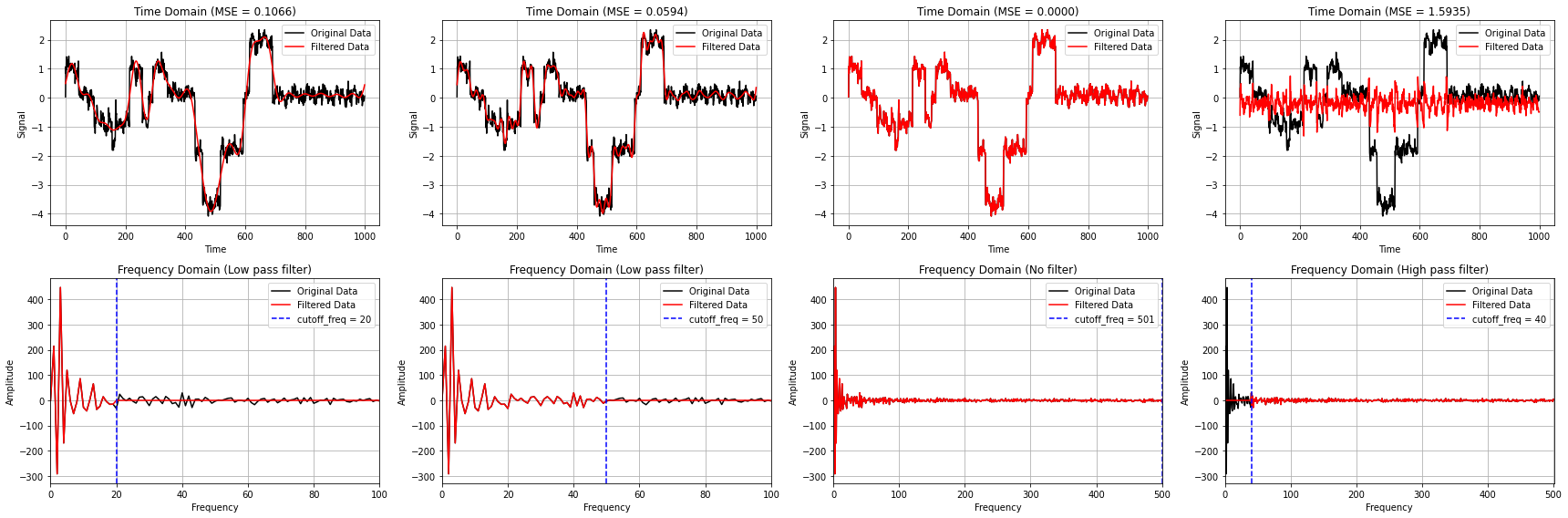}
    \caption{The effect of different cutoff frequencies using a low pass and high pass filter}
    \label{fig:enter-label}
\end{figure}

After the low pass filter on the data in the frequency domain has been applied, the data is run through a complex linear layer with input dimensions corresponding to the cut-off frequency outputting complex frequencies based on a interpolation rate $\eta$ which is calculated based on the model's output length $L_o$ and corresponding input length $L_i$. 

$$
\eta_{freq} = \frac{L_o}{L_i}
$$

After which the inverse fast Fourier transform (irFFT) is performed to project the new frequencies back into the time-domain:
$$
\hat{x}_n = \frac{1}{N} \sum_{k=0}^{N-1} X_k e^{i 2 \pi \frac{k}{N} n} \quad \text{for } n = 0, 1, \ldots, N-1,
$$
where $\hat{x}_n$ are the reconstructed samples in the time domain where inverse normalization is then performed to revert the data back to its original scale: $x = \hat{x} \sigma + \mu$

Projecting the signal back into the time domain allows the model to learn using standard loss functions such as the MSE Loss. Depending on the task, a look-back window and a horizon is generated for fore-casting, whereas a down-sampling of the original time series is done, and a reconstruction loss is computed for reconstruction tasks. 

When the signal is multichannel the model can be configured in two ways - with the linear layer shared between the channels or with each channel having an independent linear layer. In many time-series datasets the channels often share a base frequency as the data is captured from the same system, when this is the case sharing the linear layer is sufficient and cuts down on total parameters. When this doesn't hold and the signals don't share a common base frequency having individual layers can be a better option at the cost of an increase in parameters.

Another task that this projects aim to evaluate the FITS model on is anomaly detection. Anomaly detection for time series can happen with regards to signal gathering in the form of artifacts or simply data points that are of special interest. In both regards, identifying such points can be very beneficial before working with the data further.

In the reconstruction task, the original data is down-sampled first according to, and then run through a similar pipeline as the forecasting task, but here up-sampled to the same length as the segment of the timeline one wish to compare with. The loss is then a reconstruction loss with a comparison between the up-sampled time series data and the output.

In the project, the FITS model have been implemented with inspiration from the original code. The code and model is annotated with better naming for variables and functions, to help make it easier to work with should modifications be added, as well as a solid testing framework to ensure the model in the project trains exactly the same as the model that is introduced in the paper - when subject to the same hyper parameters and testing data. 

Furthermore, the training scripts and data loaders are rewritten, much like how the model is rewritten, to allow for easier readability and extension. The dataset and dataloading classes have been made more general and customizable making it easy to use custom datasets, add time series augmentations, while features used in transformer-based models such as datetime-specific features have been discarded. This allows us to experiment faster with more datasets.

To evaluate the FITS model and observe if we successfully reproduced the results from the paper, we use scripts that mimic the training properties of the original authors, and run it on the same datasets that they describe in the paper, such that the comparison and evaluation between the models are subject to the same constraints and can be compared fairly.\cite{xu2024fits}

\subsection{Interpreting FITS}
At its core FITS interpolates from the frequency components of a given input to that of a longer signal using a single weight matrix. To understand this process it makes sense to look at simple signals to clearly visualize how it works with the frequency domain.

\begin{figure}[H]
    \centering
    \includegraphics[width=\linewidth]{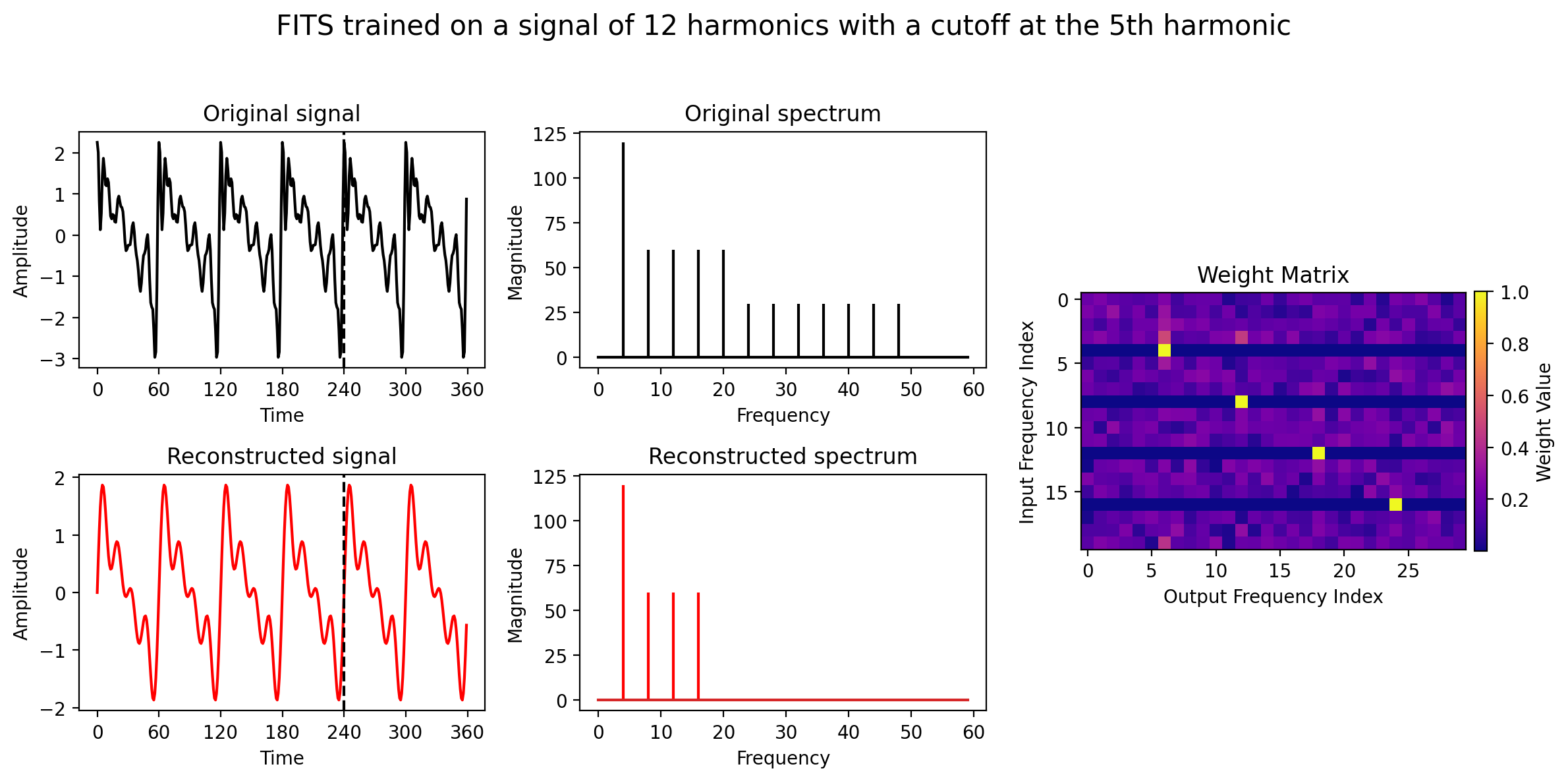} 
    \caption{FITS trained with a cutoff at the 5th harmonic of a signal of mixed sines at higher harmonics with a dominant period of 60. The left most plots show the original and reconstructed/extrapolated signal with the vertical line visualising the split between input and prediction. Middle plots are the frequency representations of the signal and the reconstruction. Right plot is the absolute value of the complex weight matrix of FITS for this data showing high activations corresponding to the harmonics.}
    \label{fig:FITS_explained1}
\end{figure}

\begin{figure}[H]
    \centering
    \includegraphics[width=\linewidth]{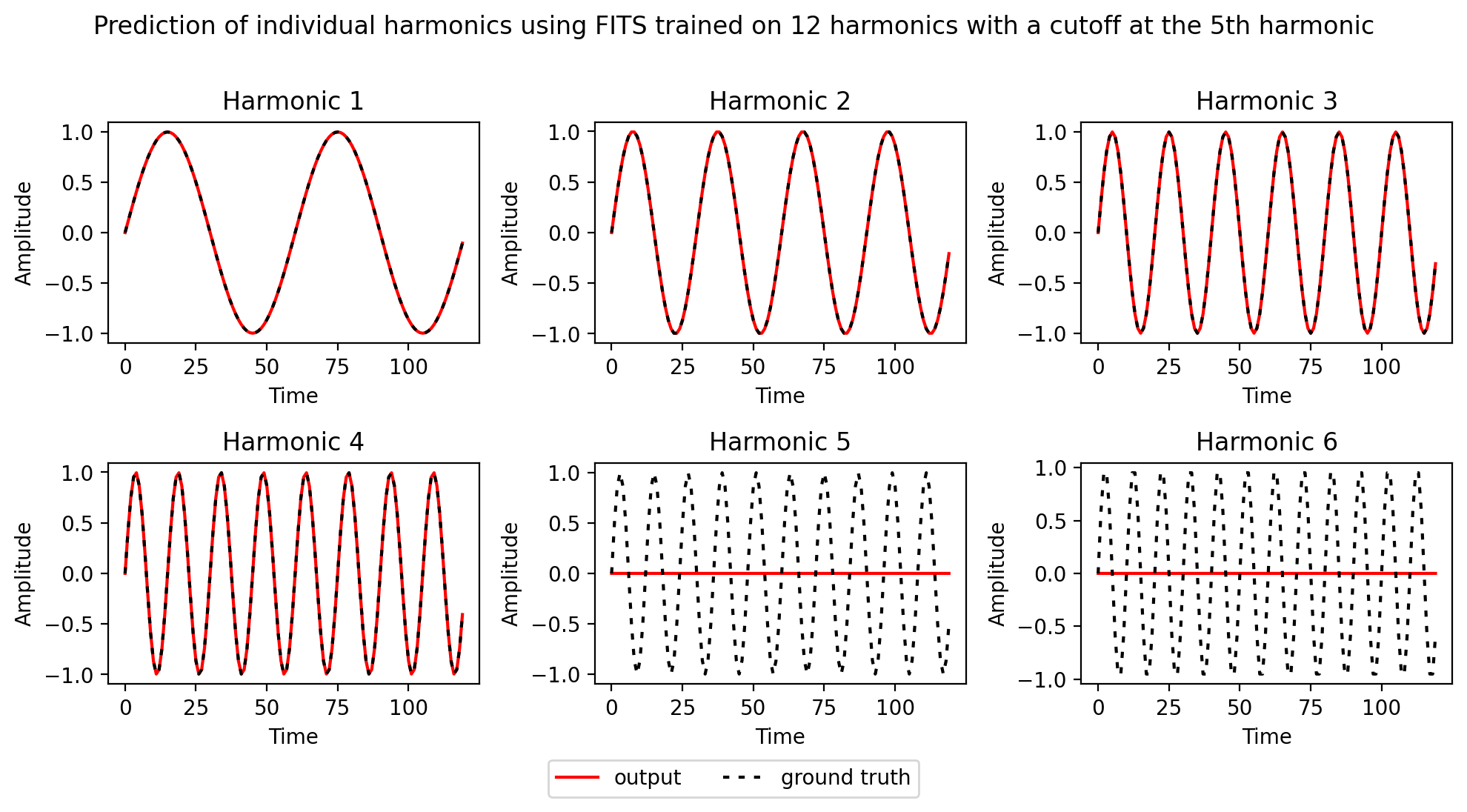}
    \caption{Inference of FITS on the harmonic base components illustrating the inherent separation of the frequency components of the signal.}
    \label{fig:FITS_explained2}
\end{figure}

When training FITS on a signal consisting of a combination of a base sine wave and multiple sine waves of higher frequency based on the higher order harmonics of the base signal it's possible to reconstruct the signal up to the cutoff frequency corresponding to the harmonic order of the signal chosen. In figure \ref{fig:FITS_explained1} this is visualized up to the 5th harmonic where everything above is discarded as it is considered "noise" due to the choice of cutoff. As the signal is deconstructed to its frequency components by the FFT, the network is able to reconstruct and predict on signals consisting of arbitrary combinations of these component frequencies up to the cutoff as seen in figure \ref{fig:FITS_explained2}. This allows FITS to interpolate between the seen frequencies of signal and makes it robust in making predictions on time series data where the signal has inherent periodicity. Even though this example is very simple it illustrates the major mechanisms of FITS with weight values corresponding the frequencies of the harmonics of the input signal to the harmonics extrapolated in the reconstruction which can be seen for a complex multichannel dataset in the appendix: \ref{fig:electricity_weights} \label{goback5}

\begin{figure}[H]
    \centering
    \includegraphics[width=1\linewidth]{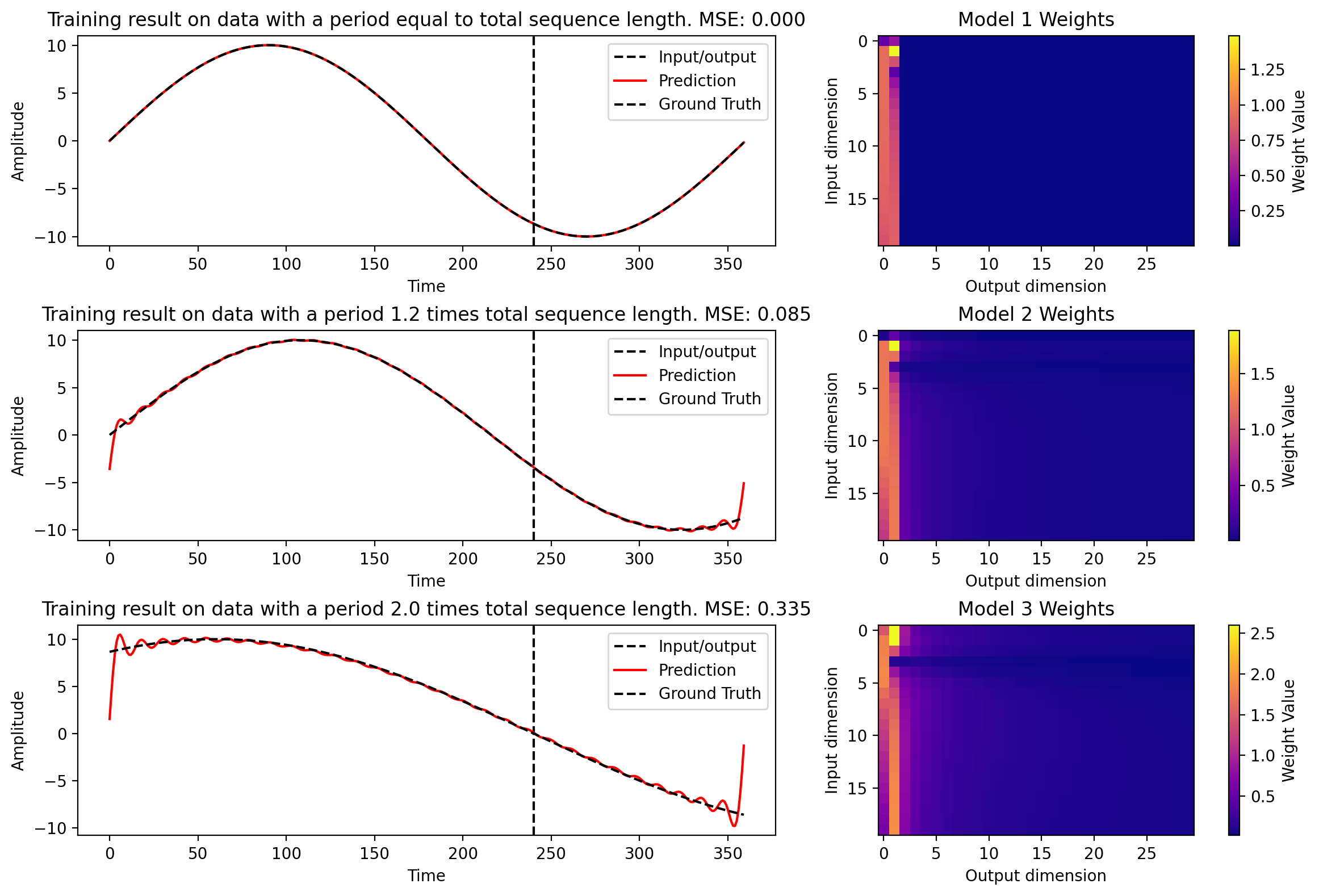}
    \caption{Increasing the length of the period of the signal that is being trained on to longer than the output allows. Output and original signal to the left and weights for each of the models to the right illustrating spectral leakage.}
    \label{fig:FITS_explained3}
\end{figure}

Where FITS struggles is with signals where a major period is greater than the total sequence length of the output. Here, it’s not possible to capture that signal as the dynamics of the period do not fit into the window. This failure mode can be seen most clearly at the beginning and end of the reconstruction, where specific artifacts appear due to the inclusion of higher frequency components than the true signal. This might occur because of spectral leakage, which happens when the signal period does not match the observation window. Leakage causes energy to spread across multiple frequency bins, leading to inaccuracies in the reconstructed signal. Plotting this, it becomes clear that the reconstructed signal is limited by the mismatch between the signal period and the window length, causing the observed artifacts. This artifact becomes more severe the further the period of the signal is from the output length, the effect of which can be seen in figure \ref{fig:FITS_explained3}. Here the weights for each of the models are also plotted in which one can see that there is spectral leakage as there are components contributing to the output in the higher frequencies of the output dimension. The models have been initialized with 0 in the weight matrix to clearly visualize the higher frequency components erroneously learned. 

When training on a multichannel signal FITS can either have $n$ number of independent linear layers corresponding to the number of channels in the signal or have one linear layer with shared weights between the $n$ channels. Using the individual configuration provides better reconstruction quality measured in MSE if the channels do not share the same frequency components.

\subsection{Deep FITS}

To explore adding layers to the complex domain in FITS, modifications to the ReLU activation function had to be introduced since ReLU is incompatible with complex numbers. We implemented the two variants ModReLU and $\mathbb{C}\text{ReLU}$ as described in \cite{deep_complex_networks} to translate ReLU to the complex domain.

$\mathbb{C}\text{ReLU}$ extends ReLU by considering the real and imaginary component for the ReLU operation as separate components:

$$
\mathbb{C}\text{ReLU}(z) = \text{ReLU}(\Re(z)) + i \text{ReLU}(\Im(z))
$$

Note that the phase of the number $z$ is modified due to the separate application of the ReLU function on the real and imaginary parts. Specifically, the magnitude of the complex number can change depending on whether the real or imaginary part (or both) are non-negative. If both parts are non-negative, the magnitude increases. If either part is negative, that part is set to zero, potentially reducing the overall magnitude. The original angle (or phase) of the complex number $z$ can be altered significantly. For instance, if $z$ lies in the third or fourth quadrant (where both the real and imaginary parts are negative), applying the ComplexReLU function will result in $\mathbb{C}\text{ReLU}(z) = 0$, thus collapsing the phase information to zero.

ModReLU works by only considering the magnitude $|z|$ of the complex number $z$ in ReLU together with a learnable parameter $b$ and scaling $z$ by it:
$$
\text{modReLU}(z) = \text{ReLU}(|z| + b) e^{i\theta_z} \begin{cases} (|z| + b)\frac{z}{|z|} \quad &\text{if } |z| + b \geq 0, \\ 0 \quad &\text{otherwise,} \end{cases}
$$

The parameter $b$ is introduced as the magnitude is complex number is never negative and ReLU would therefore not have any effect. This introduces a learnable dead-zone below which the complex number will be set to 0. 

ModReLU adjusts the activation by shifting the magnitude of the complex number by the parameter $b$. This    shift allows the network to learn a threshold below which the activations are set to zero, effectively creating a dead-zone. When $|z| + b \geq 0$, the magnitude is shifted positively and the complex number retains its phase $\theta_z$. If the condition $|z| + b < 0$ is met, the output is set to zero, meaning both the real and imaginary parts are zeroed out, thereby affecting both the magnitude and phase.

Based on experiments in \cite{deep_complex_networks} $\mathbb{C}\text{ReLU}$ seems to perform better than $\text{modReLU}$ in image classification tasks used in a convolutional neural network but as our model is quite different and uses linear layers we intend to test both. 

We experimented with adding linear layers with ModReLU and complex dropout in the upsampler and only placing the activation function between the deep layers. 

Another variant of the network we tried utilizing linear layers after the reverse Fast Fourier Transform ie after the upscaling. The numbers are now in the real domain and we thus don't need to utilize the complex neural network modifications.

Finally, an experiment was conducted by adding a linear layer that simply maps input to output and sums up the output with the standard FITS output. The idea is that any general drift or linear trend in the data might be better captured by a linear mapping from input to output. As such, a trainable parameter was introduced as well as a linear layer. When reconstructing and forecasting the signal, a sum of both of frequency interpolated signal and the linear layer would happen:

\begin{equation*}
    xy = (1 - \beta)\cdot\overline{xy} + \beta\cdot xy_{linear}
\end{equation*}

\subsection{Hybrid Models}

\subsubsection{DLinear}
The original DLinear model first decomposes the input time series into a trend component and a remainder (seasonal) component using a moving average filter with a kernel size of 25. The trend component captures the long-term progression of the time series, while the remainder represents the seasonal or periodic fluctuations. It then uses two separate linear layers to learn the patterns of each component and sums them to produce the final prediction. 

\subsubsection{DLinear + FITS}

In DLinear + FITS, we extend DLinear by incorporating FITS on the residual signal. After applying the standard DLinear decomposition and training the respective linear layers to produce predictions for the trend and seasonal parts, we compute the residual. The residual in this context is the difference between the original input time series and the sum of the predictions from DLinear during training. It represents the part of the signal that DLinear's trend and seasonal components couldn't explain or predict. This residual signal is then passed through and trained in a standard implementation of FITS to hopefully capture any remaining patterns or fluctuations that were not captured by the previous components. Finally, the predicted trend, seasonal components, and the forecast portion of FITS output are summed to produce the final prediction.

\subsubsection{FITS + DLinear}

In the FITS + DLinear model, we reverse the order of operations compared to DLinear + FITS. First, the input time series is processed through FITS, which as we know, normally returns its predictions for the entire sequence and forecast horizon. Its output then serves as training data for DLinear which produces the final prediction. This approach differs from DLinear + FITS in that FITS is applied to the entire input signal rather than just the residual of DLinear's predictions. By applying FITS first, we allow the model to capture the complex frequency patterns, ideally the main cyclical components in the data, before going through the seasonal and trend decomposition steps (and subsequent training). This may help in situations where the input signal contains periodic patterns that are not as easily separated by the moving average filter. The subsequent DLinear model can then focus on refining its predictions based on the "denoised signal" provided by FITS rather than the raw "noisy" input data. Although in hindsight, we could maybe do away with the moving average filter completely, as FITS essentially serves as an enhanced version of that (see figure \ref{fig:ma-vs-lowpass} below) or not train FITS, as the low pass filter might be enough by itself.

\begin{figure}[H]
    \centering
    \includegraphics[width=1\linewidth]{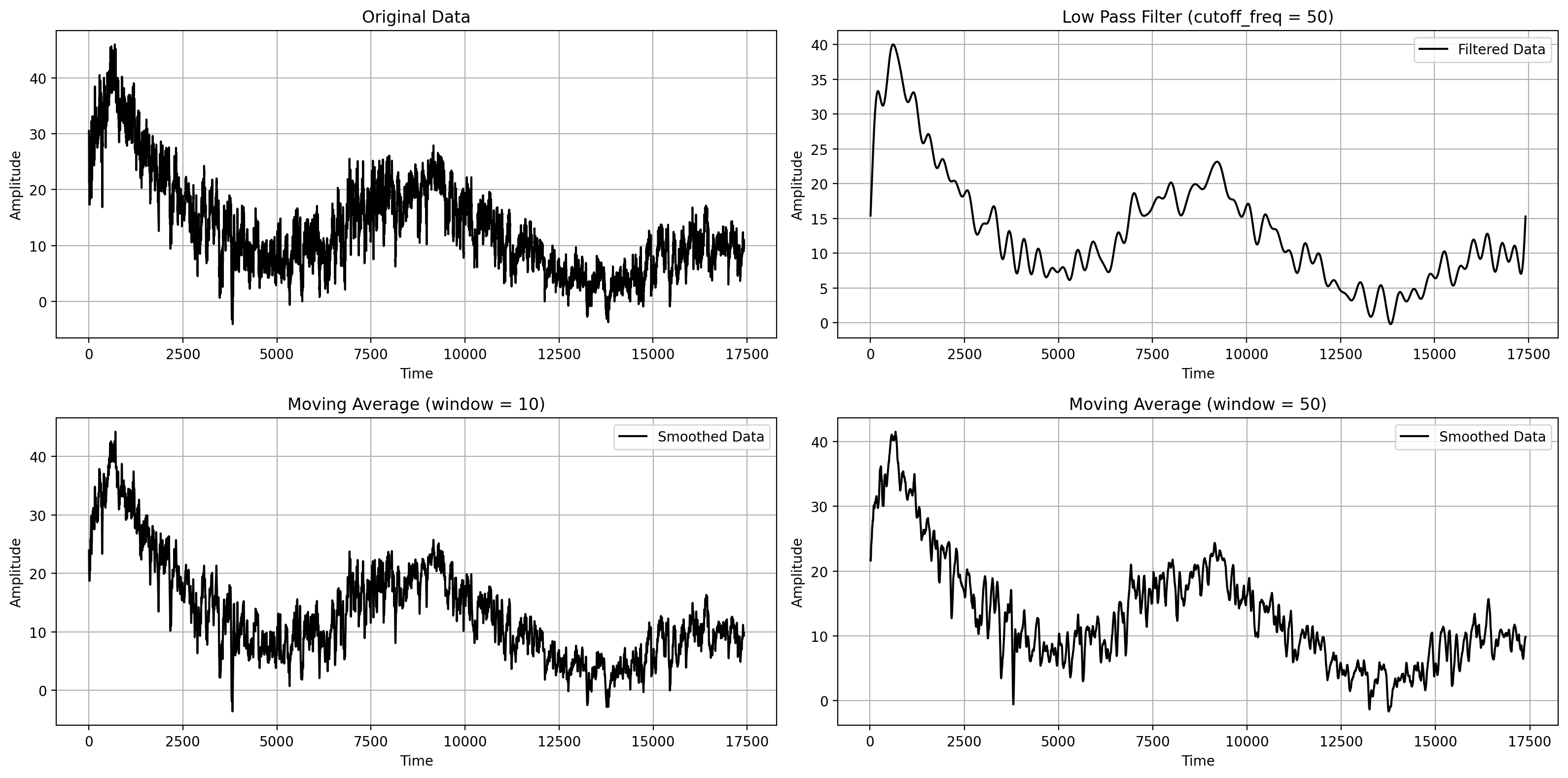}
    \caption{The effects of a moving average filter vs a low pass filter}
    \label{fig:ma-vs-lowpass}
\end{figure}

The code for the three models can be found here:

\href{https://github.com/thorhojhus/ssl_fts/tree/DLinear/models}{https://github.com/thorhojhus/ssl\_fts/tree/DLinear/models}

\subsection{Hurst exponent}

The Hurst exponent $H$ seeks to provide insight into the behavior or predictability of a time series by quantifying the tendency of a dataset to exhibit trending, mean-reverting, or random walk behaviour. $H$ always takes a value between 0 and 1 and is formally defined as\href{https://en.wikipedia.org/wiki/Hurst_exponent}{$^{1}$}$^{,}$\href{https://www.sciencedirect.com/topics/engineering/hurst-exponent}{$^2$}:

$$\mathbb{E} \left[\frac{R(n)}{S(n)}\right] = C n^H \text{ as } n \rightarrow \infty$$

\begin{figure}[H]
    \centering
    \includegraphics[width=.98\linewidth]{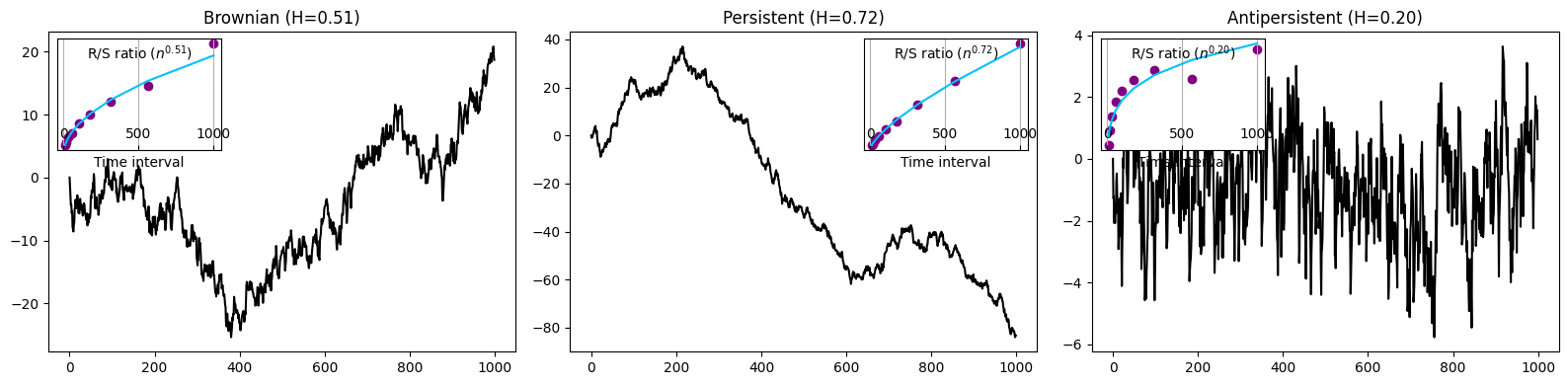}
    \includegraphics[width=.98\linewidth]{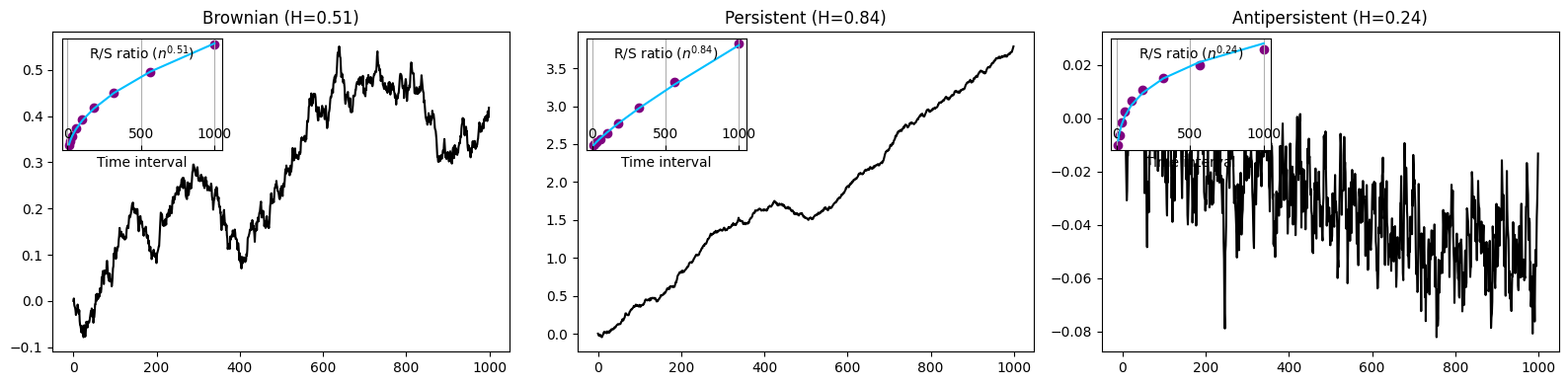}
    \caption{1 simulation (top row) \& Mean of 10000 simulations (bottom row) and their mean R/S}
    \label{fig:figure8}
\end{figure}

Where
\begin{align*}
R(n) &= \max \left(Z_1, Z_2, \ldots, Z_n\right)-\min \left(Z_1, Z_2, \ldots, Z_n\right) \tag{Calculate the range $R$} \\
Z_t &= \sum_{t=1}^t Y_t \quad \text { for } t=1,2, \ldots, n \tag{Cumulative deviation from the mean-adjusted series} \\
Y_t &= X_t-\mu \quad \text { for } t=1,2, \ldots, n \tag{Remove trend, keep fluctuations} \\
\mu &= \frac{1}{n} \sum_{t=1}^n X_t \tag{Mean of the series $X$} \\
\end{align*}
And
\begin{align*}
S(n) &= \sqrt{\frac{1}{n} \sum_{i=1}^n\left(X_i-\mu\right)^2} \tag{Standard deviation $S$ of the original series $X$} \\
\end{align*}
$H$ is then fitted as a straight line using least-squares regression on $\log [R(n)/S(n)]$ as function of $\log n$ where the slope gives $H$ (ref. figure \ref{fig:figure8}).

The rescaled range $R(n)/S(n)$ is really just a measure of the variability of a time series relative to its standard deviation. This means that the it captures the idea of how much the cumulative deviations from the mean ($R(n)$) vary compared to the typical size of fluctuations in the original series ($S(n)$).

If the time series is purely random (e.g., brown noise or random walk), the cumulative deviations from the mean ($R(n)$) will grow roughly proportionally to the square root of the series length ($n^{H=0.5}$). This is because random fluctuations tend to cancel each other out over time. This indicates short-memory, meaning each value is independent of each other and there is no correlation.

However, if the time series has long-term memory or persistence (no noise), meaning trending behaviour, i.e. high values to be followed by higher value and lower values tend to be followed by lower values, the cumulative deviations from the mean ($R(n)$) will grow faster than the square root of the series length($n^{H>0.5}$). This is because the positive (or negative) deviations tend to maintain their direction (positive or negative deviations) for longer periods than would be expected by chance, leading to larger cumulative deviations. 

Conversely, if the rescaled range $R(n)/S(n)$ grows slower than $H=0.5$, so $n^{(H<0.5)}$, it suggests that the time series is anti-persistent, meaning high values are more likely to be followed by low values, and low values are more likely to be followed by high values, with a tendency to revert to the mean value over time (white noise / quickly fluctuating series).

All Hurst Exponents used in plots have been calculated using this python library:

\href{https://pypi.org/project/hurst/}{https://pypi.org/project/hurst/}.

\subsection{Autocorrelation function (ACF)}

The autocorrelation function (ACF) is a statistical tool used to measure the correlation between a time series and a lagged version of itself. It quantifies the degree of similarity between observations as a function of the time lag between them. The ACF is also useful for identifying patterns, such as persistence or mean reversion, in a time series.

The ACF is defined as\href{https://www.sciencedirect.com/topics/chemistry/autocorrelation-function}{$^3$}:

$$\rho(k) = \frac{\text{Cov}(X_t, X_{t+k})}{\text{Var}(X_t)}$$
Where
\begin{align*}
\operatorname{Cov}(X_t, X_{t+k}) &= \frac{1}{n-k} \sum_{t=1}^{n-k} (X_t - \mu)(X_{t+k} - \mu) \tag{Autocovariance at lag $k$} \\
\operatorname{Var}(X_t) &= \frac{1}{n} \sum_{t=1}^{n} (X_t - \mu)^2 \tag{Variance} \\
\mu &= \frac{1}{n} \sum_{t=1}^n X_t \tag{Mean} \\
\end{align*}

The ACF is related to the Hurst exponent in that they both provide information about the memory or persistence of a time series. A time series with long-term memory or persistence a ($H$ close to 1) will typically exhibit slowly decaying autocorrelations and remains positive for many lags, because it mostly correlates with itself. Conversely, a time series anti-persistent behaviour ($H$ close to 0) will show rapidly decaying and quickly oscillating autocorrelations because of its erratic nature. If the time series is completely uncorrelated exhibiting no memory ($H=0$), meaning all steps are i.i.d., stationary, and uncorrelated, the ACF should be close to zero for all lags $k > 0$. Note that this is different from a random walk as every next time step becomes its new stationarity, therefore exhibiting some autocorrelation that decays slow due to the dependence of each step on the previous step.

In figure \ref{fig:figure9} below, the ACF plots for 10000 simulated random walks show similarly decaying autocorrelations to those descriptions.

\begin{figure}[H]
    \centering
    \includegraphics[width=1\linewidth]{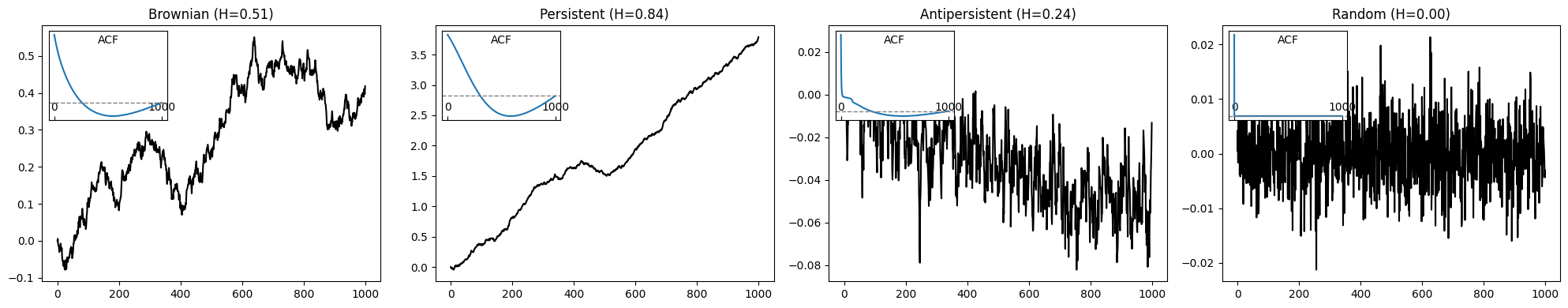}
    \caption{Mean of 10000 simulated random walks and their ACF}
    \label{fig:figure9}
\end{figure}

\begin{figure}[H]
    \centering
    \includegraphics[width=1\linewidth]{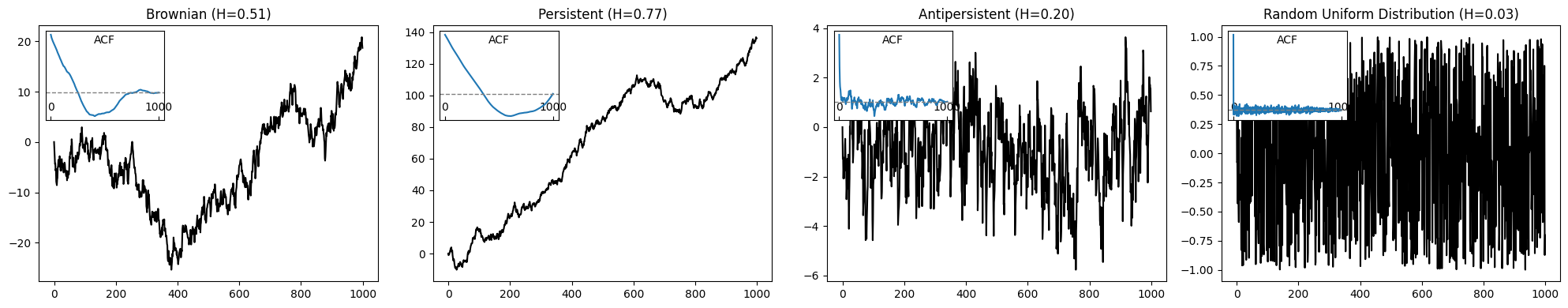}
    \caption{1 simulated random walk and its ACF}
    \label{fig:figure10}
\end{figure}

However, an ACF plot for a single example is much more likely to resemble something like figure \ref{fig:figure10} below as taking the mean smooths the otherwise common distinguishable characteristics of the three variants. 

\subsection{Baseline Models in forecasting}
As forecasting can be a difficult task, some of the simplest models can perform quite well. As we recall from section \ref{sec:motivation_hurst} \label{goback4}, if we assume a random walk without drift, meaning each data point is independent and random, no model will predict better than the simple yet surprisingly effective forecasting model \textbf{Naïve Forecasting Model}. The idea is simple; predict the next $L$ time steps to be equal to the last observed value:
\begin{equation*}
    \hat{z}_{T+l|T} = z_T,\quad\forall l \in [1\cdots L]
\end{equation*}

This provides a solid baseline for which to benchmark more sophisticated models against. If this baseline model reliably predicts better than the rest, it strongly hints that the dataset is fundamentally \underline{unpredictable}.

Another straightforward and simple baseline for time series data is simply to forecast the next $L$ values as the mean of the observed time series, a \textbf{Mean Valued Forecast}:

\begin{equation*}
    \hat{z}_{T+l|T} = \mu_z,\quad\forall l\in[1\cdots L] 
\end{equation*}

A simple \textbf{linear regression} on the time series data could also serve as a baseline but for reasons will be expanded on later, has shown to be an inferior baseline and will not be utilized.

These are the the most simple forecasting models, and we include the Naïve forecasting as a baseline comparison in the results section on page \pageref{sec:results}.

\subsection{Autoregressive Integrated Moving Average Models (ARIMA)} \label{goback3}
Moved to appendix. \ref{sec: ARIMA}

%% file: data.tex
\subsection{Real data}
Make descriptions of how the data looks plotted and asses any potential for forecasting on the data such as any patterns that are present throughout the data (or parts of it)

We use 9 common datasets used to benchmark time-series based models \cite{wu2022autoformer}. This is used as a baseline to easily reproduce previous papers' results and compare new modifications. Their descriptions are outlined in the tables below. Summary statistics have been omitted from the tables below because of the vast number of independent variables.

\texttt{electricity.csv} contains the hourly electricity consumption of 321 different clients in kWh from 2016-07-01 to 2019-07-02.\href{https://drive.google.com/drive/folders/1ZhaQwLYcnhT5zEEZhBTo03-jDwl7bt7v}{$^1$}

\texttt{ETTh1.csv}, \texttt{ETTh2.csv}, \texttt{ETTm1.csv}, and \texttt{ETTm2.csv} are all datasets of Electricity Transformer Temperature where each column measures different loads and their usefulness and \textbf{O}il \textbf{T}emperature (OT) from 2016-07-01 to 2018-06-26.\href{https://github.com/zhouhaoyi/ETDataset}{$^2$} %

\texttt{traffic.csv} contains hourly recordings of road occupancy rates from 862 different sensors on San Francisco freeways between 2015-01-01 and 2015-04-27 from California Department of Transportation.\href{https://zenodo.org/records/4656132}{$^3$}

\texttt{national\_illness.csv} contains weekly recorded influenza-like illness (ILI) from CDC in United States from 2002-01-01 and 2022-06-30.\href{https://gis.cdc.gov/grasp/fluview/fluportaldashboard.html}{$^4$}

\texttt{weather.csv} is a collection of German weather data with 21 different meteorological indicators.\href{https://www.bgc-jena.mpg.de/wetter/}{$^5$}

\texttt{exchange\_rate.csv} is a collection of daily exchange rates of eight foreign countries including Australia, British, Canada, Switzerland, China, Japan, New Zealand and Singapore ranging from 1990 to 2010.\href{https://github.com/laiguokun/multivariate-time-series-data}{$^6$}

\texttt{\$GD} and \texttt{\$MRO} stock data collected from the python library \textit{yfinance}\href{https://github.com/ranaroussi/yfinance}{$^7$}. Comprises of stock data changes over 15723 days, or approximately 43 years, of stock data. The columns consist of "Open", "High", "Low", "Close", "Volume", "Adj Close".

Furthermore, some of the data has been cleaned of invalid or extreme outlier values, like -9999 for certain variables in \texttt{weather.csv}. We have replaced those values with interpolation between the previous and next value of the corresponding variable. Other than that, no processing has been made to any of the other datasets.

\subsubsection*{Most common benchmarking datasets}

\begin{table}[H]
\centering
\label{tab:datasets_part1}
\begin{tabular}{lccccc}
\toprule
Dataset & Electricity (ECL) & ETTh1 & ETTh2 & ETTm1 & ETTm2 \\
\midrule
Prediction ($y$) & Last col. & OT & OT & OT & Channels \\
Description of $y$ & kWh  & Oil Temp. & Oil Temp. & Oil Temp. & Oil Temp. \\
Sampling Rate  & 1 hour & 1 hour & 1 hour & 15 min & 15 min \\
Channels (columns) $\times$ & 321 & 7 & 7 & 7 & 7 \\
Total Timesteps (rows) & 26,304 & 17,420 & 17,420 & 69,680 & 69,680 \\
Hurst Exp. ($y$) & 0.2449 & 0.3191 & 0.2322 & 0.3471 & 0.2319\\
\bottomrule
\multicolumn{6}{l}{\textbf{Note}: Oil Temp.: Oil temperature. Last Col.: Last column in the dataset (index -1)}\\
\multicolumn{6}{l}{\hspace{1cm} Expanded table 7 from \cite{xu2024fits}}
\end{tabular}
\end{table}

In each dataset, the last column has been renamed Original Target (OT) by other papers with no mention of it what it was before. Presumably, this was done in order to fit with the data loader. Therefore, some of the column descriptions are unknown. 

In the appendix plots of all the channels in the datasets are provided and their autocorrelation over time. From those we see clearly daily periods in all the datasets but illness, where there instead is a yearly period. 

We have decided to include two more datasets for benchmarking. The first one, \$GD, was collected using the yfinance python library to download all stocks in the S\&P500\href{https://github.com/thorhojhus/ssl_fts/blob/main/notebooks/stock_data.ipynb}{$^6$} for all available intervals (1m, 2m, 5m, 15m, 30m, 1h, 1d, 5d, 1wk, 1mo, 3mo) and then sorted in descending order by its number timesteps and afterwards its deviation from $H=0.5$. The second one, \$MRO, was collected the same way except the data was sent through a low pass filter with cutoff frequency 200, mimicking the FITS pipeline and removing "noise", before being sorted. The rationale for this method is that theoretically, these two stocks should resemble random walks the least out of the 500 and therefore have the least "unpredictability".

\begin{table}[H]
\centering
\label{tab:datasets_part2}
\begin{tabular}{lcccc|cc}
\toprule
Dataset & Traffic & Illness & Weather & Exchange rate & \$GD & \$MRO \\
\midrule
Prediction ($y$) & Last col. & OT & OT & OT & Adj Close & Adj Close \\
Description of $y$ & Road occ.  &  &  & FX rate & Price & Price\\
Sampling Rate  & 1 hour & 1 week & 10 min & 1 day & 1 day & 1 day\\
Channels (columns) $\times$ & 861 & 6 & 20 & 7 & 6 & 6\\
Total Timesteps (rows) & 17544 & 966 & 52696 & 7588 & 15723 & 15723 \\
Hurst Exp. ($y$) & 0.1637 & 0.3410 & 0.3249 & 0.5002 & 0.5792 & 0.4982\\
Hurst Exp. ($y_{\text{low\_pass}}$) &  &  &  &  & 0.3828 & 0.3086\\
\bottomrule
\multicolumn{7}{l}{\textbf{Note}: Road occ.: Road occupancy rates. Descriptions left blank are unknown.} \\
\multicolumn{7}{l}{\hspace{.8cm} A Hurst Exponent of 0.5 indicates random walk behaviour} \\
\end{tabular}
\end{table}

It is possible to train on these datasets in three different ways:
\begin{enumerate}
    \item Univariate input and univariate target, i.e. predicting $y$ based on only the previous rows of the same column (univariate regression)
    \item Multivariate input and univariate target, i.e. predicting $y$ based on all other variables (multivariable regression)
    \item Multivariate input and multivariate target, i.e. predicting all columns besides the \texttt{datetime} column based on every other column. (multiple auto regression)
\end{enumerate}

For benchmarking FITS against the previous other state-of-the-art models, we rely on the results from a previous 2023 paper \cite{SHEN2023953} as retraining all the models would be infeasible and train only on multivariate input with multivariate prediction.

\subsection{Synthetic data}

In order to help identify some of the models weaknesses and strengths, we developed a framework for generating synthetic data to allow us to manipulate trends, noise, and other characteristics. The purpose of this is to help us see how the nature of the data influences FITS' predictive power.\href{https://github.com/thorhojhus/ssl_fts/blob/main/src/synthgen.py}{$^7$}

We have not used any synthetic datasets for official benchmarking but it has been used for interpreting and experimenting with FITS.

\subsection{Training Setup}

Our training process begins by calculating the training loss on the combined input and output of the signal, providing the model with the most comprehensive information. Afterwards, we enter a fine-tuning stage, where the loss is computed between the prediction part of the network and the unseen part of the signal. In summary, we first consider both $x$ and $y$, and then focus solely on $y$.

We use a training, validation, and test split of the datasets to train the models effectively. To prevent overfitting on the training set, we use the validation set to adjust the learning rate and implement early stopping if no improvement is observed within a specified number of epochs. At the end of each stage, we calculate the test loss in terms of Mean Squared Error (MSE) exclusively on the prediction part of the output.

The dataset is split into training, validation, and test sets in a 70:10:20 ratio respectively. For ETT we split 60:20:20 as that is what others report on.

%% file: results.tex
Below we report the results of our various models and baselines on all the different dataset trained and test with multivariate input and multivariate output similar to the reported 

\subsection{Common benchmark datasets}

To reproduce the results reported in \cite{xu2024fits} we run with exactly the same model parameters and splits on the datasets as reported and read from their source code. All the runs are done mutivariate and we report the MSE over all the channels in the dataset

The parameters for the model are as follows:

For the ETTh1 dataset, the lookback window (seq\_len) is set to 360, the base period (base\_T) is 24 hours, and the harmonic order (H\_order) is 6. In contrast, the ETTh2 dataset uses a lookback window of 720, with the remaining parameters unchanged. Both ETTm1 and ETTm2 share the same parameters: a lookback window of 720, a base period of 96 (representing 15-minute intervals per day), and a harmonic order of 14. The Electricity dataset features a lookback window of 720, a base period of 24 hours, and a harmonic order of 10. Similarly, the weather dataset has a lookback window of 720, a base period of 144 (equivalent to 10-minute intervals per day), and a harmonic order of 12 but notably is the only dataset that is trained on in individual mode - having independent layers for all channels. Traffic has a base period of 24 hours and harmonic order of 10. 

The base period (base\_T) for each dataset reflects the number of particular time intervals within a day at which the dataset was sampled, while the harmonic order denotes the n-th harmonic derived from base\_T, utilized to determine the cutoff frequency. Note that for national illness (ILI) the base period is number of weeks in a year. 

The primary results are provided in table \ref{tab:gigatable}, where we also include exchange rate and national illness (ILI). We include comparable models and repeat of the last value into the forecast horizon as a baseline. ARIMA (Auto-ARIMA) was only tested on 100 samples and on forecast horizon of 96 due to the computational complexity and time required.

Our reproduced scores have some minor discrepencies compared to the reported, which might be caused by the slightly different data handling or different seeds. We also notice from the results provided in the log files in the original FITS repo, that the numbers are not rounded but floored to the significant digit (e.g., 0.34857 is floored to 0.348), while ours have been rounded often causing slight differences in the reported scores. A particularly notable difference is in the ETTh2 dataset at the 192 forecast horizon, where there is a discrepancy of 0.01, for which we lack a clear explanation. None of these differences are however large enough to change the rankings compared to other models.

\include{gigatable}

\subsection{Results with variants of FITS}

We present our findings from our research on enhancing the FITS model. In Table \ref{tab:Fits_variants_ETTm1}, we examined the effects of different lookback window lengths [96, 192, 336, 720] on various models using the ETTm1 dataset. The real-valued deep FITS consistently underperformed, while the other models showed similar results. The bypass layer model performed slightly better with shorter lookback windows [96, 192], whereas deep FITS showed marginal improvements with longer lookback windows [336, 720].

\begin{table*}[ht]
\centering
\makebox[\textwidth][c]{
\resizebox{\textwidth}{!}{
\begin{tabular}{@{}c|cc|cc|cc|cc@{}}
\toprule
& \multicolumn{8}{c}{ETTm1} \\
\midrule
Lookback & \multicolumn{2}{c|}{96} & \multicolumn{2}{c|}{192} & \multicolumn{2}{c|}{336} & \multicolumn{2}{c}{720} \\ \cmidrule{1-9}
Model & MSE & MAE & MSE & MAE & MSE & MAE & MSE & MAE \\ \midrule
real\_deep\_FITS & 0.5771 & 0.5104 & 0.5311 & 0.4899 & 0.4727 & 0.4602 & 0.4261 & 0.4287 \\
FITS\_bypass\_layer & 0.4861 & 0.4485 & 0.4405 & 0.4229 & 0.4275 & 0.4165 & 0.4155 & 0.4113 \\
deep\_FITS (ModReLU) & 0.4871 & 0.4514 & 0.4416 & 0.4263 & 0.4272 & 0.4199 & 0.4138 & 0.4144 \\
deep\_FITS\_after\_upscaler & 0.4866 & 0.4488 & 0.4407 & 0.4227 & 0.4274 & 0.4161 & 0.4138 & 0.4113 \\ \bottomrule
\end{tabular}
}}
\caption{Long-term forecasting results in MSE and MAE with different lookback lengths on ETTm1 with forecast horizon of 720 steps.}
\label{tab:Fits_variants_ETTm1}
\end{table*}

Table \ref{table:FITS_variants} presents the models' performance across all ETT datasets. The models were configured with harmonic order 6 and, where applicable, 2 hidden layers with 128 neurons per layer. Despite the significant increase in parameters compared to the original FITS, none of the variants demonstrated substantial improvements.

\begin{table*}[ht]
\centering
\makebox[\textwidth][c]
{
\resizebox{\textwidth}{!}{
\begin{tabular}{@{}c|cccc|cccc|cccc|cccc@{}}
\toprule
Dataset & \multicolumn{4}{c|}{ETTh1} & \multicolumn{4}{c|}{ETTh2} & \multicolumn{4}{c|}{ETTm1} & \multicolumn{4}{c}{ETTm2} \\ \midrule
Horizon & 96 & 192 & 336 & 720 & 96 & 192 & 336 & 720 & 96 & 192 & 336 & 720 & 96 & 192 & 336 & 720 \\ \midrule

FITS                & 0.3804 & 0.4173 & 0.4426 & \textbf{0.4363} & 0.2707 & 0.3418 & 0.3588 & \textbf{0.3796} & 0.3076 & 0.3372 & 0.3655 & 0.4148 & \textbf{0.1617} & 0.2157 & 0.2687 & \textbf{0.3494} \\

deep\_FITS (ModReLU) & \textbf{0.3692} & \textbf{0.4050} & \textbf{0.4333} & 0.4440 & 0.2687 & 0.3414 & 0.3593 & 0.3830 & 0.3100 & \textbf{0.3371} & \textbf{0.3643} & \textbf{0.4137} & 0.1722 & \textbf{0.2155} & 0.2682 & 0.3501 \\

real\_deep\_FITS    & 0.3792 & 0.4158 & 0.4494 & 0.5164 & 0.2877 & 0.3619 & 0.3783 & 0.4251 & 0.3117 & 0.3437 & 0.3672 & 0.4248 & 0.1719 & 0.2299 & 0.2825 & 0.3586 \\

FITS\_bypass\_layer & 0.3761 & 0.4153 & 0.4462 & 0.4635 & \textbf{0.2685} & \textbf{0.3398} & \textbf{0.3573} & 0.3826 & \textbf{0.3062} & 0.3359 & 0.3655 & 0.4155 & 0.1617 & \textbf{0.2155} & \textbf{0.2674} & 0.3495 \\

deep\_FITS\_after\_upscaler & 0.3804 & 0.4178 & 0.4427 & 0.4365 & 0.2703 & 0.3415 & 0.3586 & 0.3797 & 0.3079 & 0.3373 & 0.3657 & 0.4149 & 0.1618 & 0.2161 & 0.2680 & \textbf{0.3494} \\
\bottomrule
\multicolumn{6}{l}{} \\
\end{tabular}
}}
\caption{Long-term forecasting results on the ETT datasets in MSE with a look-back window of 720 and different forecast horizons. real\_deep\_FITS, deep\_FITS and deep\_FITS\_after\_upscaler use 3 layers and 128 neurons per layer.}
\label{table:FITS_variants}
\end{table*}

To evaluate whether deep FITS with ModReLU actually performed statistically better than FITS on ETTh1 at a forecast horizon of 96, as indicated by the results in Table \ref{table:FITS_variants}, we trained each model 10 times with different random seeds to calculate the mean and standard deviation of the MSE. We then conducted a paired t-test to assess the statistical significance of the difference between the two models' performance.

\begin{table}[h!]
\centering
\small
\begin{tabular}{lccc}
\hline
\textbf{Model} & \textbf{Mean MSE} & \textbf{Std Dev} & \textbf{p-value} \\
\hline
FITS & 0.3806 & 0.0004 & \multirow{2}{*}{$3.42 \times 10^{-13}$} \\
deep\_FITS (ModReLU) & 0.3692 & 0.0002 & \\
\hline
\end{tabular}
\vspace{.2cm}
\caption{Comparison of FITS and deep\_FITS models on ETTh1 dataset (lookback: 720, horizon: 96).}
\label{table:mse_comparison}
\end{table}

The results in table \ref{table:mse_comparison} show that deep FITS with ModReLU achieved a lower mean MSE (0.3692) compared to FITS (0.3806), with a smaller standard deviation as well. A paired t-test yielded a p-value of $3.42 \times 10^{-13}$, indicating a statistically significant difference between the two models' performance.

We also present an analysis of the performance of three Deep FITS variants: $\mathbb{C}\text{ReLU}$, ModReLU, and Deep FITS After Upscaler, on the ETTh1 dataset in table \ref{table:mse_results_deep_fits_comparison}. Our investigation explores the effect of varying the number of layers and hidden neurons, utilizing a consistent lookback and forecast horizon of 720 time steps. The baseline '0 layers' configuration represents the original FITS model, with its linear layer embedded within a sequential module for consistency.

\begin{table*}[ht]
\centering
\makebox[\textwidth][c]
{
\resizebox{\textwidth}{!}{
\begin{tabular}{@{}c|cccc|cccc|cccc@{}}
\toprule
Layers & \multicolumn{4}{c|}{$\mathbb{C}\text{ReLU}$ Deep FITS} & \multicolumn{4}{c|}{ModReLU Deep FITS} & \multicolumn{4}{c}{Deep FITS After Upscaler} \\
\midrule
Hidden Units & 64 & 128 & 256 & 512 & 64 & 128 & 256 & 512 & 64 & 128 & 256 & 512 \\
\midrule
0 & \textbf{0.4363} & \textbf{0.4363} & \textbf{0.4363} & \textbf{0.4363} & \textbf{0.4363} & \textbf{0.4363} & \textbf{0.4363} & \textbf{0.4363} & 0.4363 & 0.4363 & \textbf{0.4363} & 0.4363 \\
1 & 0.4929 & 0.4870 & 0.5198 & 0.5315 & 0.4379 & 0.4498 & 0.4413 & 0.4494 & 0.4367 & 0.4366 & 0.4364 & 0.4364 \\
2 & 0.4759 & 0.4816 & 0.5010 & 0.5214 & 0.4526 & 0.4440 & 0.4478 & 0.4492 & \textbf{0.4362} & 0.4363 & 0.4366 & 0.4364 \\
3 & 0.4797 & 0.4522 & 0.4626 & 0.4857 & 0.4607 & 0.4515 & 0.4506 & 0.4470 & 0.4365 & \textbf{0.4360} & 0.4364 & \textbf{0.4357} \\
4 & 0.4972 & 0.4561 & 0.4565 & 0.4822 & 0.4736 & 0.4588 & 0.4530 & 0.4477 & 0.4366 & 0.4361 & 0.4365 & 0.4361 \\
\bottomrule
\end{tabular}
}}
\caption{MSE results for $\mathbb{C}\text{ReLU}$ Deep FITS, ModReLU Deep FITS, and Deep FITS after upscaler on ETTh1 (forecast horizon: 720, lookback window: 720, harmonic order: 6).}
\label{table:mse_results_deep_fits_comparison}
\end{table*}

Our findings show that the Deep FITS After Upscaler variant achieves marginally superior performance, resulting in the lowest MSE when configured with 3 layers and 512 hidden units per layer, and with the lowest deterioration of MSE overall. $\mathbb{C}\text{ReLU}$ consistently performs the worst among the three variants across different configurations. For instance, with 1 layer and 512 hidden units, $\mathbb{C}\text{ReLU}$ records an MSE of 0.5315, significantly higher than the other variants. ModReLU shows slightly better performance than $\mathbb{C}\text{ReLU}$ but still lags behind the Deep FITS After Upscaler, and the unmodified FITS.

\include{minitable}

%% file: gigatable.tex
\begin{table*}[ht]
\centering
\makebox[\textwidth][c]{%
\resizebox{1.05\textwidth}{!}{%
\begin{tabular}{c|c|c|cc|c|cccccccc|cc|c}
\hline
\multicolumn{2}{c|}{Methods}&{IMP.}&\multicolumn{2}{c|}{FITS (ours)}&\multicolumn{1}{c|}{FITS*}&\multicolumn{2}{c|}{DLinear*}&\multicolumn{2}{c|}{NLinear*}&\multicolumn{2}{c|}{PatchTST/64 } &\multicolumn{2}{c|}{PatchTST/42}&\multicolumn{2}{c}{\emph{Repeat}*}&\multicolumn{1}{|c}{ARIMA}\\
\hline
\multicolumn{2}{c|}{Metric} & MSE & MSE&MAE & MSE & MSE&MAE & MSE&MAE & MSE&MAE & MSE&MAE & MSE&MAE & MSE\\
\hline
\multirow{4}{*}{\rotatebox{90}{Weather}}
& 96 & 4.20\%
& \textbf{0.143} & \textbf{0.195}
& \textbf{0.143}
& 0.176 & 0.237 & 0.182 & 0.232 %
& \underline{0.149} & \underline{0.198} & 0.152 & 0.199 %
& 0.259 & 0.254 %
& 0.239 \\ %
& 192 & 4.30\%
& \textbf{0.186} & \textbf{0.236}
& \textbf{0.186}
& 0.220 & 0.282 & 0.225 & 0.269 %
& \underline{0.194} & \underline{0.241} & 0.197 & 0.243 %
& 0.309 & 0.292 \\ %
& 336 & 3.81\%
& \textbf{0.236} & \textbf{0.277}
& \textbf{0.236}
& 0.265 & 0.319 & 0.271 & 0.301 %
& \underline{0.245} & \underline{0.282} & 0.249 & 0.283 %
& 0.377 & 0.338 \\ %
& 720 & 1.95\%
& 0.308 & \textbf{0.330}
& \textbf{0.307}
& 0.323 & 0.362 & 0.338 & 0.348 %
& \underline{0.314} & \underline{0.334} & 0.320 & 0.335 %
& 0.465 & 0.394 \\ %
\hline
\multirow{4}{*}{\rotatebox{90}{Traffic}}
& 96 & -6.74\%
& 0.386 & 0.266
& 0.385
& 0.410 & 0.282 & 0.410 & 0.279 %
& \textbf{0.360} & \textbf{0.249} & \underline{0.367} & \underline{0.251} %
& 2.723 & 1.079 %
& 1.448 \\ %
& 192 & -4.77\%
& \underline{0.398} & 0.273
& 0.397
& 0.423 & 0.287 & 0.423 & 0.284 %
& \textbf{0.379} & \textbf{0.256} & \underline{0.385} & \underline{0.259} %
& 2.756 & 1.087 \\ %
& 336 & -5.08\%
& 0.413 & 0.277
& 0.410
& 0.436 & 0.296 & 0.435 & 0.290 %
& \textbf{0.392} & \textbf{0.264} & \underline{0.398} & \underline{0.265} %
& 2.791 & 1.095 \\ %
& 720 & -3.79\%
& 0.449 & 0.297
& 0.448
& 0.466 & 0.315 & 0.464 & 0.307 %
& \textbf{0.432} & \textbf{0.286} & \underline{0.434} & \underline{0.287} %
& 2.811 & 1.097 \\ %
\hline
\multirow{4}{*}{\rotatebox{90}{Electricity}}
& 96 & -3.73\%  
& 0.134 & 0.229 
& 0.134
& 0.140 & 0.237 & 0.141 & 0.237 %
& \textbf{0.129} & \textbf{0.222} & \underline{0.130} & \textbf{0.222} %
& 1.588 & 0.946 \\ %
& 192 & -3.29\%
& 0.152 & 0.243
& 0.149
& 0.153 & 0.249 & 0.154 & 0.248 %
& \textbf{0.147} & \textbf{0.240} & \underline{0.148} & \textbf{0.240} %
& 1.595 & 0.950 \\ %
& 336 & -1.21\%
& \underline{0.165} & \textbf{0.256}
& \underline{0.165}
& 0.169 & 0.267 & 0.171 & 0.265 %
& \textbf{0.163} & \underline{0.259} & 0.167 & 0.261 %
& 1.617 & 0.961 \\ %
& 720 & -3.43\%
& 0.204 & 0.292
& 0.203
& 0.203 & 0.301 & 0.210 & 0.297 %
& \textbf{0.197} & \textbf{0.290} & \underline{0.202} & \underline{0.291} %
& 1.647 & 0.975 \\ %
\hline
\multirow{4}{*}{\rotatebox{90}{ILI}}
& 24 & -47.18\%
& 2.497 & 0.988
& -
& 2.215 & 1.081 & 1.683 & 0.858 %
& \textbf{1.319} & \textbf{0.754} & \underline{1.522} & \underline{0.814} %
& 6.587 & 1.701 %
& 9.249 %
\\ 
& 36 & -40.59\%
& 2.407 & 0.995
& -
& 1.963 & 0.963 & 1.703 & \underline{0.859} %
& \underline{1.579} & 0.870 & \textbf{1.430} & \textbf{0.834} %
& 7.130 & 1.884 \\ %
& 48 & -37.99\%
& 2.472 & 1.038
& -
& 2.130 & 1.024 & 1.719 & 0.884 %
& \textbf{1.553} & \textbf{0.815} & \underline{1.673} & 0.854 %
& 6.575 & 1.798 \\ %
& 60 & -39.28\%
& 2.421 & 1.033
& -
& 2.368 & 1.096 & 1.819 & 0.917 %
& \textbf{1.470} & \textbf{0.788} & \underline{1.529} & \underline{0.862} %
& 5.893 & 1.677 \\ %
\hline
\multirow{4}{*}{\rotatebox{90}{ETTh1}}
& 96 & 5.63\%
& 0.373 & 0.395
& \underline{0.372}
& 0.375 & 0.399 & 0.374 & \textbf{0.394} %
& \textbf{0.370} & 0.400 & 0.375 & 0.399 %
& 1.295 &0.713 %
& 0.839 \\ %
& 192 & 1.97\%
& 0.407 & \textbf{0.415}
& \textbf{0.404}
& \underline{0.405} & 0.416 & 0.408 & \textbf{0.415} %
& 0.413 & 0.429 & 0.414 & 0.421 %
& 1.325 & 0.733 \\ %
& 336 & -0.23\%
& 0.429 & 0.429
& \underline{0.427}
& 0.439 & 0.443 & 0.429 & \textbf{0.427} %
& \textbf{0.422} & 0.440 & 0.431 & 0.436 %
& 1.323 &0.744 \\ %
& 720 & 5.84\%
& \underline{0.428} & \textbf{0.448}
& \textbf{0.424}
& 0.472 & 0.490 & 0.440 & \underline{0.453} %
& 0.447 & 0.468 & 0.449 & 0.466 %
& 1.339 &0.756 \\ %
\hline
\multirow{4}{*}{\rotatebox{90}{ETTh2}}
& 96 & 1.48\%
& \textbf{0.270} & \textbf{0.322}
& \underline{0.271}
& 0.289 & 0.353 & 0.277 & 0.338 %
& 0.274 & 0.337 & 0.274 & \underline{0.336} %
& 0.432 & 0.422 %
& 0.444 \\ %
& 192 & -21.35\%
& 0.341 & \textbf{0.365}
& \textbf{0.331}
& 0.383 & 0.418 & 0.344 & 0.381 %
& 0.341 & 0.382 & \underline{0.339} & \underline{0.379} %
& 0.534 & 0.473 \\ %
& 336 & -8.36\%
& 0.359 & 0.439
& 0.354
& 0.448 & 0.465 & 0.357 & 0.400 %
& \textbf{0.329} & \underline{0.384} & \underline{0.331} & \textbf{0.380} %
& 0.591 & 0.508 \\ %
& 720 & -0.26\%
& 0.380 & 0.457
& \textbf{0.377}
& 0.605 & 0.551 & 0.394 & \underline{0.436} %
& \underline{0.379} & \textbf{0.422} & \underline{0.379} & \textbf{0.422} %
& 0.588 & 0.517 \\ %
\hline
\multirow{4}{*}{\rotatebox{90}{ETTm1}}
& 96  & -5.84\%
& 0.308 & 0.345
& 0.303
& 0.299 & \underline{0.343} & 0.306 & 0.348 %
& \underline{0.293} & 0.346 & \textbf{0.290} & \textbf{0.342} %
& 1.214 & 0.665 %
& 0.910 \\ %
& 192 & -1.48\%
& 0.337 & \textbf{0.363}
& 0.337
& 0.335 & \underline{0.365} & 0.349 & 0.375 %
& \underline{0.333} & 0.370 & \textbf{0.332} & 0.369 %
& 1.261 & 0.690 \\ %
& 336 & 0.00\%
& \textbf{0.366} & \textbf{0.381}
& \textbf{0.366}
& \underline{0.369} & \underline{0.386} & 0.375 & 0.388 %
& \underline{0.369} & 0.392 & \textbf{0.366} & 0.392 %
& 1.283 & 0.707 \\ %
& 720 & 1.20\%
& \textbf{0.415} & \textbf{0.408}
& \textbf{0.415}
& 0.425 & 0.421 & 0.433 & 0.422 %
& \underline{0.416} & \underline{0.420} & 0.420 & 0.424 %
& 1.319 & 0.729 \\ %
\hline
\multirow{4}{*}{\rotatebox{90}{ETTm2}} 
& 96  & 1.85\%
& \textbf{0.162} & \textbf{0.242}
& \textbf{0.162}
& 0.167 & 0.262 & 0.167 & \underline{0.255} %
& 0.166 & 0.260 & \underline{0.165} & \underline{0.255} %
& 0.266 & 0.328 %
& 0.224 \\ %
& 192 & 1.85\%
& \textbf{0.216} & \textbf{0.276}
& \textbf{0.216}
& 0.224 & 0.303 & 0.221 & 0.293 %
& 0.223 & 0.303 & \underline{0.220} & \underline{0.292} %
& 0.340 & 0.371 \\ %
& 336 & 2.24\%
& \textbf{0.268} & \textbf{0.309}
& \textbf{0.268}
& 0.281 & 0.342 & \underline{0.274} & \underline{0.327} %
& \underline{0.274} & 0.342 & 0.278 & 0.329 %
& 0.412 & 0.410 \\ %
& 720 & 3.72\%
& \underline{0.349} & \textbf{0.360}
& \textbf{0.348}
& 0.397 & 0.421 & 0.368 & \underline{0.384} %
& 0.362 & 0.421 & 0.367 & 0.385 %
& 0.521 &0.465 \\ %
\hline
\multirow{4}{*}{\rotatebox{90}{Exchange}}
& 96 & -14.94\%
& 0.087 & 0.204
& -
& \underline{0.081} & \underline{0.203} & 0.089 & 0.208 %
& - & - & - & - %
& \underline{0.081} & \textbf{0.196} %
& \textbf{0.074} \\ 
& 192 & -12.78\%
& 0.180 & 0.295
& -
& \textbf{0.157} & \underline{0.293} & 0.180 & 0.300 %
& - & - & - & - %
& \underline{0.167} & \textbf{0.289} \\ %
& 336 & -8.41\%
& 0.333 & \underline{0.405} 
& -
& \textbf{0.305} & 0.414 & \underline{0.331} & 0.415 %
& - & - & - & - %
& \textbf{0.305} & \textbf{0.396} \\ %
& 720 & -32.39\%
& 0.951 & 0.713
& -
& \textbf{0.643} & \textbf{0.601} & 1.033 & 0.780 %
& - & - & - & - %
& \underline{0.823} & \underline{0.681} \\ %
\hline
\end{tabular}
}}
\begin{tablenotes}
    \scriptsize
    {
    \item *Results borrowed from \cite{xu2024fits}, \cite{DLinear}, and \cite{PatchTST}.
    }
\end{tablenotes}
\caption{Multivariate long-term forecasting errors in terms of MSE and MAE, the lower the better. Note that the ILI dataset has prediction length $T \in \{24,36,48,60\}$ whereas the others have prediction length $T \in \{96,192,336,720\}$. \emph{Repeat} repeats the last value in the look-back window and serves as our baseline. The best results are highlighted in bold and the second best results are highlighted with underline. IMP denotes the relative improvement or downgrade from FITS (ours) and the second best or best model.}
\label{tab:gigatable}
\end{table*}

%% file: minitable.tex
\subsection{Price datasets}

Price datasets with various error metrics across 381, 369, 351, and 303 test samples for horizons 96, 192, 336, and 720 respectively with a set lookback-window of 336. The standard FITS models have their cutoff frequency noted in parentheses (10 \& 100), whereas the hybrid FITS models have a cutoff frequency of 49 for DLinear + FITS and 168 for FITS + DLinear.

Mean Squared Error (MSE):
$$\text{MSE} = \frac{1}{n} \sum_{i=1}^{n} (y_i - \hat{y}_i)^2$$
Mean Absolute Error (MAE):
$$\text{MAE} = \frac{1}{n} \sum_{i=1}^{n} |y_i - \hat{y}_i|$$
Squared Error  (SE):
$$\text{SE} = (y_{T} - \hat{y}_{T})^2$$

Relative Root Mean Squared Error (RRMSE):
$$\text{RRMSE} = \frac{\sqrt{\text{MSE}}}{\frac{1}{n} \sum_{i=1}^{n} |y_i|}$$

Where $n$ is the length of the forecasted horizon (96, 192, 336, and 720), $y_i$ represents the true value, $\hat{y}_i$ represents the predicted value, and $\hat{y}_T$ \& $y_T$ represents the last value in lookback window.

The training and test logs for these results can be found here: 

\href{https://github.com/thorhojhus/ssl_fts/tree/DLinear/results}{https://github.com/thorhojhus/ssl\_fts/tree/DLinear/results}

\begin{table*}[ht!]
\centering
\makebox[\textwidth][c]{
\resizebox{1\textwidth}{!}{
\begin{tabular}{@{}c|c|cccc|cccc|cccc|c@{}}
\toprule
Dataset & & \multicolumn{4}{c|}{Exchange Rate} & \multicolumn{4}{c|}{General Dynamics $\underline{GD}$} & \multicolumn{4}{c|}{Marathon Oil Corp $\underline{MRO}$} & Best \\ 
\midrule
Model & Horizon & MSE & MAE & SE & RRMSE & MSE & MAE & SE & RRMSE & MSE & MAE & SE & RRMSE & 48 \\ 
\midrule
\multirow{4}{4em}{Repeat}
& 96 & 0.081 & \textbf{0.196} & 0.164 & 19.53\% & \textbf{1.196} & \textbf{0.763} & \textbf{2.283} & \textbf{10.98\%} & 4.465 & \textbf{1.246} & 6.730 & 36.19\% & \multirow{4}{*}{17/48} \\
& 192 & 0.167 & \textbf{0.289} & 0.342 & 28.41\% & \textbf{2.250} & \textbf{1.077} & 4.923 & \textbf{15.07\%} & 6.562 & \textbf{1.653} & 10.663 & 44.26\% \\
& 336 & 0.306 & \underline{0.397} & 0.626 & 38.85\% & \textbf{3.899} & \textbf{1.444} & \underline{8.558} & \textbf{19.78\%} & 10.411 & \underline{2.190} & 18.627 & 56.68\% \\
& 720 & \textbf{0.813} & \textbf{0.677} & 1.781 & \textbf{63.52\%} & 9.859 & 2.366 & 21.451 & 31.22\% & 20.380 & 3.233 & 33.096 & 85.56\% \\
\midrule
\multirow{4}{4em}{DLinear}
& 96 & \underline{0.081} & 0.203 & 0.184 & \underline{19.51\%} & 1.659 & 0.948 & \underline{2.356} & 12.93\% & \underline{3.826} & \underline{1.270} & \textbf{5.770} & \underline{33.50\%} & \multirow{4}{*}{9/48} \\
& 192 & \textbf{0.157} & \underline{0.293} & 0.321 & \textbf{27.51\%} & 2.430 & 1.153 & 4.825 & 15.66\% & 6.203 & 1.765 & 10.376 & 43.03\% \\
& 336 & \underline{0.299} & 0.417 & \textbf{0.492} & \underline{38.44\%} & \underline{4.189} & \underline{1.551} & \textbf{8.089} & \underline{20.50\%} & \textbf{9.121} & \textbf{2.178} & \textbf{15.831} & \textbf{53.05\%} \\
& 720 & 1.040 & 0.767 & 1.804 & 71.86\% & 8.923 & 2.341 & \underline{15.157} & 29.70\% & 16.783 & \underline{2.968} & 19.187 & 77.64\% \\
\midrule
\multirow{4}{4em}{DLinear + FITS}
& 96 & 0.089 & 0.214 & \textbf{0.152} & 20.48\% & 1.660 & 0.933 & 3.268 & 12.93\% & 3.838 & 1.276 & 5.890 & 33.55\% & \multirow{4}{*}{6/48} \\
& 192 & 0.173 & 0.308 & \textbf{0.295} & 28.93\% & 2.382 & \underline{1.128} & \textbf{4.686} & 15.51\% & \textbf{5.824} & 1.696 & \underline{9.691} & \textbf{41.70\%} \\
& 336 & 0.300 & 0.416 & 0.702 & 38.51\% & 5.116 & 1.707 & 8.590 & 22.66\% & \underline{9.322} & 2.203 & 17.687 & \underline{53.63\%} \\
& 720 & 0.962 & 0.736 & \textbf{1.485} & 69.12\% & \underline{8.889} & \underline{2.323} & 15.236 & \underline{29.65\%} & \underline{16.773} & 3.021 & \underline{17.534} & \underline{77.62\%} \\
\midrule
\multirow{4}{4em}{FITS + DLinear}
& 96 & \textbf{0.079} & \underline{0.200} & \underline{0.156} & \textbf{19.31\%} & 1.775 & 0.946 & 3.212 & 13.37\% & \textbf{3.823} & 1.272 & \underline{5.774} & \textbf{33.48\%} & \multirow{4}{*}{16/48} \\
& 192 & \underline{0.157} & 0.294 & \underline{0.302} & \underline{27.52\%} & \underline{2.374} & 1.136 & \underline{4.733} & \underline{15.48\%} & \underline{5.852} & \underline{1.691} & \textbf{9.671} & \underline{41.80\%} \\
& 336 & \textbf{0.276} & \textbf{0.396} & \underline{0.544} & \textbf{36.94\%} & 5.887 & 1.851 & 10.428 & 24.30\% & 9.376 & 2.218 & \underline{16.519} & 53.78\% \\
& 720 & 0.961 & 0.736 & \underline{1.728} & 69.06\% & \textbf{8.814} & \textbf{2.301} & \textbf{15.031} & \textbf{29.52\%} & \textbf{15.462} & \textbf{2.882} & \textbf{17.467} & \textbf{74.52\%} \\
\midrule
\multirow{4}{4em}{FITS}
& 96 & 0.089 & 0.212 & 0.192 & 20.46\% & 1.539 & 0.899 & 2.634 & 12.45\% & 3.975 & 1.304 & 6.195 & 34.14\% & \multirow{4}{*}{0/48} \\
& 192 & 0.184 & 0.310 & 0.378 & 29.81\% & 2.691 & 1.208 & 5.281 & 16.48\% & 6.062 & 1.716 & 10.265 & 42.54\% \\
& 336 & 0.331 & 0.421 & 0.717 & 40.40\% & 4.795 & 1.635 & 9.223 & 21.93\% & 9.977 & 2.259 & 17.624 & 55.48\% \\
& 720 & \underline{0.913} & \underline{0.732} & 1.878 & \underline{67.32\%} & 11.150 & 2.599 & 25.127 & 33.21\% & 18.319 & 3.179 & 26.543 & 81.12\% \\
\midrule
\multirow{4}{4em}{FITS100}
& 96 & 0.088 & 0.212 & 0.196 & 20.43\% & \underline{1.531} & \underline{0.895} & 2.509 & \underline{12.42\%} & 3.963 & 1.302 & 6.695 & 34.09\% & \multirow{4}{*}{0/48} \\
& 192 & 0.184 & 0.310 & 0.371 & 29.79\% & 2.692 & 1.208 & 5.098 & 16.49\% & 6.053 & 1.716 & 10.786 & 42.51\% \\
& 336 & 0.331 & 0.421 & 0.695 & 40.44\% & 4.787 & 1.633 & 8.921 & 21.92\% & 9.973 & 2.260 & 17.936 & 55.47\% \\
& 720 & 0.920 & 0.735 & 1.937 & 67.59\% & 11.161 & 2.602 & 25.952 & 33.22\% & 18.325 & 3.181 & 27.608 & 81.13\% \\
\midrule
\multirow{4}{4em}{FITS10}
& 96 & 0.101 & 0.228 & 0.211 & 21.78\% & 1.620 & 0.929 & 2.670 & 12.78\% & 4.101 & 1.347 & 6.437 & 34.68\% & \multirow{4}{*}{0/48} \\
& 192 & 0.199 & 0.325 & 0.408 & 30.99\% & 2.787 & 1.236 & 5.564 & 16.77\% & 6.168 & 1.747 & 10.849 & 42.91\% \\
& 336 & 0.352 & 0.436 & 0.750 & 41.69\% & 4.900 & 1.662 & 10.084 & 22.17\% & 10.095 & 2.287 & 17.827 & 55.81\% \\
& 720 & 0.993 & 0.764 & 2.144 & 70.22\% & 11.310 & 2.630 & 21.824 & 33.44\% & 18.343 & 3.192 & 25.342 & 81.17\% \\
\midrule
\end{tabular}
}}
\caption{Multivariate input on multivariate columns}
\label{tab:multivariate_multivariate}
\end{table*}

\begin{table*}[ht!]
\centering
\makebox[\textwidth][c]{%
\resizebox{1\textwidth}{!}{%
\begin{tabular}{@{}c|c|cccc|cccc|cccc|c@{}}
\toprule
Dataset & & \multicolumn{4}{c|}{Exchange Rate} & \multicolumn{4}{c|}{General Dynamics $\underline{GD}$} & \multicolumn{4}{c|}{Marathon Oil Corp $\underline{MRO}$} & Best \\ 
\midrule
Model & Horizon & MSE & MAE & SE & RRMSE & MSE & MAE & SE & RRMSE & MSE & MAE & SE & RRMSE & 48 \\ 
\midrule
\multirow{4}{4em}{Repeat}
& 96 & \textbf{0.088} & \textbf{0.221} & \textbf{0.179} & \textbf{17.68\%} & \textbf{2.180} & \textbf{1.045} & 4.423 & \textbf{9.57\%} & \textbf{2.870} & \textbf{1.261} & \underline{5.347} & \textbf{20.17\%} & \multirow{4}{*}{26/48} \\
& 192 & \textbf{0.188} & \textbf{0.333} & 0.386 & \textbf{26.03\%} & \textbf{4.301} & \textbf{1.531} & 9.898 & \textbf{13.48\%} & \textbf{5.362} & \textbf{1.761} & \textbf{10.657} & \textbf{27.86\%} \\
& 336 & 0.370 & 0.468 & 0.809 & 36.49\% & \textbf{7.579} & \textbf{2.093} & 17.440 & \textbf{17.87\%} & \textbf{10.261} & \textbf{2.421} & 21.428 & \textbf{39.26\%} \\
& 720 & 1.002 & \underline{0.765} & 1.831 & 58.32\% & 20.163 & 3.575 & 46.714 & 29.08\% & 23.185 & 3.680 & 39.829 & 63.92\% \\
\midrule
\multirow{4}{4em}{DLinear}
& 96 & \underline{0.099} & \underline{0.242} & \underline{0.192} & \underline{18.82\%} & 2.459 & 1.170 & \underline{4.054} & 10.16\% & \underline{2.921} & 1.310 & \textbf{5.254} & \underline{20.35\%} & \multirow{4}{*}{8/48} \\
& 192 & 0.197 & 0.355 & 0.404 & 26.65\% & \underline{4.335} & \underline{1.542} & \underline{9.311} & \underline{13.53\%} & 5.714 & 1.828 & 12.139 & 28.76\% \\
& 336 & \textbf{0.317} & \textbf{0.458} & \underline{0.657} & \textbf{33.77\%} & \underline{7.986} & \underline{2.160} & \textbf{15.729} & \underline{18.35\%} & 10.830 & 2.488 & 24.803 & 40.33\% \\
& 720 & \underline{0.981} & 0.783 & \textbf{0.928} & \underline{57.69\%} & \textbf{17.167} & 3.397 & \underline{30.434} & \textbf{26.83\%} & \underline{20.396} & \underline{3.536} & 36.740 & \underline{59.95\%} \\
\midrule
\multirow{4}{4em}{DLinear + FITS}
& 96 & 0.107 & 0.254 & 0.201 & 19.52\% & 2.840 & 1.247 & \textbf{3.979} & 10.92\% & 2.997 & 1.321 & 5.574 & 20.62\% & \multirow{4}{*}{6/48} \\
& 192 & 0.255 & 0.397 & \underline{0.360} & 30.32\% & 4.944 & 1.709 & 9.626 & 14.45\% & 5.714 & 1.825 & 12.242 & 28.76\% \\
& 336 & 0.404 & 0.506 & 0.759 & 38.14\% & 9.204 & 2.395 & 17.318 & 19.70\% & 10.638 & 2.504 & \textbf{19.348} & 39.97\% \\
& 720 & 1.057 & 0.803 & 1.072 & 59.89\% & \underline{17.583} & \textbf{3.280} & 37.847 & \underline{27.15\%} & \textbf{19.986} & \textbf{3.499} & \underline{34.045} & \textbf{59.35\%} \\
\midrule
\multirow{4}{4em}{FITS + DLinear}
& 96 & 0.106 & 0.249 & 0.210 & 19.42\% & 2.604 & 1.161 & 5.208 & 10.46\% & 3.024 & \underline{1.309} & 5.617 & 20.71\% & \multirow{4}{*}{7/48} \\
& 192 & \underline{0.197} & \underline{0.351} & \textbf{0.333} & \underline{26.62\%} & 4.419 & 1.571 & \textbf{9.304} & 13.66\% & \underline{5.562} & \underline{1.819} & 11.092 & \underline{28.37\%} \\
& 336 & \underline{0.337} & \underline{0.460} & \textbf{0.575} & \underline{34.83\%} & 10.053 & 2.544 & 17.222 & 20.58\% & \underline{10.474} & \underline{2.466} & 23.334 & \underline{39.66\%} \\
& 720 & \textbf{0.873} & \textbf{0.732} & \underline{1.043} & \textbf{54.42\%} & 17.726 & \underline{3.311} & \textbf{29.015} & 27.26\% & 21.393 & 3.610 & 40.060 & 61.40\% \\
\midrule
\multirow{4}{4em}{FITS}
& 96 & 0.120 & 0.277 & 0.264 & 20.64\% & \underline{2.392} & \underline{1.125} & 4.278 & \underline{10.03\%} & 3.163 & 1.385 & 6.150 & 21.18\% & \multirow{4}{*}{0/48} \\
& 192 & 0.243 & 0.393 & 0.472 & 29.58\% & 4.723 & 1.637 & 9.919 & 14.12\% & 5.701 & 1.880 & \underline{11.051} & 28.72\% \\
& 336 & 0.405 & 0.485 & 0.851 & 38.17\% & 8.658 & 2.267 & \underline{16.092} & 19.10\% & 10.886 & 2.541 & 20.806 & 40.43\% \\
& 720 & 1.066 & 0.808 & 2.415 & 60.14\% & 23.621 & 3.900 & 59.070 & 31.47\% & 22.865 & 3.720 & 36.404 & 63.48\% \\
\midrule
\multirow{4}{4em}{FITS100}
& 96 & 0.119 & 0.273 & 0.247 & 20.54\% & 2.459 & 1.153 & 4.802 & 10.17\% & 3.174 & 1.391 & 6.469 & 21.22\% & \multirow{4}{*}{0/48} \\
& 192 & 0.239 & 0.389 & 0.443 & 29.36\% & 4.738 & 1.642 & 10.412 & 14.15\% & 5.701 & 1.881 & 11.448 & 28.73\% \\
& 336 & 0.404 & 0.487 & 0.817 & 38.13\% & 8.704 & 2.278 & 17.656 & 19.15\% & 10.892 & 2.545 & 20.770 & 40.44\% \\
& 720 & 1.048 & 0.802 & 2.064 & 59.64\% & 23.623 & 3.919 & 64.024 & 31.47\% & 22.869 & 3.723 & 35.469 & 63.48\% \\
\midrule
\multirow{4}{4em}{FITS10}
& 96 & 0.143 & 0.303 & 0.276 & 22.57\% & 2.704 & 1.239 & 5.193 & 10.66\% & 3.457 & 1.482 & 5.547 & 22.14\% & \multirow{4}{*}{1/48} \\
& 192 & 0.244 & 0.391 & 0.507 & 29.61\% & 5.167 & 1.749 & 12.032 & 14.77\% & 6.021 & 1.957 & 11.073 & 29.52\% \\
& 336 & 0.422 & 0.505 & 0.852 & 38.97\% & 9.303 & 2.394 & 21.951 & 19.80\% & 11.191 & 2.598 & \underline{20.616} & 40.99\% \\
& 720 & 1.065 & 0.822 & 1.739 & 60.12\% & 24.351 & 4.081 & 48.666 & 31.95\% & 22.852 & 3.737 & \textbf{33.523} & 63.46\% \\
\midrule
\end{tabular}
}}
\caption{Multivariate input on univariate column}
\label{tab:multivariate_univariate}
\end{table*}

\begin{table*}[ht!]
\centering
\makebox[\textwidth][c]{%
\resizebox{1\textwidth}{!}{%
\begin{tabular}{@{}c|c|cccc|cccc|cccc|c@{}}
\toprule
Dataset & & \multicolumn{4}{c|}{Exchange Rate} & \multicolumn{4}{c|}{General Dynamics $\underline{GD}$} & \multicolumn{4}{c|}{Marathon Oil Corp $\underline{MRO}$} & Best \\ 
\midrule
Model & Horizon & MSE & MAE & SE & RRMSE & MSE & MAE & SE & RRMSE & MSE & MAE & SE & RRMSE & 48 \\ 
\midrule
\multirow{4}{4em}{Repeat}
& 96 & \textbf{0.088} & \textbf{0.221} & \textbf{0.179} & \textbf{17.68\%} & \textbf{2.180} & \textbf{1.045} & 4.423 & \textbf{9.57\%} & \textbf{2.870} & \textbf{1.261} & \underline{5.347} & \textbf{20.17\%} & \multirow{4}{*}{26/48} \\
& 192 & \textbf{0.188} & \textbf{0.333} & 0.386 & \textbf{26.03\%} & \textbf{4.301} & \textbf{1.531} & 9.898 & \textbf{13.48\%} & \textbf{5.362} & \textbf{1.761} & \textbf{10.657} & \textbf{27.86\%} \\
& 336 & 0.370 & 0.468 & 0.809 & 36.49\% & \textbf{7.579} & \textbf{2.093} & 17.440 & \textbf{17.87\%} & \textbf{10.261} & \textbf{2.421} & 21.428 & \textbf{39.26\%} \\
& 720 & 1.002 & \underline{0.765} & 1.831 & 58.32\% & 20.163 & 3.575 & 46.714 & 29.08\% & 23.185 & 3.680 & 39.829 & 63.92\% \\
\midrule
\multirow{4}{4em}{DLinear}
& 96 & \underline{0.099} & \underline{0.242} & \underline{0.192} & \underline{18.82\%} & 2.459 & 1.170 & \underline{4.054} & 10.16\% & \underline{2.921} & 1.310 & \textbf{5.254} & \underline{20.35\%} & \multirow{4}{*}{8/48} \\
& 192 & 0.197 & 0.355 & 0.404 & 26.65\% & \underline{4.335} & \underline{1.542} & \underline{9.311} & \underline{13.53\%} & 5.714 & 1.828 & 12.139 & 28.76\% \\
& 336 & \textbf{0.317} & \textbf{0.458} & \underline{0.657} & \textbf{33.77\%} & \underline{7.986} & \underline{2.160} & \textbf{15.729} & \underline{18.35\%} & 10.830 & 2.488 & 24.803 & 40.33\% \\
& 720 & \underline{0.981} & 0.783 & \textbf{0.928} & \underline{57.69\%} & \textbf{17.167} & 3.397 & \underline{30.434} & \textbf{26.83\%} & \underline{20.396} & \underline{3.536} & 36.740 & \underline{59.95\%} \\
\midrule
\multirow{4}{4em}{DLinear + FITS}
& 96 & 0.107 & 0.254 & 0.201 & 19.52\% & 2.840 & 1.247 & \textbf{3.979} & 10.92\% & 2.997 & 1.321 & 5.574 & 20.62\% & \multirow{4}{*}{6/48} \\
& 192 & 0.255 & 0.397 & \underline{0.360} & 30.32\% & 4.944 & 1.709 & 9.626 & 14.45\% & 5.714 & 1.825 & 12.242 & 28.76\% \\
& 336 & 0.404 & 0.506 & 0.759 & 38.14\% & 9.204 & 2.395 & 17.318 & 19.70\% & 10.638 & 2.504 & \textbf{19.348} & 39.97\% \\
& 720 & 1.057 & 0.803 & 1.072 & 59.89\% & \underline{17.583} & \textbf{3.280} & 37.847 & \underline{27.15\%} & \textbf{19.986} & \textbf{3.499} & \underline{34.045} & \textbf{59.35\%} \\
\midrule
\multirow{4}{4em}{FITS + DLinear}
& 96 & 0.106 & 0.249 & 0.210 & 19.42\% & 2.604 & 1.161 & 5.208 & 10.46\% & 3.024 & \underline{1.309} & 5.617 & 20.71\% & \multirow{4}{*}{7/48} \\
& 192 & \underline{0.197} & \underline{0.351} & \textbf{0.333} & \underline{26.62\%} & 4.419 & 1.571 & \textbf{9.304} & 13.66\% & \underline{5.562} & \underline{1.819} & 11.092 & \underline{28.37\%} \\
& 336 & \underline{0.337} & \underline{0.460} & \textbf{0.575} & \underline{34.83\%} & 10.053 & 2.544 & 17.222 & 20.58\% & \underline{10.474} & \underline{2.466} & 23.334 & \underline{39.66\%} \\
& 720 & \textbf{0.873} & \textbf{0.732} & \underline{1.043} & \textbf{54.42\%} & 17.726 & \underline{3.311} & \textbf{29.015} & 27.26\% & 21.393 & 3.610 & 40.060 & 61.40\% \\
\midrule
\multirow{4}{4em}{FITS}
& 96 & 0.120 & 0.277 & 0.264 & 20.64\% & \underline{2.392} & \underline{1.125} & 4.278 & \underline{10.03\%} & 3.163 & 1.385 & 6.150 & 21.18\% & \multirow{4}{*}{0/48} \\
& 192 & 0.243 & 0.393 & 0.472 & 29.58\% & 4.723 & 1.637 & 9.919 & 14.12\% & 5.701 & 1.880 & \underline{11.051} & 28.72\% \\
& 336 & 0.405 & 0.485 & 0.851 & 38.17\% & 8.658 & 2.267 & \underline{16.092} & 19.10\% & 10.886 & 2.541 & 20.806 & 40.43\% \\
& 720 & 1.066 & 0.808 & 2.415 & 60.14\% & 23.621 & 3.900 & 59.070 & 31.47\% & 22.865 & 3.720 & 36.404 & 63.48\% \\
\midrule
\multirow{4}{4em}{FITS100}
& 96 & 0.119 & 0.273 & 0.247 & 20.54\% & 2.459 & 1.153 & 4.802 & 10.17\% & 3.174 & 1.391 & 6.469 & 21.22\% & \multirow{4}{*}{0/48} \\
& 192 & 0.239 & 0.389 & 0.443 & 29.36\% & 4.738 & 1.642 & 10.412 & 14.15\% & 5.701 & 1.881 & 11.448 & 28.73\% \\
& 336 & 0.404 & 0.487 & 0.817 & 38.13\% & 8.704 & 2.278 & 17.656 & 19.15\% & 10.892 & 2.545 & 20.770 & 40.44\% \\
& 720 & 1.048 & 0.802 & 2.064 & 59.64\% & 23.623 & 3.919 & 64.024 & 31.47\% & 22.869 & 3.723 & 35.469 & 63.48\% \\
\midrule
\multirow{4}{4em}{FITS10}
& 96 & 0.143 & 0.303 & 0.276 & 22.57\% & 2.704 & 1.239 & 5.193 & 10.66\% & 3.457 & 1.482 & 5.547 & 22.14\% & \multirow{4}{*}{1/48} \\
& 192 & 0.244 & 0.391 & 0.507 & 29.61\% & 5.167 & 1.749 & 12.032 & 14.77\% & 6.021 & 1.957 & 11.073 & 29.52\% \\
& 336 & 0.422 & 0.505 & 0.852 & 38.97\% & 9.303 & 2.394 & 21.951 & 19.80\% & 11.191 & 2.598 & \underline{20.616} & 40.99\% \\
& 720 & 1.065 & 0.822 & 1.739 & 60.12\% & 24.351 & 4.081 & 48.666 & 31.95\% & 22.852 & 3.737 & \textbf{33.523} & 63.46\% \\
\midrule
\end{tabular}
}}
\caption{Univariate input on univariate column}
\label{tab:univariate_univariate}
\end{table*}

\newpage

Note: While method \ref{tab:multivariate_univariate} has access to other variables during training, both methods only optimize for the target column since their loss functions ignore non-target predictions. This explains why their tables show such similar results -- they're effectively doing the same optimization. Attempting to make method \ref{tab:multivariate_univariate} mode predict or share weights from all columns resulted in substantially worse performance, probably because it diverts the model's focus from the actual target variable.

%% file: discussion.tex
\subsection{What does the Hurst Exponent really measure?}

In theory, a random walk should yield a Hurst Exponent of approximately $H=0.5$. However, as can be seen from figure  \ref{fig:dicussion1} below, data which clearly has a non-random underlying signal, can also yield a $H \approx 0.5$ when non-stationary and noisy enough.

Although one of the steps  when estimating $H$ is detrending the (entire) signal, we notice that if a noisy signal has a, or more, linear trends, the Hurst exponent also seems to become an unreliable measure of unpredictability. Similarly, given that the $H$ value for simple sine waves or combinations of sine waves without noise indicates "anti-persistent behavior" and is also lower than that of an actual (approximated) random walk, one should exercise caution before using it as a definitive measure.

In retrospect, it seems that the Hurst Exponent seems less useful for quantifying predictability in terms of trends or random walk behaviour but instead serves as a decent measure of the amount of noise in a signal (ref. figure \ref{fig:figure10}), with its simple to understand value between 0 (white noise) and 1 (no noise). 

In any case, the Hurst exponent is a quick, inexpensive, and effective calculation, and while inconclusive by itself, it can serve in conjunction with other indicators like the autocorrelation function (ACF) or Naïve Forecast (NF) baseline benchmark to paint a broader picture.

\begin{figure}[H]
    \centering
    \includegraphics[width=1\linewidth]{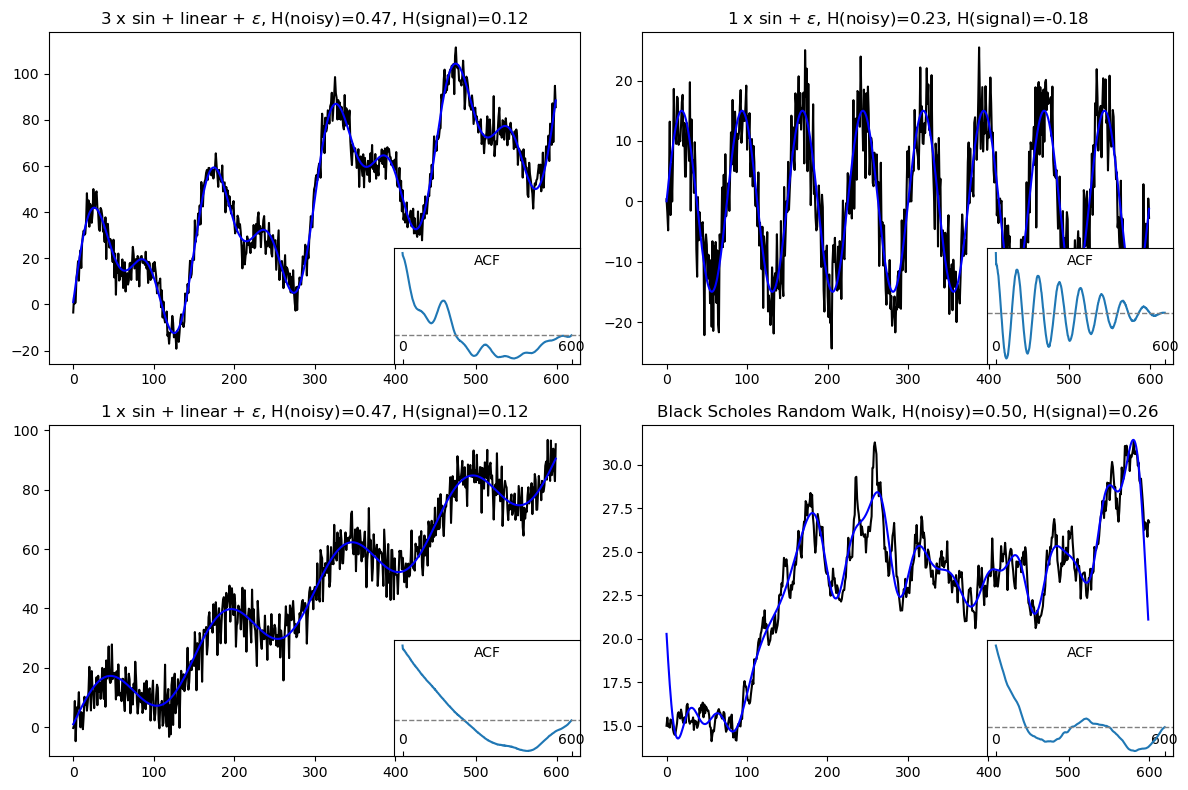}
    \caption{Hurst Exponent for different types of synthetically generated data. The ACF plot is of the noisy (black) graph. The  "Signal" in the random walk has had a simple low pass filter applied.}
    \label{fig:dicussion1}
\end{figure}

\subsection{Autocorrelation Functions (ACF) -- an insight into why FITS performs differently}
As explained earlier, FITS upscales a signal to a longer length using the frequency domain, relying on the fast Fourier transform for the translation back and forth between time and frequency domain, with the weights learning the proper interpolation between the input and output frequencies. 

While a Fourier transformation can decompose arbitrary signals into frequency components, not all signals are necessarily modeled appropriately in the frequency domain. Fourier transformation excels in analyzing stationary signals—those whose statistical properties do not change over time. In such cases, the signal can be accurately represented as a sum of sinusoids with different frequencies, amplitudes, and phases. But when data has little to no periodic attributes, either globally or locally, FITS will not be able to model the data when, as the decomposed frequency components have just been used to fit to non-periodic attributes and they therefore hold no forecasting value. Most of the datasets we've tested are modelled well in the frequency domain, whereas exchange rate and the stock data in particular are unsuitable for FITS with naive baselines consistently performing better. 

Even though the channels in a dataset might be heavily correlated in a non periodic manner, FITS will not be able to model these correlations. This limits the utility of FITS in time series applications where the data lacks these periodic or seasonal attributes, such as systems where the input is random such as randomly applied speeder or brake pressure in relation to the acceleration of an electric car, where those attributes are directly correlated, but non-periodic.

Autocorrelation functions, have shown to be incredibly informative (ref. figure \ref{fig:figure18}) in understanding why FITS performs better on datasets like (ETTh1, ETTh2, ETTm1, ETTm2, Weather, Electricity, \& Traffic) compared to other datasets which were omitted in the original paper. These datasets show clear periodicity, often coinciding with 24h cycles or compared to the more uninformative ACF plot of the Exchange Rate dataset where the correlation seems non periodic and converges towards zero correlation. And the datasets where there is clear periodicity from the ACF plots we can also see that FITS is able to perform really well.

\subsection{Evaluation of ARIMA}

In the results section, specifically table \ref{tab:gigatable}, ARIMA is included due to it being a standard model for statistical forecasting. The results of ARIMA do not look particularly favorable, however, compared to the other models, but we assert that it is due to the fact that ARIMA models do not perform well on the long-range forecasting horizons, the uncertainty grows at larger time steps. Depending on the values of $p$ and $q$, ARIMA achieves best performs on short horizons, especially if no seasonality is in the data.

Along with that fact, we do not report ARIMA results for the forecast horizons of over 96, equally much due to the performance of ARIMA at such forecast lengths, but also due to the fact that the computation time for it is abnormally high compared to FITS - especially when the data contains a large amount of channels.

The FITS model we do a lot better for the aforementioned cases, as it represents the data in the frequency domain after a Fourier transformation of the data, and then learns a linear interpolation of the frequencies; the ARIMA model is fitted only on the observed part of the test data, and trained on the forecasting part. This is unlike the FITS model, which as a training set on which it trains both reconstruction and forecasting.

\subsection{FITS and the various attempts to improving it}

All of the models incorporating additional layers in the FITS interpolation step performed worse at the longest forecast horizon \ref{tab:Fits_variants_ETTm1} with a far higher parameter count. $\mathbb{C}\text{ReLU}$ deteriorated performance to a higher degree than ModReLU most likely due to implicit change in the phase of the complex numbers \ref{table:mse_results_deep_fits_comparison}. 

The real valued complex upscaler also performed quite poorly on everything but the shorter forecast horizons on ETTh1, likely due to similar problems as the $\mathbb{C}\text{ReLU}$ model but now causing problems throughout the entire network as the complex numbers no longer had proper representation. At shorter forecast horizons [96, 192, 336] the results are somewhat inconclusive as a minor statistically significant improvement was acheived \ref{table:mse_comparison} indicating that the ModReLU variant of deep FITS is able to learn more features in the frequency representation of the signal compared to a single weight matrix. Some of the reasons for the small improvements might simply be because the lowest MSE possible on the ETT datasets is around the previous FITS scores due to the noise inherent in the datasets and much further improvement might not be possible. On the longer forecast horizon [720] none of the models convincingly beat the default FITS implementation could be due to several factors. The increased model complexity of the deep FITS variants may not be capturing meaningful long-term patterns that generalize well to the test set for extended forecast horizons. This suggests that the added complexity might not be necessary or beneficial for longer-term predictions in these datasets. Additionally, the default FITS implementation may already be capturing the most relevant frequency components for long-term forecasting, with the additional complexity introduced by the deep variants not adding substantial value in terms of identifying more meaningful long-term patterns in the frequency domain.

The use of the Fourier transformation in the pipeline seems to be generally an excellent way to help forecast better, especially on data that shows seasonality and/or periodic patterns. This is the case for most of the datasets that the performance is evaluated on. More on this can be read in the data assessment section of Appendix \ref{appendix:A}. The addition of extra layers for processing in our experiments showed that these were unnecessary and that the default FITS implementation may already be capturing the most relevant frequency components for long-term forecasting. The additional complexity introduced by the deep variants might not be adding substantial value in terms of identifying more meaningful long-term patterns in the frequency domain.

\subsection{DLinear vs FITS and hybrid model performance}

DLinear and FITS represent two relatively simple yet very effective approaches to time series forecasting compared to prior models. While both utilize linear layers, they differ significantly in how they transform and process the input data.

In table \ref{tab:multivariate_multivariate} and \ref{tab:gigatable}, we compared the results of DLinear vs FITS, which reveal some interesting patterns in the performance of these models.

On datasets with clear periodic patterns (e.g., ETT datasets, Weather, Electricity, Traffic), FITS often outperforms DLinear, especially at longer forecast horizons. This aligns with FITS' strength in capturing frequency-domain patterns.

For datasets with less obvious periodicity or more random-resembling behavior (like the pricing datasets in table \ref{tab:multivariate_multivariate}), DLinear tends to perform better than FITS, though often only marginally better than the naive baseline.

It would seem from the SE scores, that if one prioritizes what a stock price looks like in 192, 336, or 720 days, irregardless of the fluctuations along the way, the DLinear + FITS and FITS + DLinear hybrid models seem like the better option for this purpose. However, for capturing non-periodic and prices without drift like the Exchange Rate dataset, DLinear slightly outperforms the Naive Repeat baseline, compared to the rest.

The hybrid models also tend to outperform each individual model at longer forecast horizons and for multivariate targets. This suggests that the two approaches combined can complement each other, with FITS capturing periodic patterns and DLinear handling non-periodic trends and seasonality. It also suggests that having more cross-column context when predicting each column can yield better results, assuming the columns are uniquely informative about each other.

This improvement, although minor, suggests that there is potential in combining frequency-domain, time-domain, and perhaps exploring other mathematical representations of the data for gain new insights into the patterns of the time series. Future research could explore more sophisticated ways of integrating these complementary techniques to create more robust and versatile forecasting models.

In conclusion, it seems that architecturally simpler models beat bigger and more advanced models on time series forecasting tasks when properly decomposed into components that provide signals about the data's trend, fluctuations, and periodicity. 

%% file: appendix.tex
\appendix

\subsection*{Weights}

\begin{figure}[H]
    \centering
    \includegraphics[width=1\linewidth]{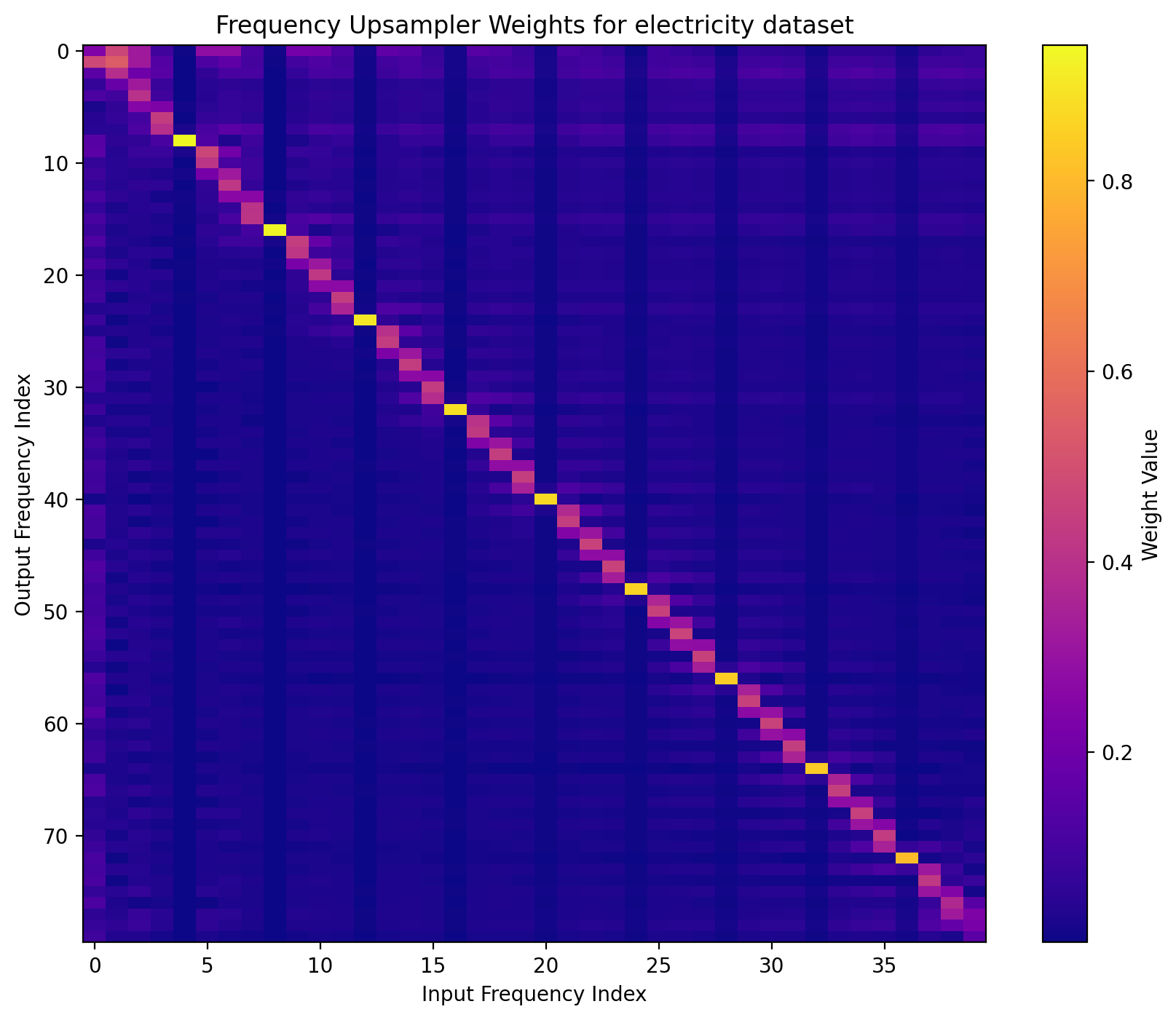}
    \caption{Weights for the electricity dataset [$\color{red} \rightarrow$ \ref{goback5}]}
    \label{fig:electricity_weights}
\end{figure}

\section*{Data assessment}
A problem that goes throughout most of the datasets, is their documentation, which is rather lacking. The lack of proper documentation makes it hard to make assumptions about the nature of the datasets, in the sense of how they will behave in the future. But even though the datasets are not documented properly, it is still possible to use them for evaluating our time series models as it is their performance in relation to each other they get evaluated on.
\newpage

\subsection*{ETT datasets}
\begin{figure}[H]
    \centering
    \begin{minipage}[t]{0.125\textwidth}
        \centering
        \includegraphics[width=\textwidth]{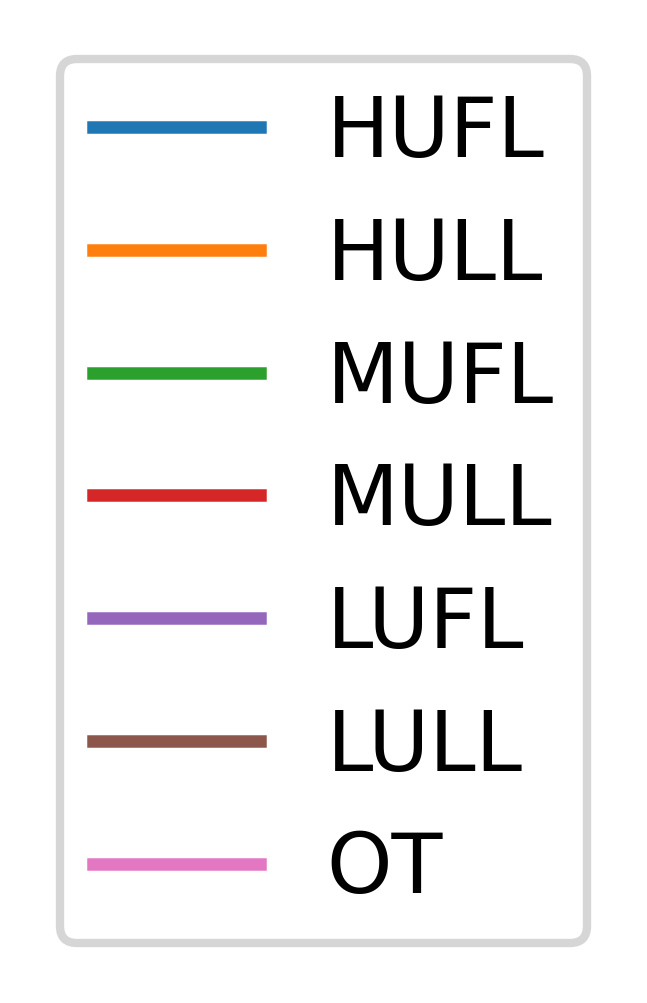}
        \caption*{Legend}
        \label{fig:ETTlegend}
    \end{minipage}
    \hfill
    \begin{minipage}[t]{0.86\textwidth}
        \centering
        \begin{minipage}[b]{0.49\textwidth}
            \centering
            \includegraphics[width=\textwidth]{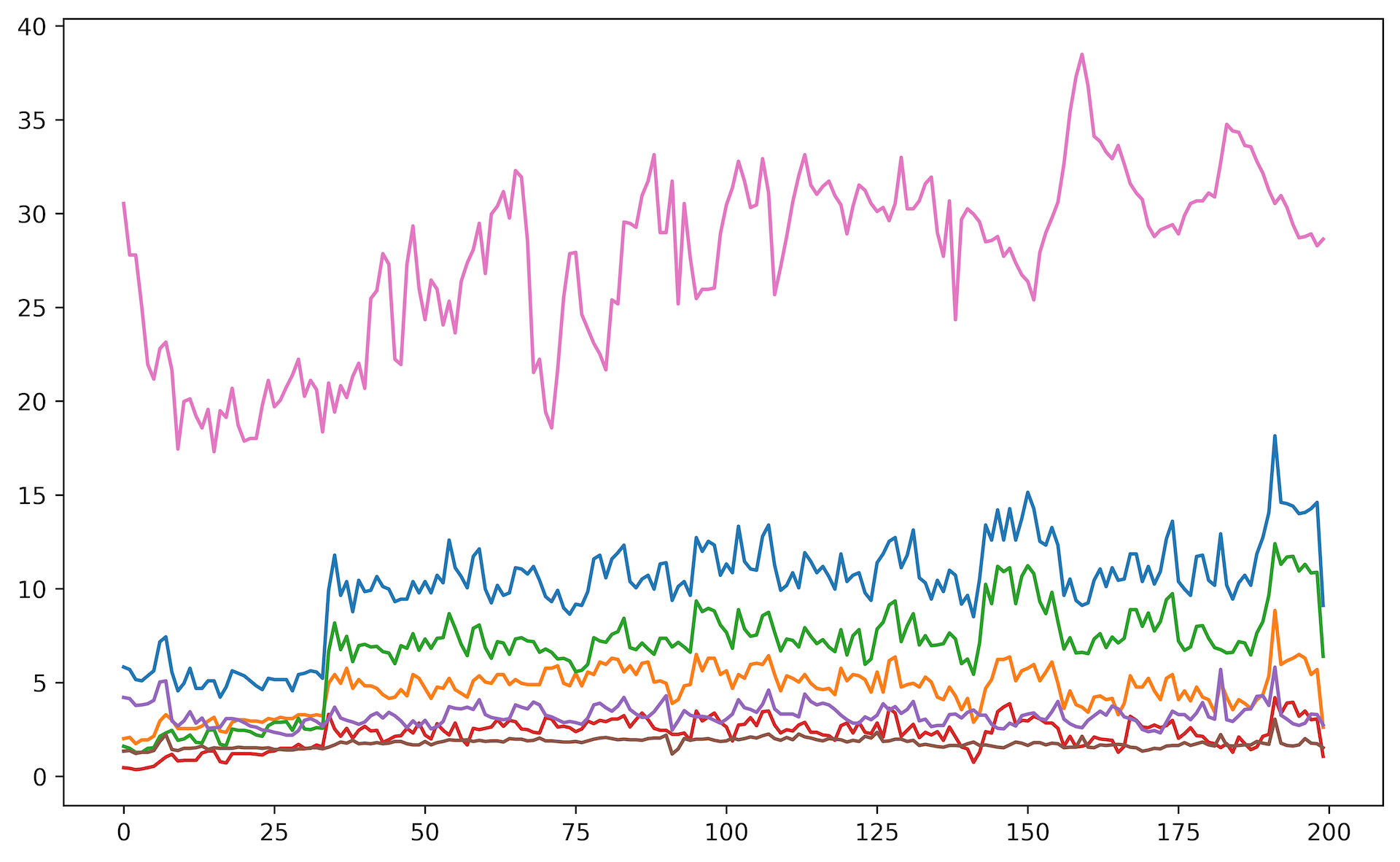}
            \caption{Plot of the channels in ETTh1}
            \label{fig:figure1}
        \end{minipage}
        \hfill
        \begin{minipage}[b]{0.49\textwidth}
            \centering
            \includegraphics[width=\textwidth]{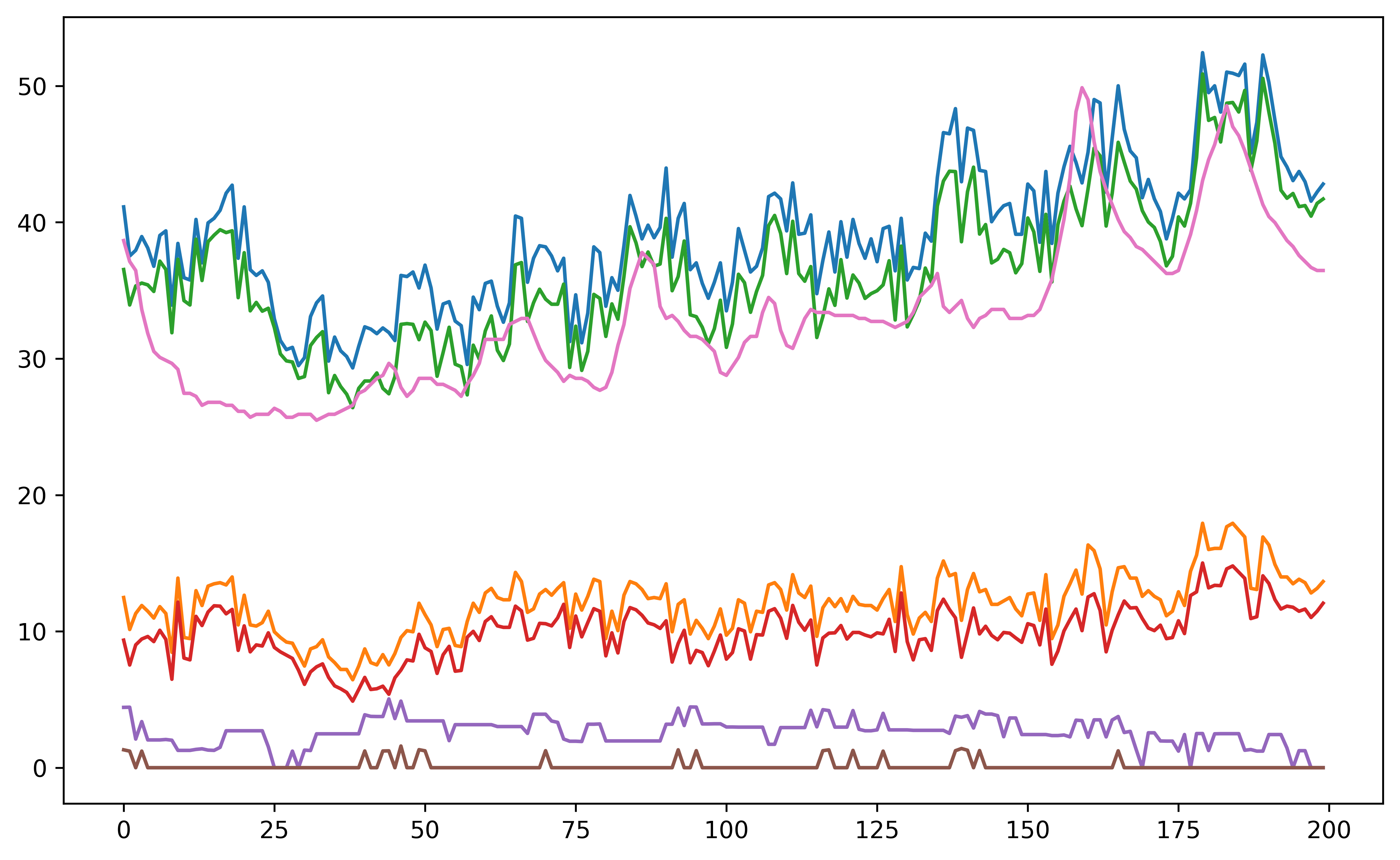}
            \caption{Plot of the channels in ETTh2}
            \label{fig:figure2}
        \end{minipage}
        \vfill
        \begin{minipage}[b]{0.49\textwidth}
            \centering
            \includegraphics[width=\textwidth]{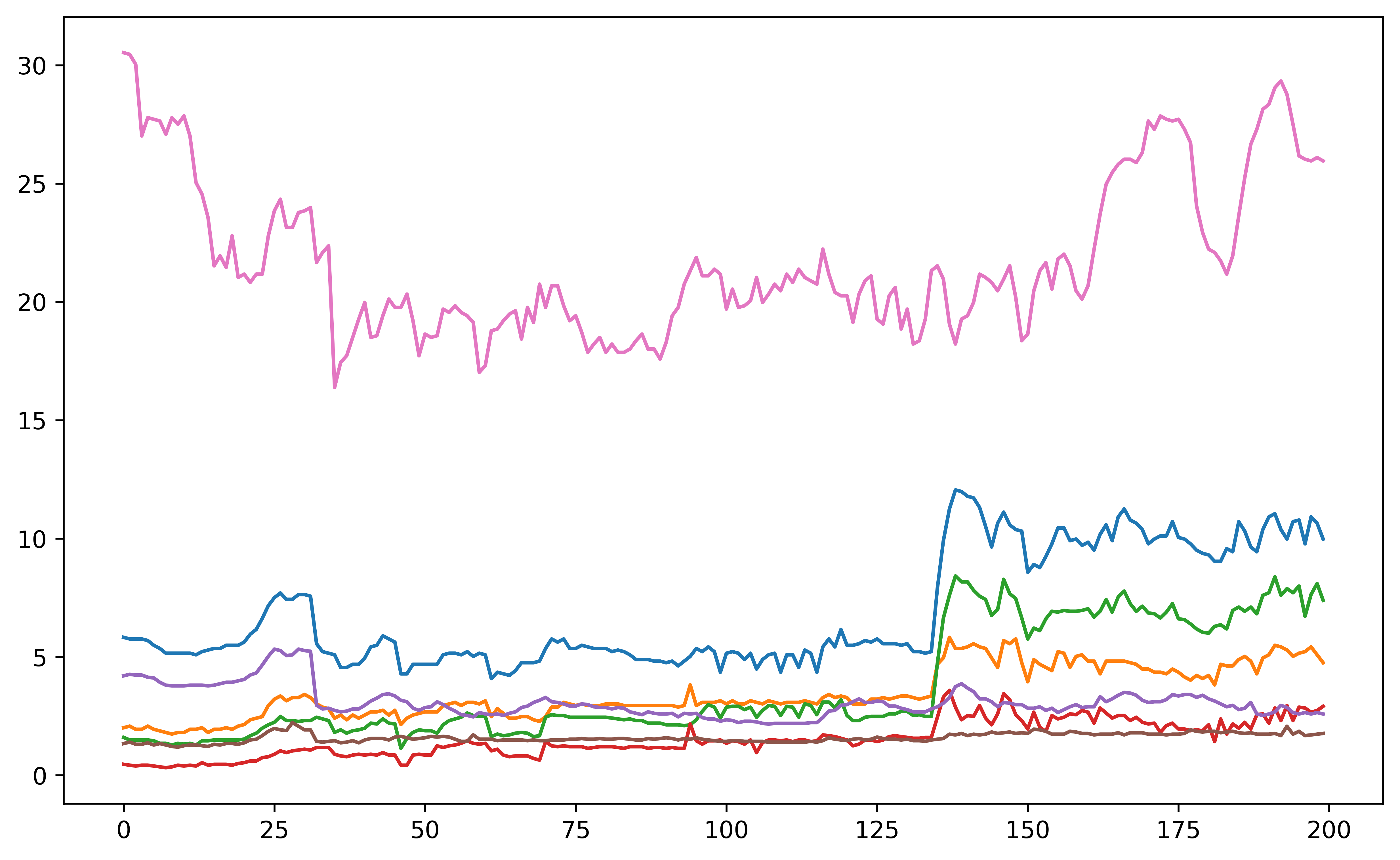}
            \caption{Plot of the channels in ETTm1}
            \label{fig:figure3}
        \end{minipage}
        \hfill
        \begin{minipage}[b]{0.49\textwidth}
            \centering
            \includegraphics[width=\textwidth]{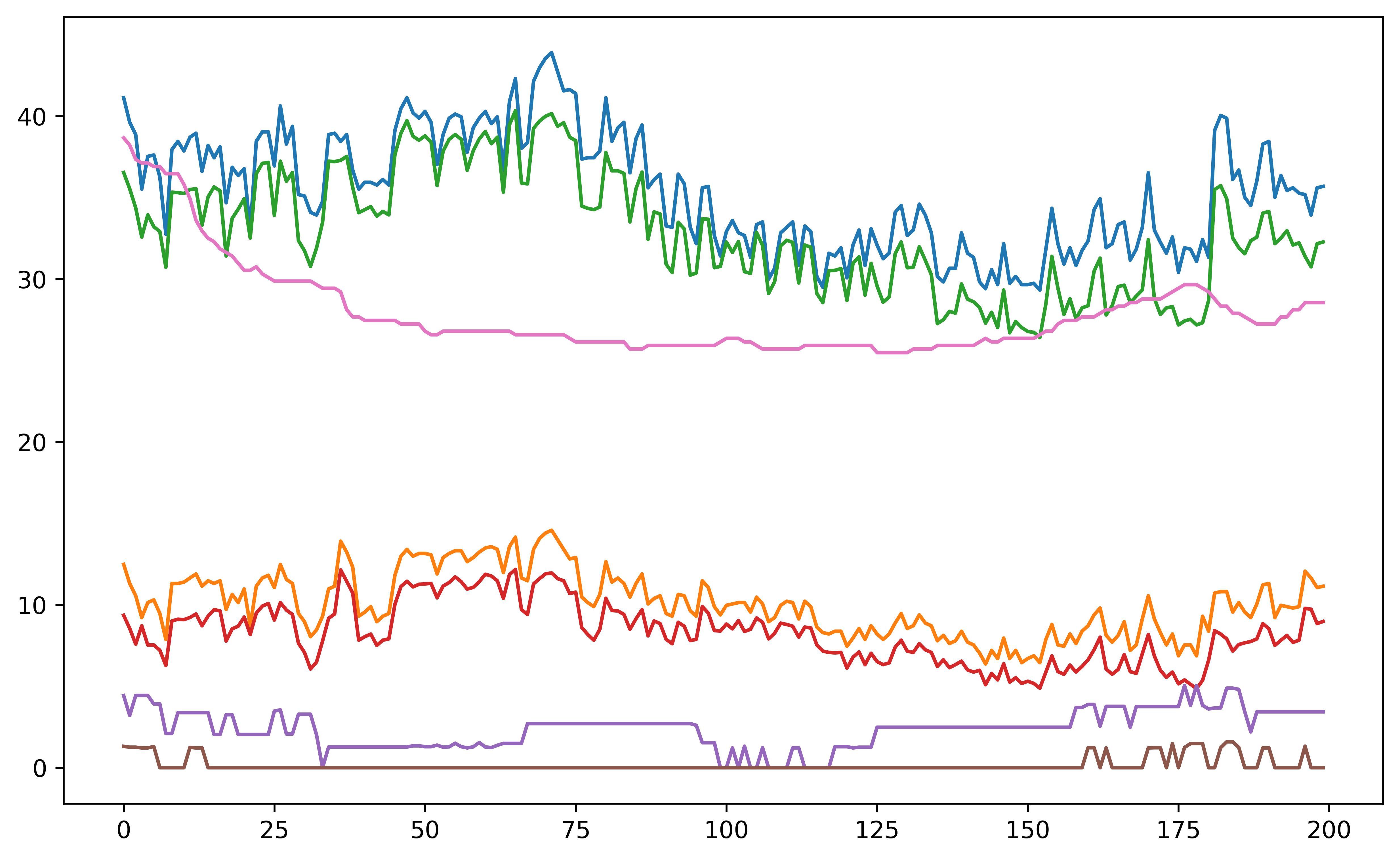}
            \caption{Plot of the channels in ETTm2}
            \label{fig:figure4}
        \end{minipage}
    \end{minipage}
\end{figure}
These plots are all clippings of 200 time steps from each of their respective whole datasets. ETThX and ETTmX are from the same datasets, where ETTmX samples each 15th minute, where ETThX samples once every hour, the X indicates the source of the data sampling.

The column OT is the column we forecast on since it denotes the oil temperature of the transformer station it is recorded at. It is quite apparent that all of the of the channels are quite correlated apart from the OT channel from doing a visual assessment alone.

\subsection*{Price datasets}

\begin{figure}[H]
    \centering
    \begin{minipage}[t]{0.125\textwidth}
        \centering
        \includegraphics[width=\textwidth]{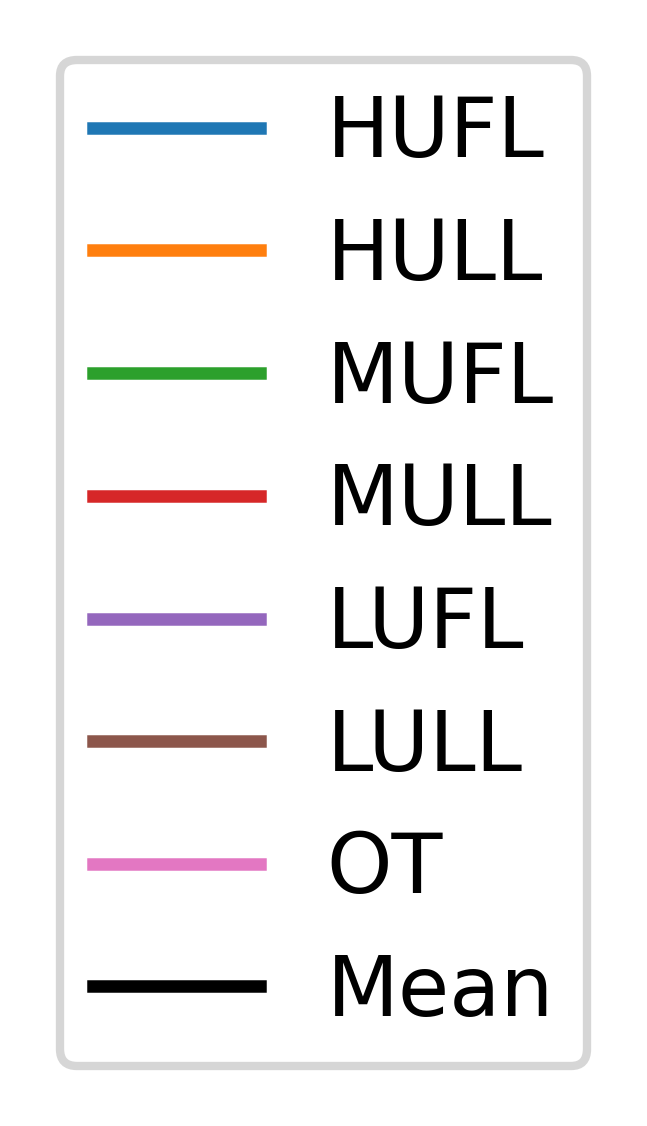}
        \caption*{Legend}
        \label{fig:ETTlegendWmean}
    \end{minipage}
    \hfill
    \begin{minipage}[t]{0.86\textwidth}
        \centering
        \begin{minipage}[b]{0.49\textwidth}
            \centering
            \includegraphics[width=\textwidth]{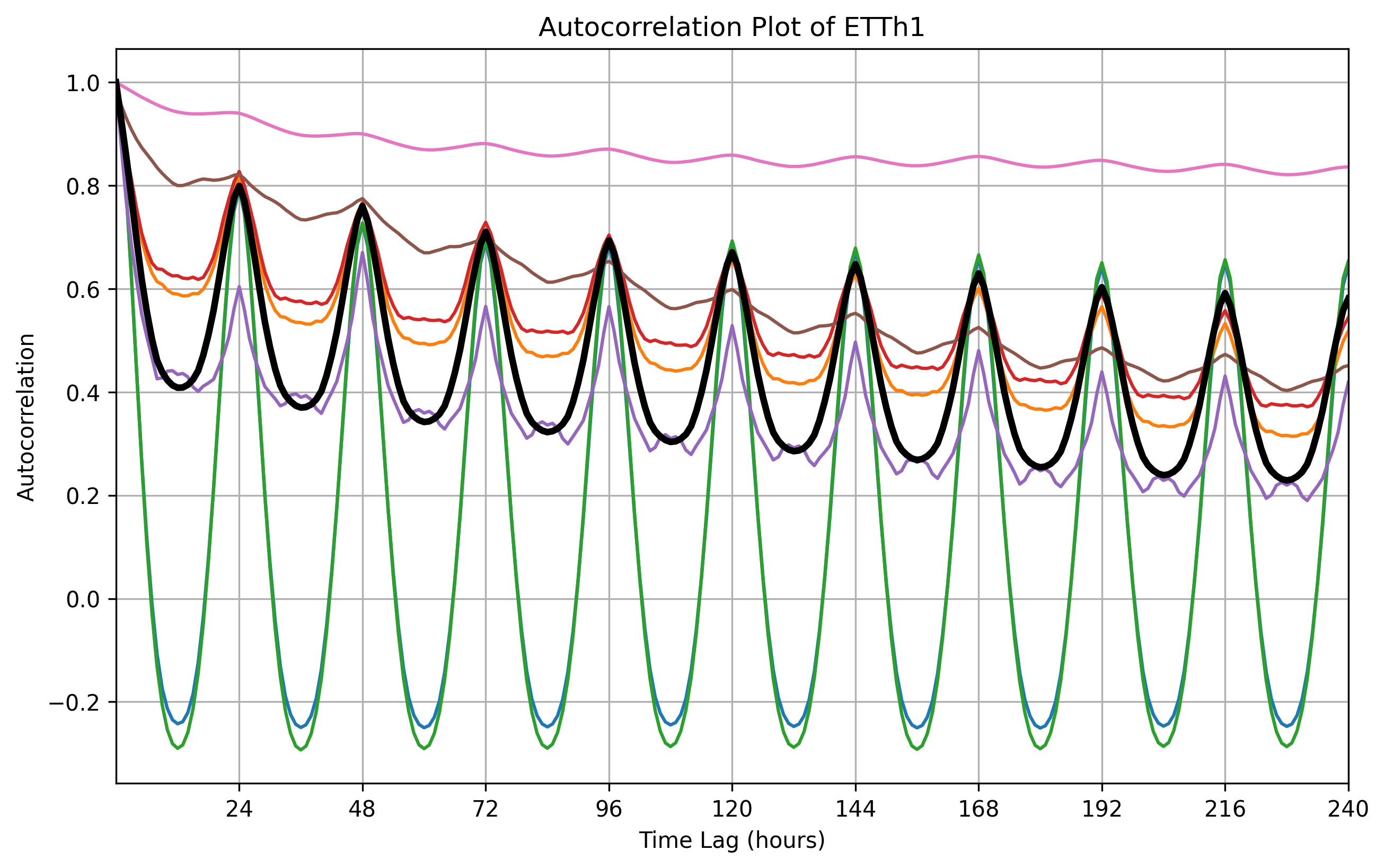}
            \caption{Time lag correlation, ETTh1}
            \label{fig:figure18}
        \end{minipage}
        \hfill
        \begin{minipage}[b]{0.49\textwidth}
            \centering
            \includegraphics[width=\textwidth]{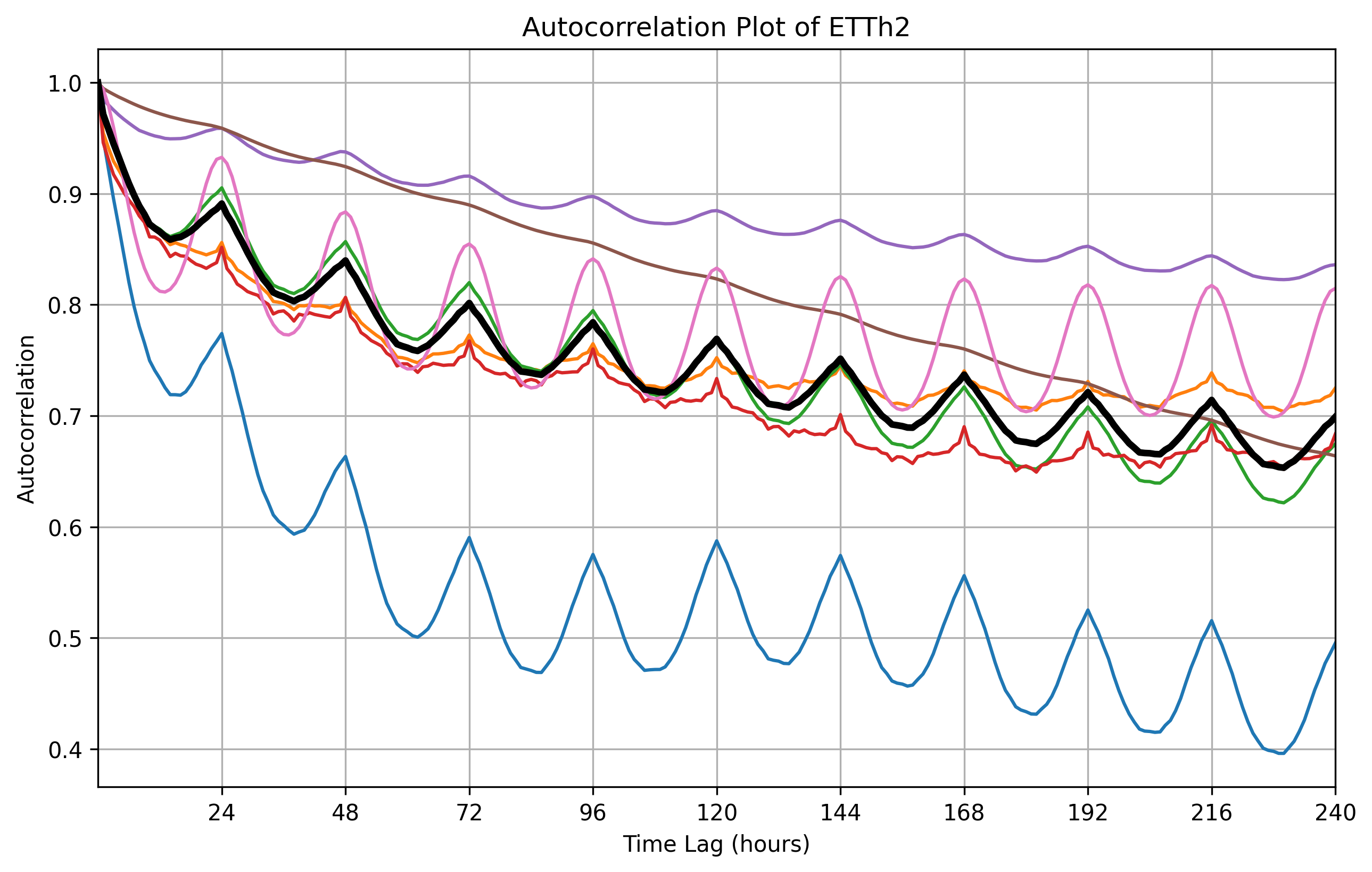}
            \caption{Time lag correlation, ETTh2}
            \label{fig:figure19}
        \end{minipage}
        \vfill
        \begin{minipage}[b]{0.49\textwidth}
            \centering
            \includegraphics[width=\textwidth]{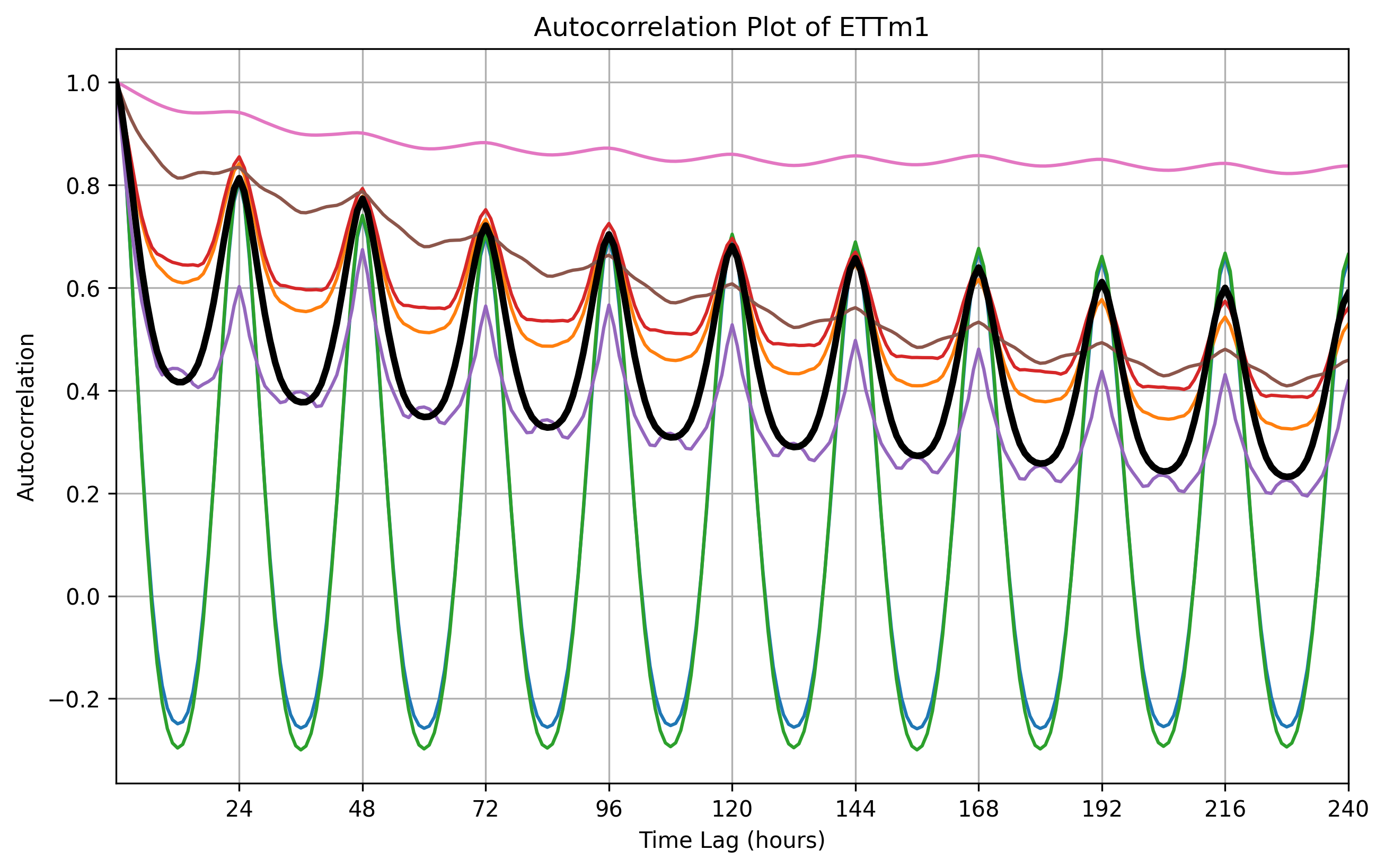}
            \caption{Time lag correlation, ETTm1}
            \label{fig:figure20}
        \end{minipage}
        \hfill
        \begin{minipage}[b]{0.49\textwidth}
            \centering
            \includegraphics[width=\textwidth]{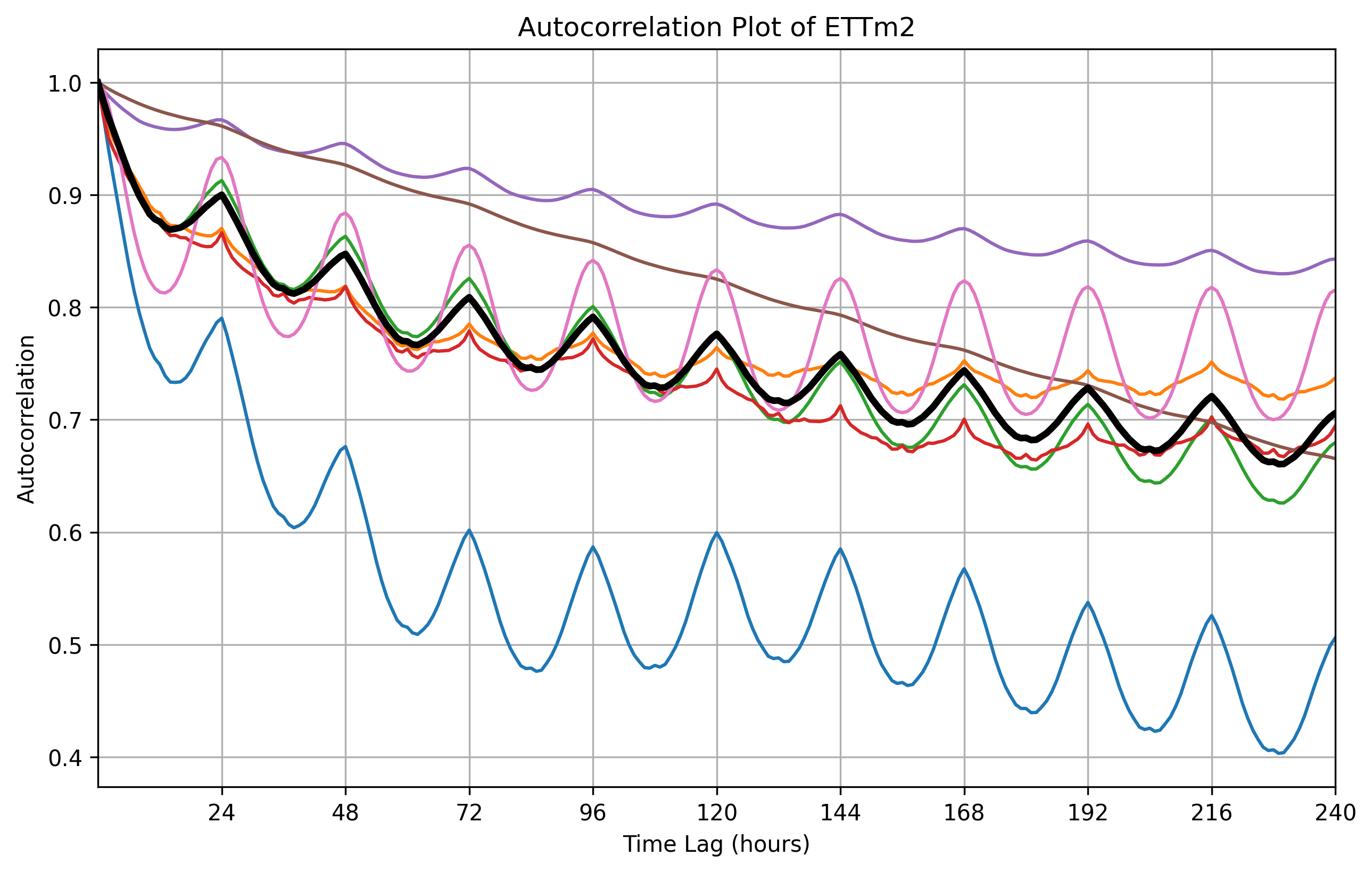}
            \caption{Time lag correlation, ETTm2}
            \label{fig:figure21}
        \end{minipage}
    \end{minipage}
\end{figure}
Afterwards we performed auto correlation on the data and saw that peaks between different channels happened at similar positions in the data and at high correlation values. This indicates strong correlation between the channels, which we also expected from our first visual assessment of the dataset. We can also observe that OT is highly correlated with its earlier and slightly positively correlated with the other channels as it shows slight peaking at the same positions. Because of the periodic peaking with high time lag correlation this indicates high periodicity in the data, which in theory should mean that there are frequencies our models should be able to pick up.

\begin{figure}[H]
    \centering
    \includegraphics[width=0.8\linewidth]{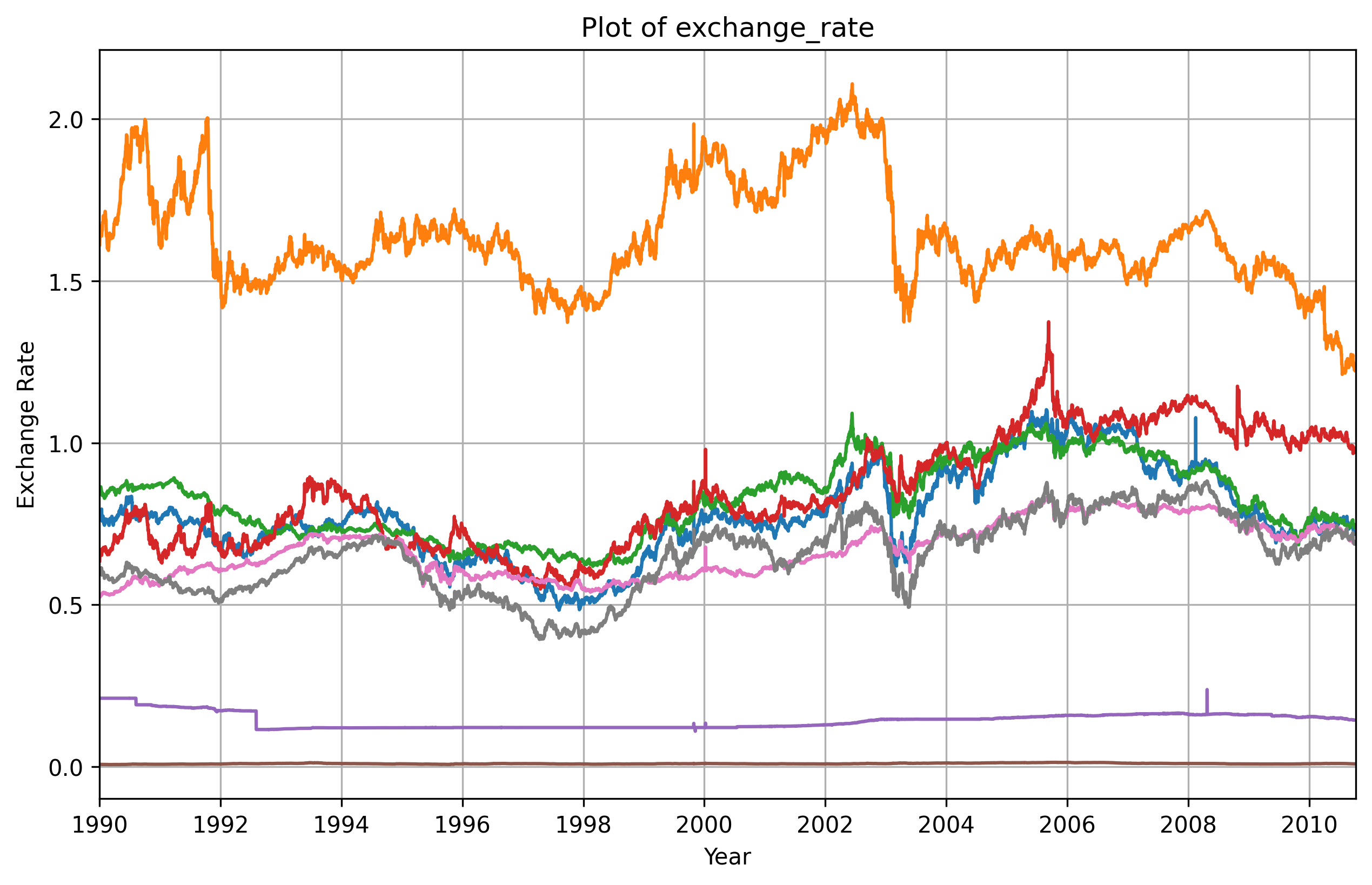}
    \label{fig:Exchange}
\end{figure}
This dataset consists of the exchange rates between different currencies. It isn't known which channel corresponds to which currency. On this small cutout of the data, there seems to be no discernible signs of correlation or periodicity between the channels.

\begin{figure}[H]
    \centering
    \includegraphics[width=0.8\linewidth]{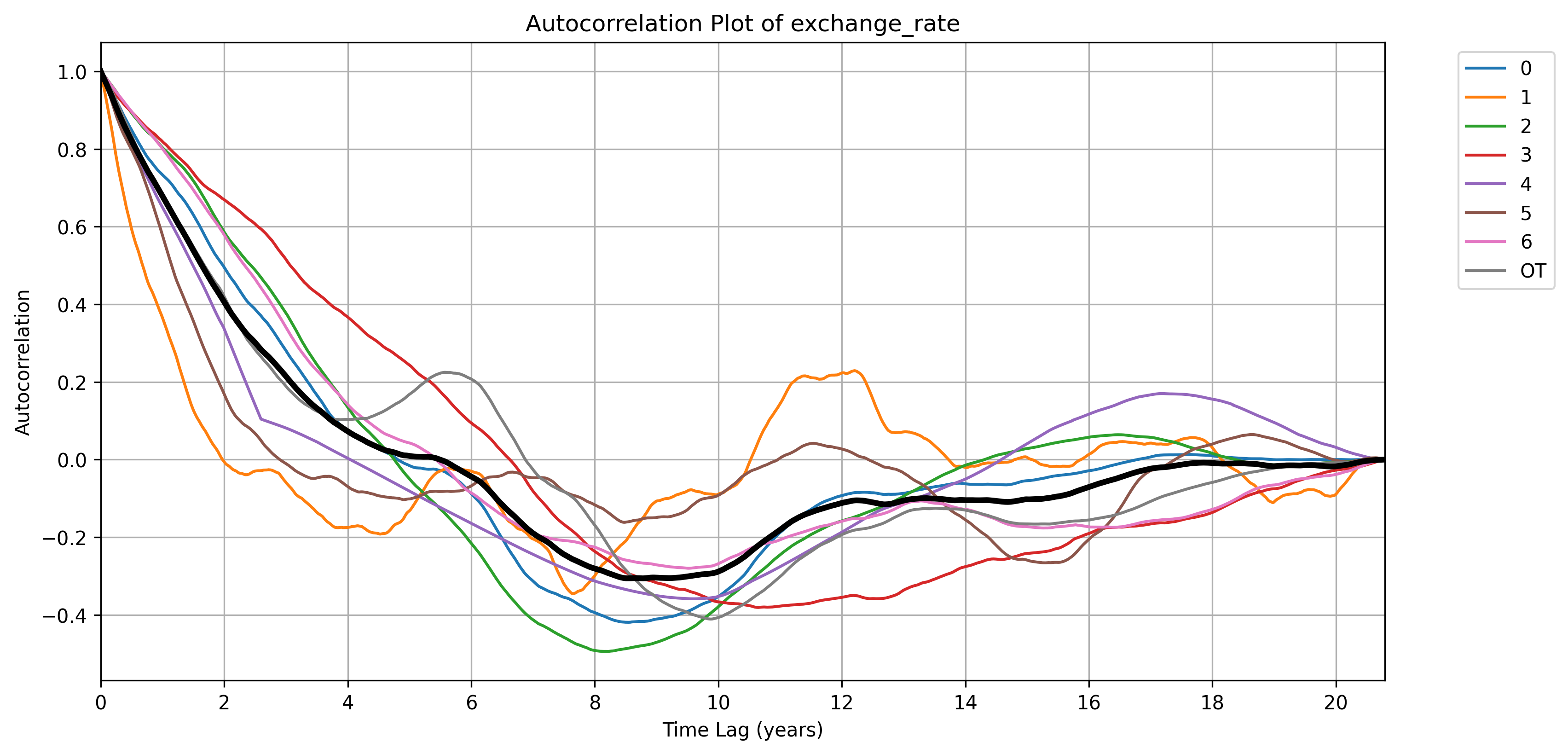}
    \label{fig:autocorr_plot_exchange_rate}
\end{figure}

\begin{figure}[H]
    \centering
    \includegraphics[width=1\linewidth]{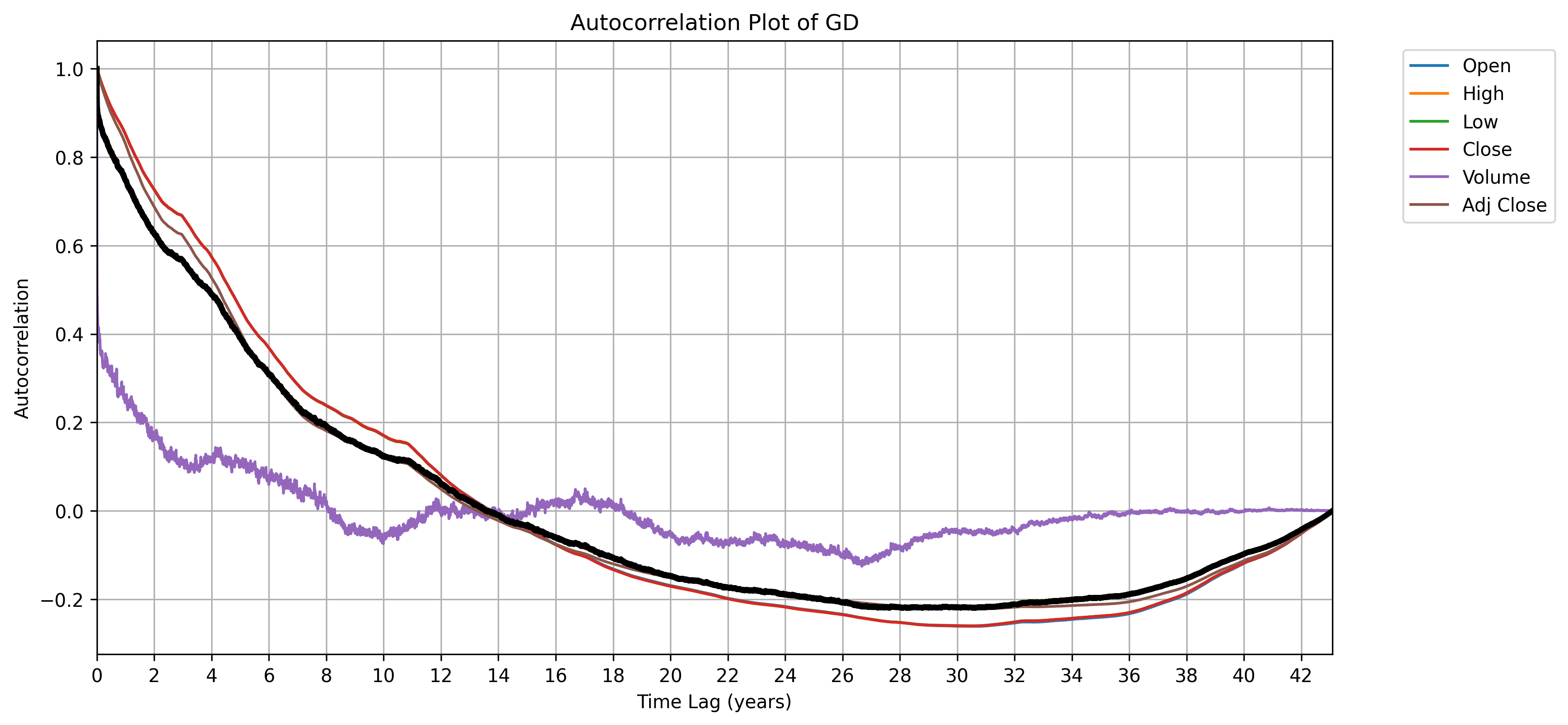}
    \label{fig:autocorr_plot_GD}
    \caption{Autocorrelation plot of \$GD}
\end{figure}

\begin{figure}[H]
    \centering
    \includegraphics[width=1\linewidth]{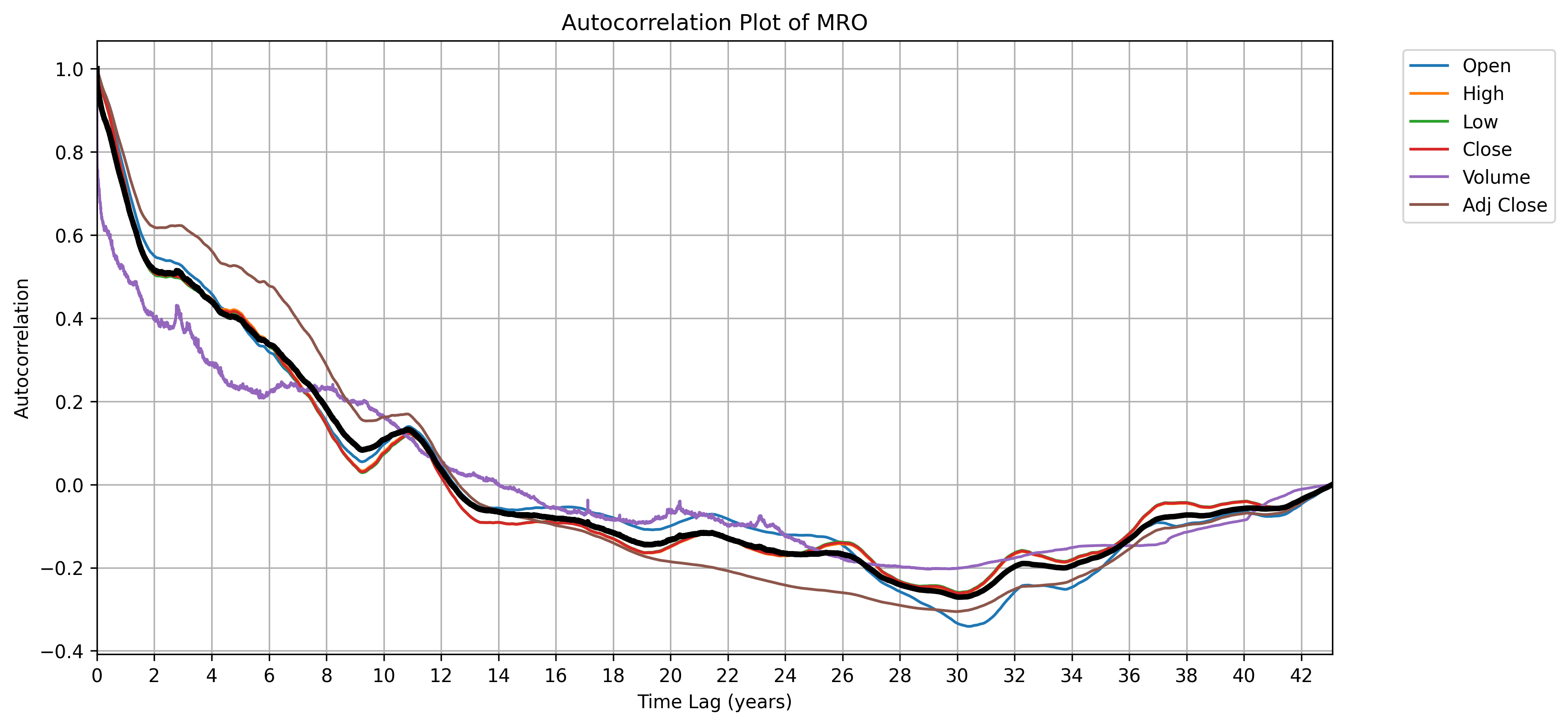}
    \label{fig:autocorr_plot_MRO}
    \caption{Autocorrelation plot of \$MRO}
\end{figure}

When performing auto correlation on the dataset we observe that there is no periodicity, but quite high correlation between previous values in most channels, except for channel 1, but the correlation is slowly diverging, which could indicate that it would be hard to make predictions based on frequencies. 
\newpage
\begin{figure}[H]
    \centering
    \includegraphics[width=0.8\linewidth]{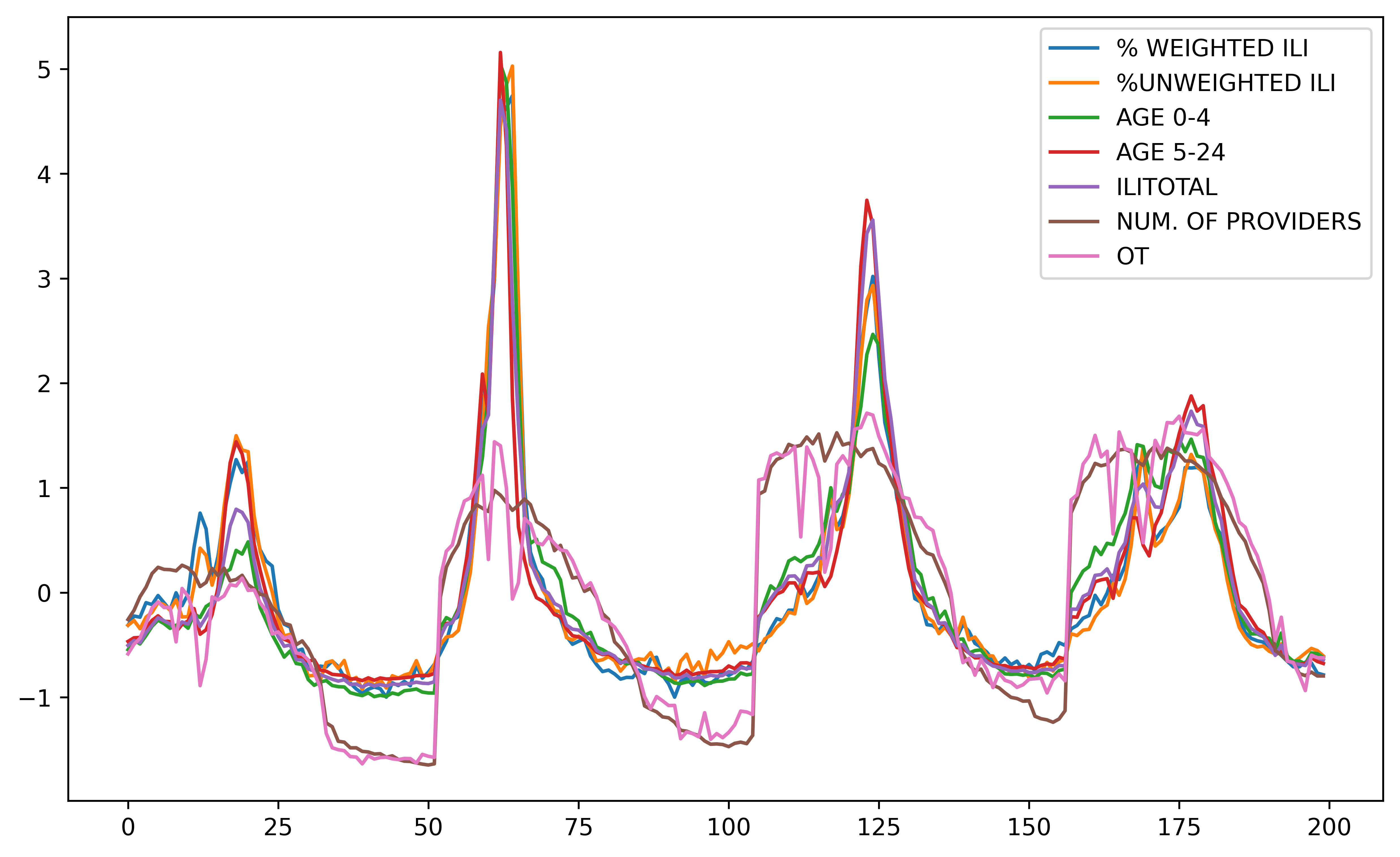}
    \label{fig:NationalIllness}
\end{figure}
It is quite apparent that there are infection patterns present as the peaks of the dataset seem to peak around the same time every year, as the x-axis represents weeks and there is a heavy increase in value of the OT at the same time of year. There is also peaking patterns in the rest of the dataset

\begin{figure}[H]
    \centering
    \includegraphics[width=0.8\linewidth]{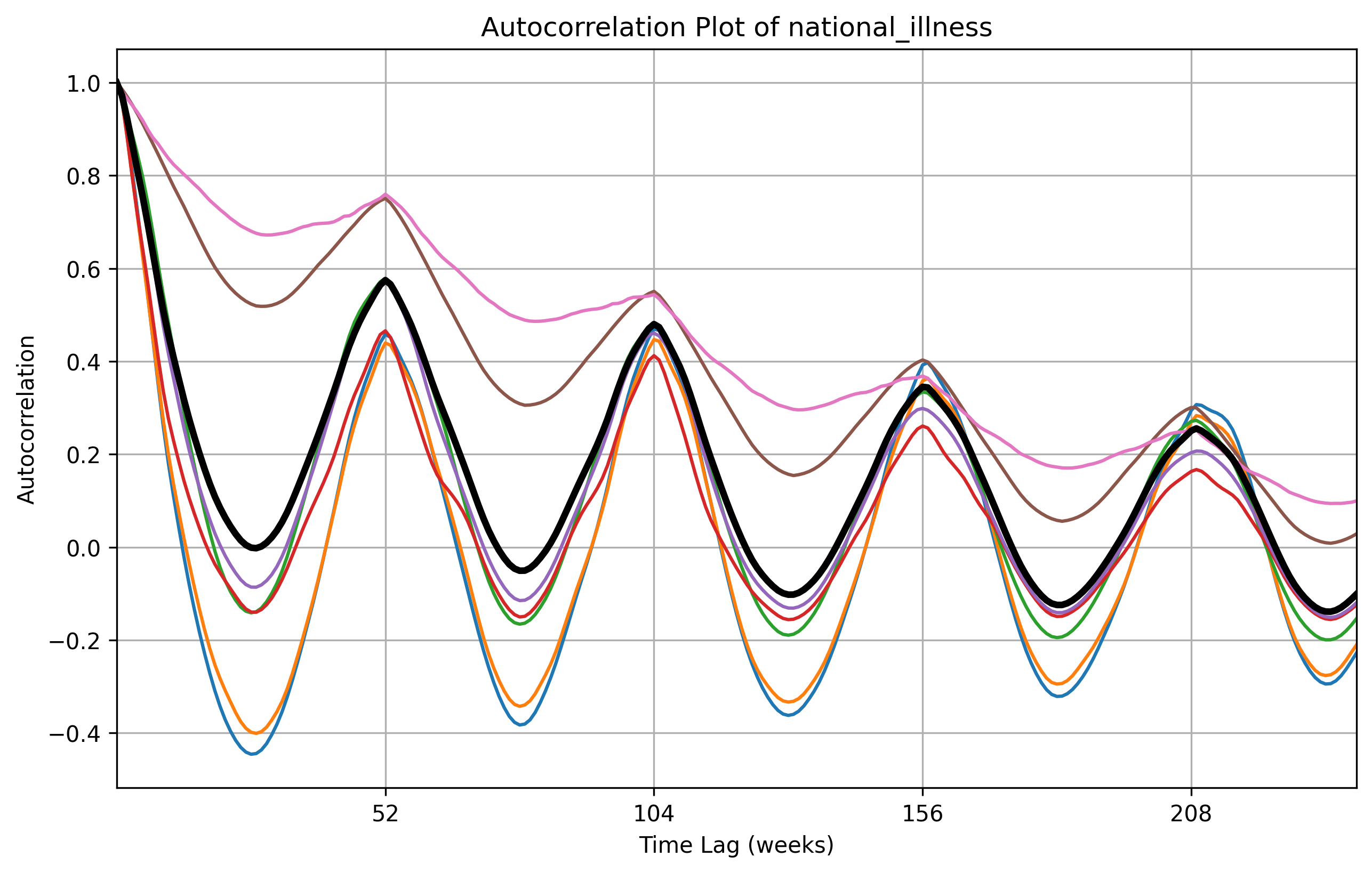}
    \label{fig:autocorr_plot_national_illness}
\end{figure}

We chose to plot the time lag in weeks because of the assumption that infection numbers are seasonal. The autocorrelation plot shows peaking around the same time of year (52 weeks) each time, which indicates that the dataset is periodic. OT (pink) also shows periodicity, though its prediction value lessens over time which makes sense as infection patterns change from each year.

\newpage
\begin{figure}[H]
    \centering
    \includegraphics[width=0.8\linewidth]{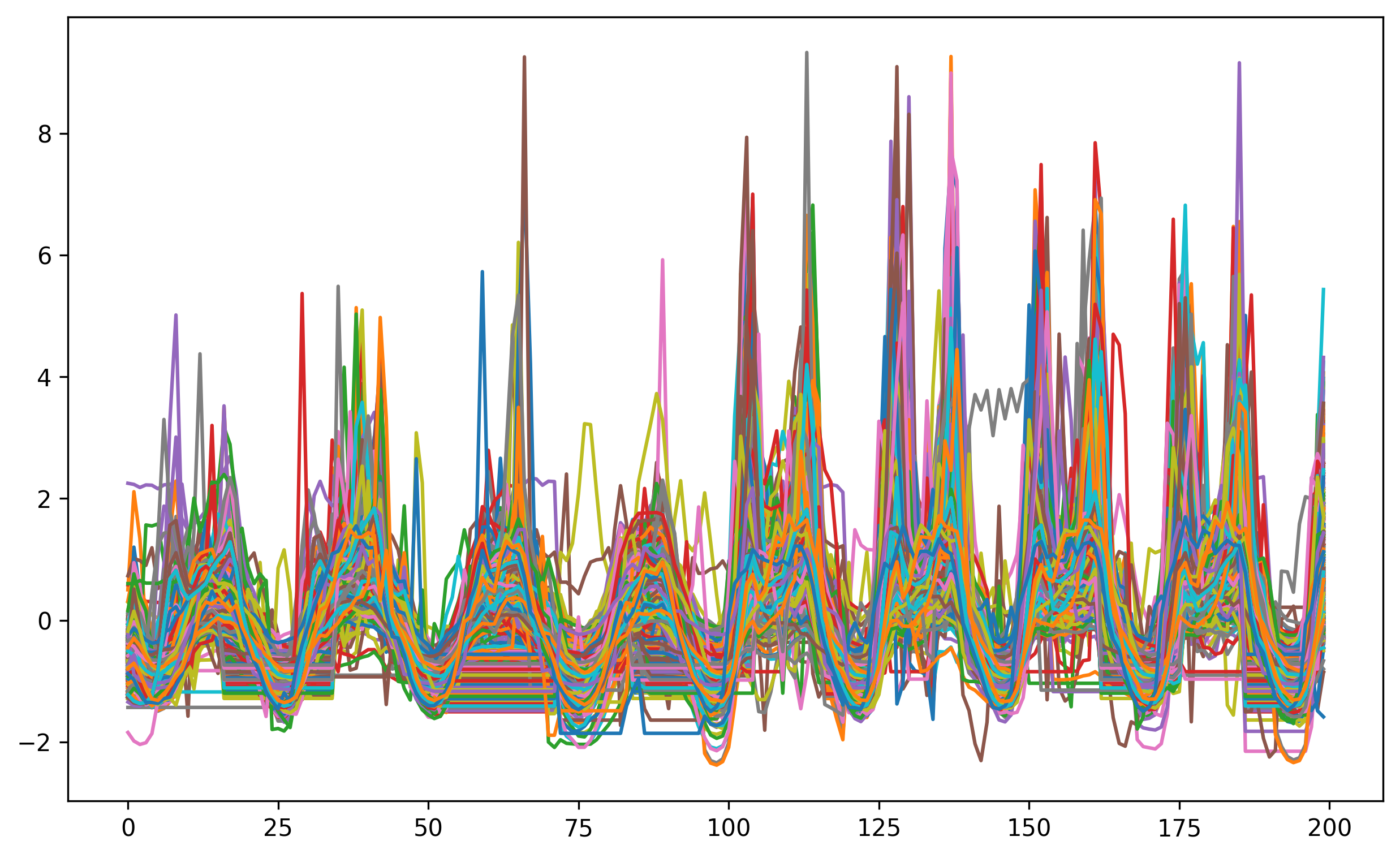}
    \label{fig:Traffic}
\end{figure}
The traffic dataset shows quite a bit of periodicity every 24 hours. This also makes sense as traffic patterns are quite affected by the time of day.
\begin{figure}[H]
    \centering
    \includegraphics[width=0.8\linewidth]{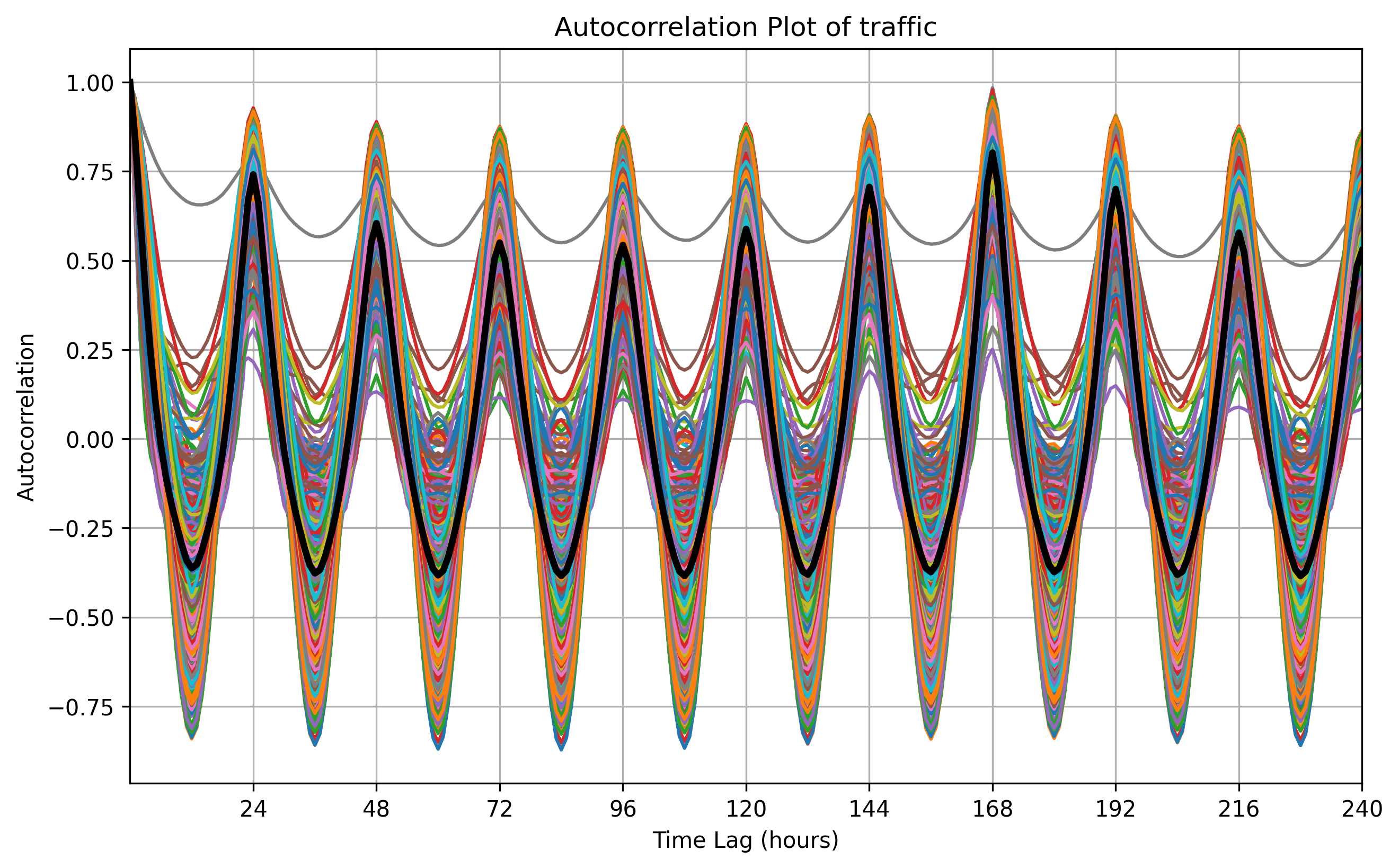}
    \label{fig:autocorr_plot_traffic}
\end{figure}
Performing time correlation on the dataset we can observe heavy periodicity with peaks sitting around the same time of day each day. Again this would indicate that there are predictable frequencies in the data which FITS will be able to effectively forecast on.
\newpage
\begin{figure}[H]
    \centering
    \begin{minipage}[t]{1.1\textwidth}
        \centering
        \begin{minipage}[b]{0.49\textwidth}
            \centering
            \includegraphics[width=\textwidth]{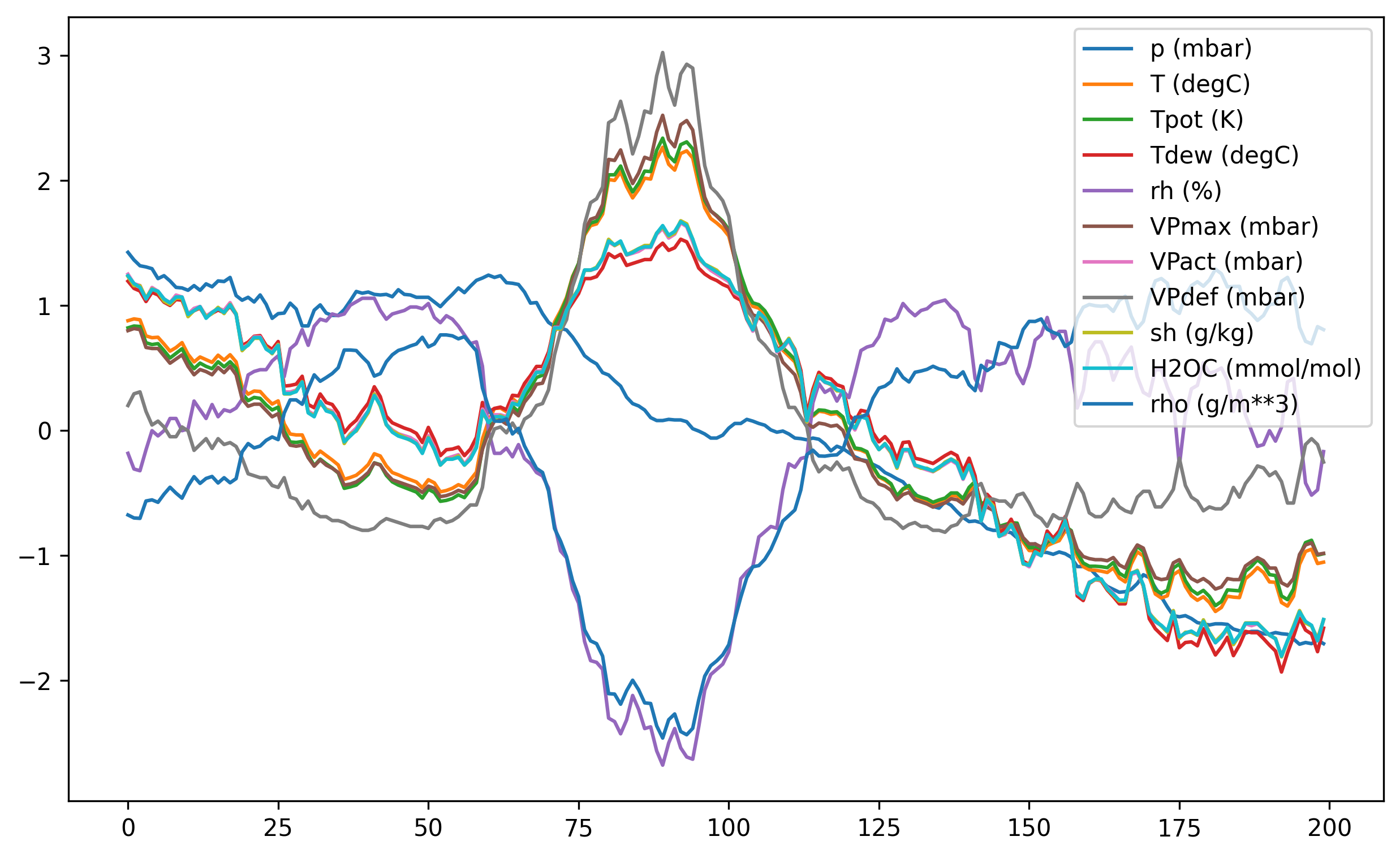}
            \caption{Plot, which is part of the weather dataset, the X-axis is Time, where each time step is 10 minutes}
            \label{fig:weather1}
        \end{minipage}
        \hfill
        \begin{minipage}[b]{0.49\textwidth}
            \centering
            \includegraphics[width=\textwidth]{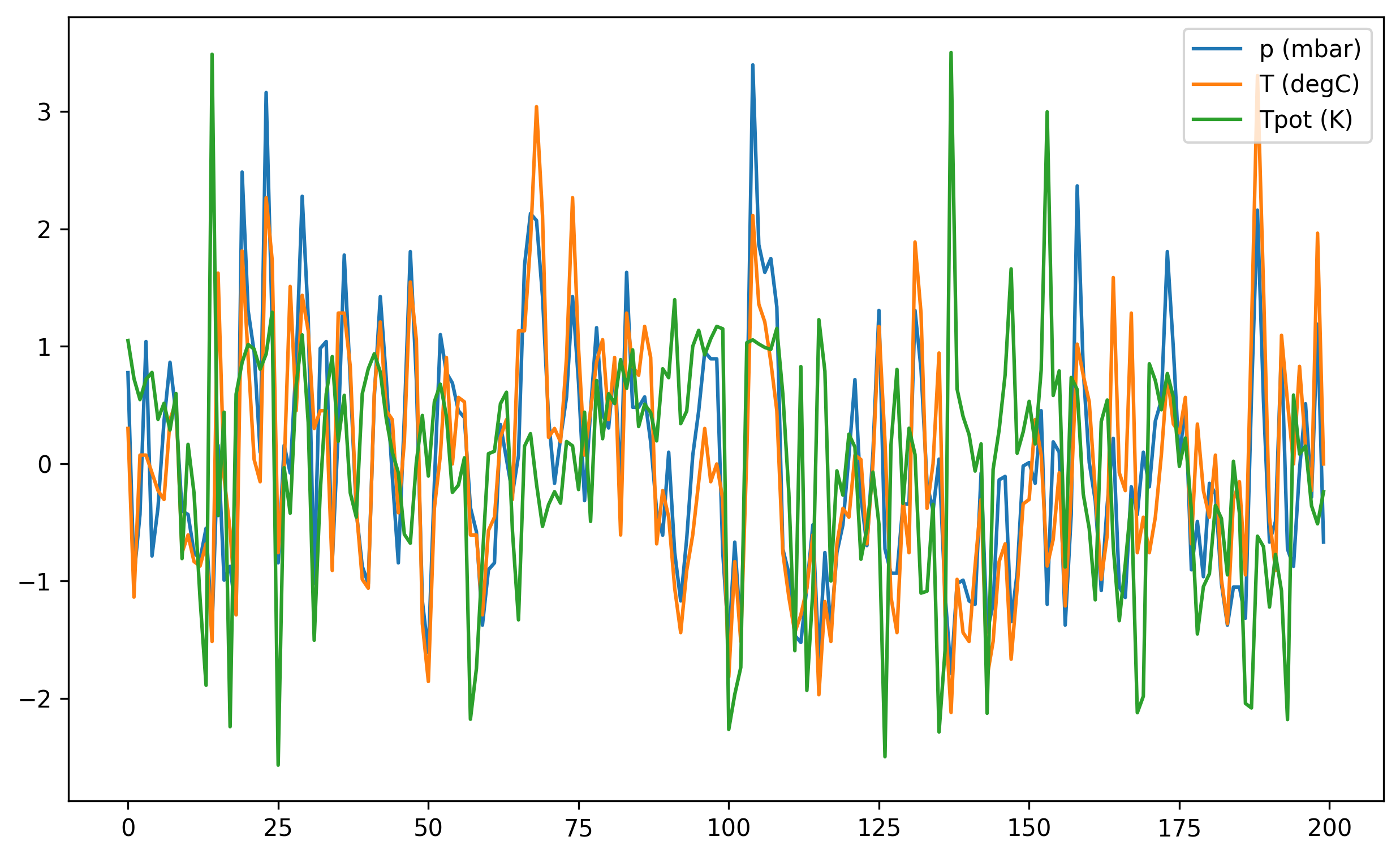}
            \caption{Plot, which is part of the weather dataset, the X-axis is Time, where each time step is 10 minutes}
            \label{fig:Weather2}
        \end{minipage}
        \vfill
        \begin{minipage}[b]{0.49\textwidth}
            \centering
            \includegraphics[width=\textwidth]{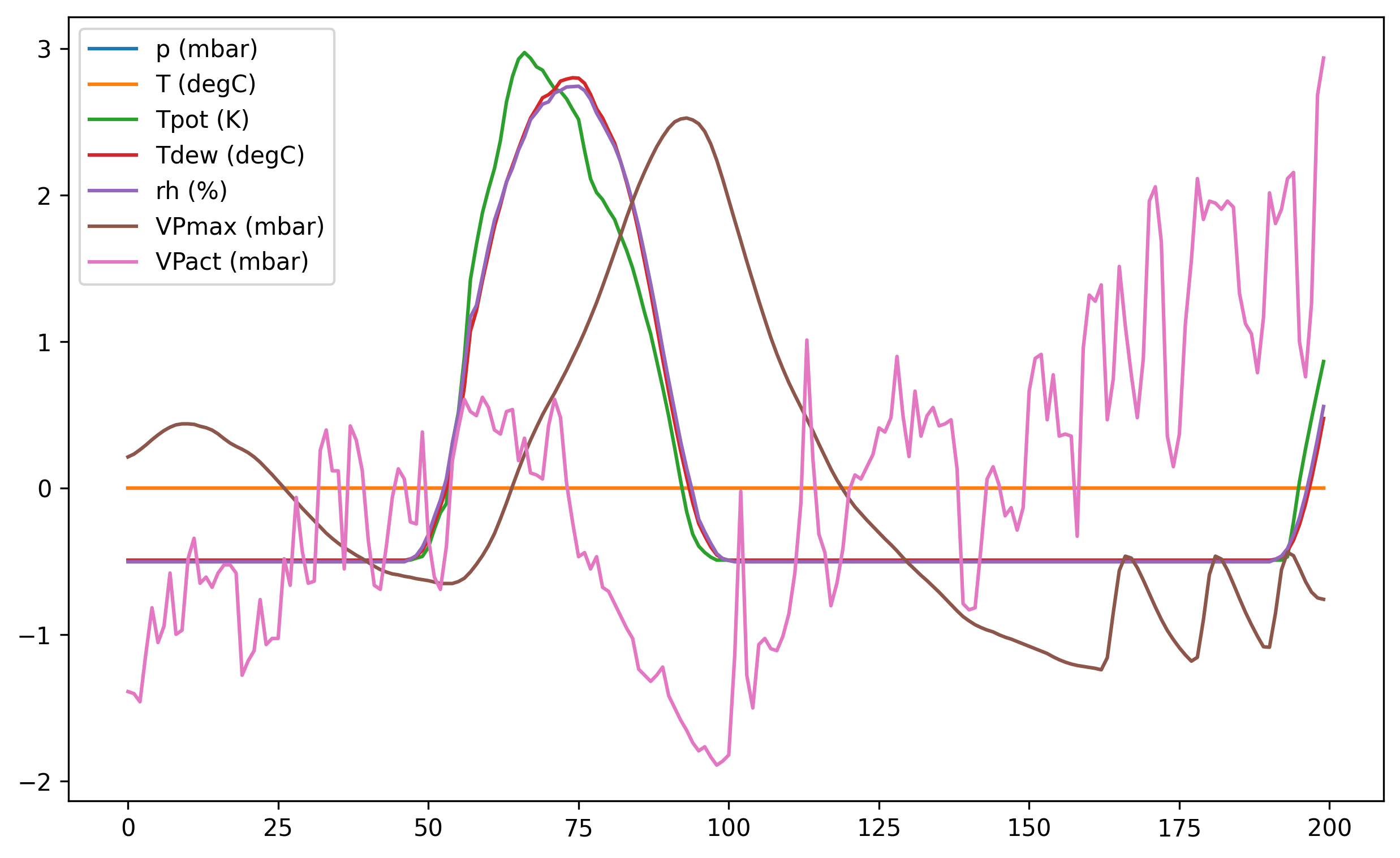}
            \caption{Plot, which is part of the weather dataset, the X-axis is Time, where each time step is 10 minutes}
            \label{fig:Weather3}
        \end{minipage}
        \hfill
        \begin{minipage}[b]{0.49\textwidth}
            \centering
            \includegraphics[width=\textwidth]{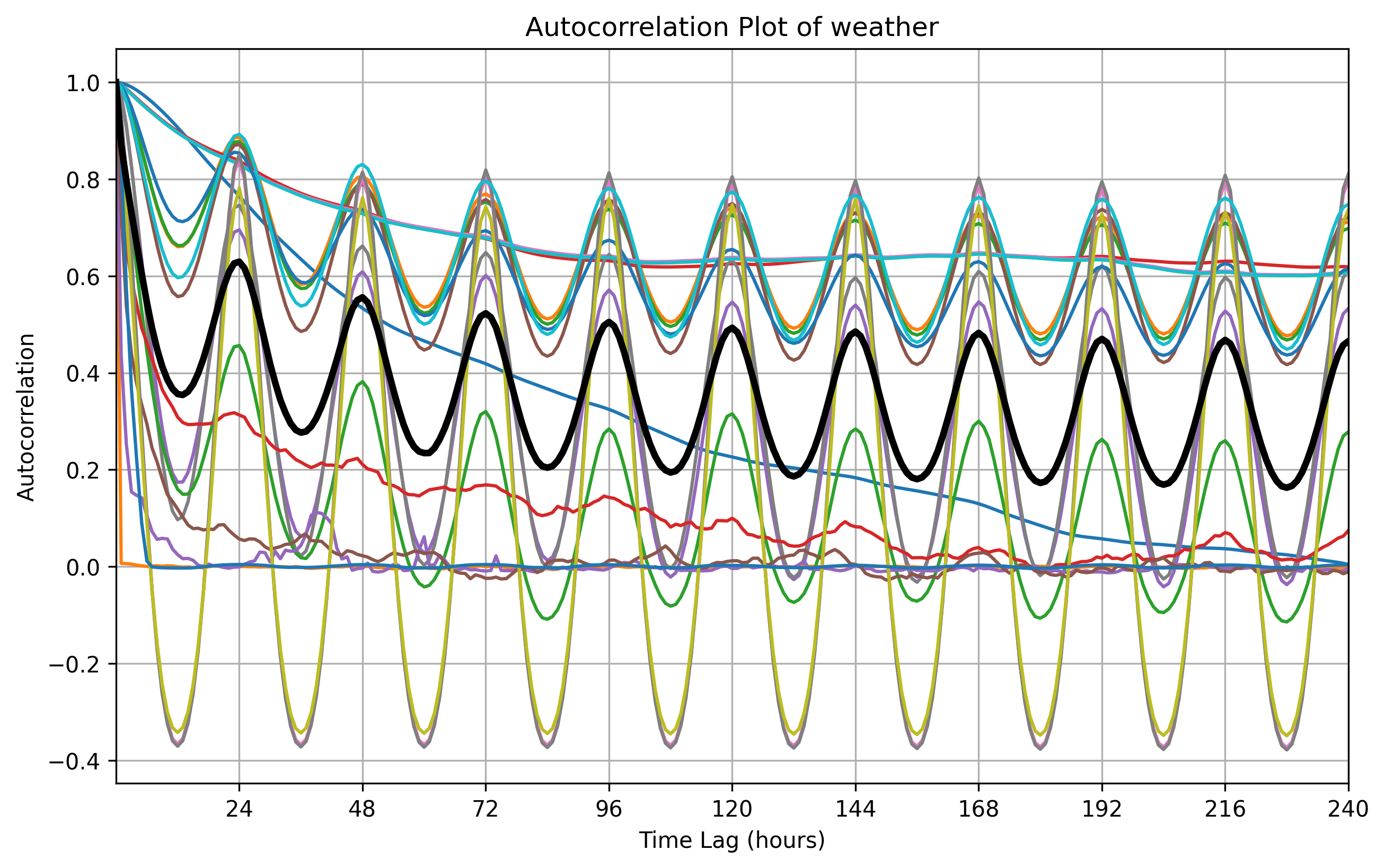}
            \caption{Time lag correlation of all channels}
            \label{fig:autocorr_plot_weather}
        \end{minipage}
    \end{minipage}
\end{figure}

We split the weather data in several plots for better readability. We can also observe from the plots that there seems to be quite high correlation between some of the channels. This makes perfectly sense as weather channels are heavily correlated such as water content and dew point. From the auto correlation plot we can observe heavy correlation and periodicity between most of the channels, where some of the channels converge to a correlation of 0.

\begin{figure}[H]
    \centering
    \includegraphics[width=1\linewidth]{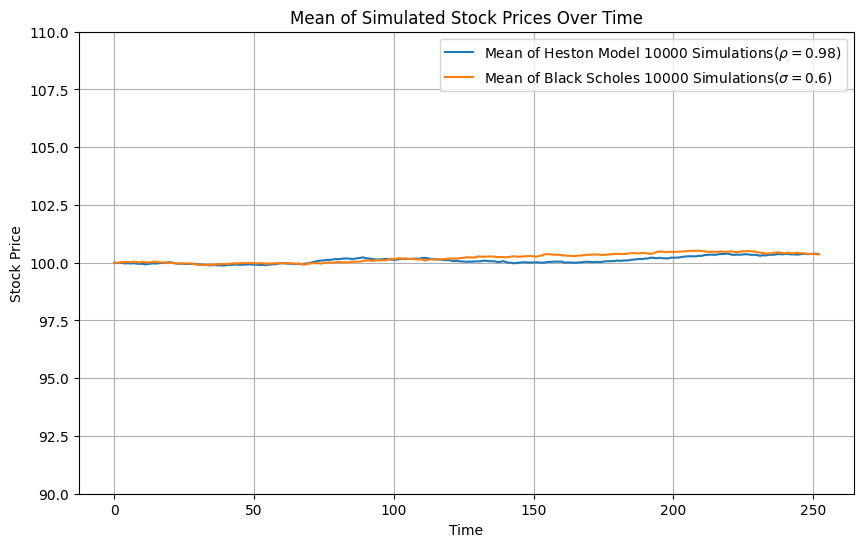}
    \caption{Infinite random walks $\implies$ naive repeat (\href{https://github.com/thorhojhus/ssl_fts/blob/main/notebooks/compare_random_walk.ipynb}{compare\_random\_walk.ipynb}) [$\color{red} \rightarrow$ \ref{goback2}]}
    \label{fig:compare_random_walk}
\end{figure}

\vspace{1cm}
\subsection*{Fast Fourier Transform algorithm \label{appendix:FFT}[$\color{red} \rightarrow$ \ref{goback6}]}
Implementation based on \href{https://www.youtube.com/watch?v=h7apO7q16V0}{The Fast Fourier Transform (FFT): Most Ingenious Algorithm Ever?}.
\begin{enumerate}
    \item \textbf{Bit-Reversal Permutation:}
    Start by reordering the input array using a bit-reversal permutation where the binary representation of the indices are reversed resulting in a new index. This step rearranges the original data points so that they align with their final positions early in the process, facilitating the efficient combination in later steps.
    
    \begin{equation}x'_{\text{bit-reversed}(k)} = x_k, \quad \text{for all } k \in [0, N-1]\end{equation}

    \item \textbf{Divide and Conquer:}
    Recursively divide the reordered array into smaller arrays. the lenght of the current data input is larger than $N$, this step splits the data into two halves, each of size $N/2$, and applies the FFT recursively.
    
    \item \textbf{Butterfly Combinations:}
    For each level of recursion, combine the DFTs of smaller arrays using the butterfly operations, which are simple add-and-subtract operations with twiddle factor multiplications which are constant trigonometric coefficients corresponding to the $N$ roots of unity for the given input:
    
    \begin{equation}
    \begin{aligned}
    X_k &= E_k + O_k \cdot e^{-2\pi i k / N}, \quad \text{for } k = 0, \ldots, \frac{N}{2}-1 \\
    X_{k+\frac{N}{2}} &= E_k - O_k \cdot e^{-2\pi i k / N}, \quad \text{for } k = 0, \ldots, \frac{N}{2}-1
    \end{aligned}
    \end{equation}
    where $E_k$ and $O_k$ are the DFTs of the even and odd indexed parts of the array, respectively.

    \item \textbf{Recursive Computation:}
    Continue the recursive process until the full DFT is computed. When the base case ($N=1$) is reached, the DFT of a single element is the element itself.
    
    \item \textbf{Combine Results:}
    Combine all small DFT results back into the final DFT of the original array using the relationships established in the butterfly operations.
\end{enumerate}

In pure Python the FFT and the inverse FFT it can be expressed as follows:

\begin{minipage}{.45\textwidth}
\begin{lstlisting}[language=Python, style=mystyle]
def FFT(x):
    # Must to be a power of 2
    n = len(x)

    # Base case
    if n == 1:
        return x
    # Twiddle factor
    w = exp(-2j * pi/n) 
    # Even indexed coefficients
    xe = x[0::2]
    # Odd indexed coefficients
    xo = x[1::2] 

    # Recursive calls
    ye = FFT(xe)
    yo = FFT(xo)

    # Initialize the result
    y = [0] * n
    for k in range(n//2):
        # Compute twiddle factor
        wk = w ** k
        # Butterfly combinations
        y[k] = ye[k]+yo[k]*wk
        y[k+n//2] = ye[k]-yo[k]*wk
    return y
\end{lstlisting}
\end{minipage}
\begin{minipage}{.45\textwidth}
\begin{lstlisting}[language=Python, style=mystyle]
def IFFT(x):
    # Must to be a power of 2
    n = len(x)

    # Base case
    if n == 1:
        return x
    # Twiddle factor
    w = exp(2j * pi/n)
    # Even indexed coefficients
    xe = x[0::2]
    # Odd indexed coefficients
    xo = x[1::2]

    # Recursive calls
    ye = IFFT(xe)
    yo = IFFT(xo)

    # Initialize the result
    y = [0] * n
    for k in range(n // 2):
        # Compute twiddle factor
        wk = w ** k
        # Butterfly combinations
        y[k] = ye[k]+yo[k]*wk
        y[k+n//2] = ye[k]-yo[k]*wk
    return y
\end{lstlisting}
\end{minipage}

Note that this example doesn't normalize the data from the inverse FFT nor does it implement the bit-reversal permutation. The inverse FFT can be expressed similarly to FFT just with the opposite sign in the twiddle factor and afterward should be scaled by $\frac{x}{n} \text{ for } x \in \mathbf{x}$.
\vspace{1cm}
\subsection*{Autoregressive Integrated Moving Average Models (ARIMA) \label{sec: ARIMA} [$\color{red} \rightarrow$ \ref{goback0}] [$\color{red} \rightarrow$ \ref{goback3}]}

In this chapter, an overview in short detail the individual components of the complete ARIMA model, as well as estimation of the parameters using a maximum likelihood framework, will be described.

\subsubsection*{Autoregressive Model (AR)}
The autoregressive model is a type of linear model in which the current value is expressed as a finite, linear combination of the previous values of the model $\tilde{z}$ and a error term of white noise $\epsilon$. The term $\tilde{z}_t$ is defined as the deviation from the mean of the data $\tilde{z}_t=z_t-\mu$. The autoregressive model with weights $\phi$ can be expressed as:
\begin{equation}
\tilde{z}_t = \sum_{i=1}^{p} \phi_{i}\tilde{z}_{t-i} + \epsilon_t
\label{AR}
\end{equation}
Where the order $p$ is the number of lagged values of the time series included in the model and indicates how many previous time steps the current value of the series is regressed upon.

In contrast to a linear regression model can be stated as:
$$
\tilde{z} = \sum_{i=1}^{p} \phi_i \tilde{x}_i + \epsilon
$$
Where the dependent variable $\tilde{z}$ is regressed on the independent variables $\bm{x}$ with the error $\epsilon$. 

The primary difference between the autoregressive model and the linear regression model, is that the variable $z$ is regressed on previous values of itself -- this is why it is called an \emph{auto}regressive model. 

Using the backshift operator defined as $\bm{B}^kx_t = x_{t-k}$, we can express the autoregressive operator of order $p$ in a more concise format as:
$$
\phi(\bm{B}) = 1-\sum_{i=1}^{p} \phi_i B^i
$$

This allows us to express the autoregressive model as:
\begin{equation}
\phi(\bm{B})\tilde{z}_t = \epsilon_t
\label{backshift_op}
\end{equation}

Which is equivalent to:

$$
\tilde{z}_t = \phi^{-1}(\bm{B})\epsilon_t
$$

This gives us the model parameters $\mu, \phi_1, \phi_2, ..., \phi_p, \sigma^2_{\epsilon}$ which is a total of $p+2$ parameters. 

The model is stationary, meaning constant mean $\mu$ and variance $\sigma^2_{\epsilon}$, if the roots to the characteristic polynomial of the autoregressive equation $\phi(\bm{B})=1-\sum_{i}^p \phi_i B^i=0$, are greater than 1 \cite{arma_book}. For the AR(1) process, this imposes the condition of $\phi_1$ < 1, for example.

\subsubsection*{Moving Average Model (MA)}

Another important model for time series deviation representation is the moving average model which is defined as:
\begin{align*}
    \tilde z_t &= \epsilon_t - \theta_1\epsilon_{t-1} - \cdots \theta_{q}\epsilon_{t-q} \\
    &= \epsilon_t - \sum_{i=1}^q \theta_{i}\epsilon_{t-i}
\label{MA}
\end{align*}

In this model, $\tilde{z}_t$ represents the deviation of the time series from its mean value at time $t$. The term $\epsilon_t$ denotes the white noise error at time $t$, while $\theta_1, \theta_2, \ldots, \theta_q$ are parameters that quantify the influence of the past $q$ error terms on the current value. The moving average model employs a linear combination of these $q$ past error terms to describe the current deviation. This approach captures dependencies between consecutive errors, thereby enhancing forecasting accuracy.

Using the backshift operator $\bm{B}$, it can concisely be written as:
\begin{equation*}
    \tilde{z}_t = \theta(\bm{B})\epsilon_t
\end{equation*}

where the notation for the backshift operator is the same as in equation (\ref{backshift_op}) with $\phi$ replaced by $\theta$. In the moving average model, the parameters $\theta_1,\cdots, \theta_q$ are all unknown, as well as the mean $\mu$ and the variance $\sigma^2_t$, and have to be estimated.

\subsubsection*{Mixed Autoregressive-Moving Average Models (ARMA)}
The ARMA model consists of an autoregressive componenent a moving average componenent mixed together to form the model form of
\begin{align}
    \tilde{z}_t &= \phi_1\tilde{z}_{t-1} + \cdots + \phi_{p}\tilde{z}_{t-p} + \epsilon_t - \theta_1\epsilon_{t-1} - \cdots - \theta_q\epsilon_{t-q} \\
    &= \epsilon_t + \sum_{i=1}^p \phi_i\tilde{z}_{t-i} - \sum_{j=1}^{q}\theta_j\epsilon_{t-q}
\end{align}

As such, it has a total of $p + q + 2$ parameters that needs to be estimated from the data, which are: $\mu, \epsilon_t^2, \phi_i, \theta_j$.\cite{arma_book}

\subsection*{ARMA parameter estimation}

\textbf{AR(p)}

Now consider the general AR model:

\begin{equation}
z_t = \mu + \phi_1z_{t-1} + \cdots \phi_pz_{t-p} + \epsilon_t
\label{ARP}
\end{equation}

If we consider the same conditional maximum likelihood framework as before, we now assert that the $p$ first observations are deterministic. The likelihood function, $\Theta = (\mu, \sigma^2, \phi_1, \cdots, \phi_p)$ is now:

\begin{equation}
    \mathcal{L}(\Theta) = \prod_{t=p+1}^T\frac{1}{\sqrt{2\pi\sigma^2}}\exp\left(\frac{-(z_t - \mu - \phi_1z_{t-1} - \phi_2z_{t-2}- \cdots - \phi_pz_{t-p})^2}{2\sigma^2}\right)
\end{equation}

And the equivalent log likelihood functions, simplified directly and omitting intermediary steps:

\begin{equation}
    \log\mathcal{L}(\Theta) = -\frac{T-p}{2}\cdot\log(2\pi) - \frac{T-p}{2}\cdot\log(\sigma^2) - \sum_{t=p+1}^T\frac{(z_t - \mu - \phi_1z_{t-1} - \phi_2z_{t-2} - \cdots - \phi_pz_{t-p})^2}{2\sigma^2}
\end{equation}

Likewise with AR(1), the normal equations for the OLS regression can be set up, and the variance is similar to AR(1) simply:

\begin{equation}
    \hat{\sigma}^2 = \frac{1}{T-p}\sum_{t=p+1}^T (z_t - \hat{\mu} - \hat{\phi_1}z_{t-1} - \hat{\phi_2}z_{t-2} - \cdots - \hat{\phi_p}z_{t-p})^2
\end{equation}

\textbf{MA(q)}

Now consider the general MA(q) process:

\begin{equation}
    z_t = \mu + \epsilon_t + \theta_1\epsilon_{t-1} + \theta_2\epsilon_{t-2} + \cdots + \theta_q\epsilon_{t-q}
\end{equation}

Using a similar approach, assume the first $q$ errors are 0; $\epsilon_0=\epsilon_{-1}=\cdots=\epsilon_{-q+1}=0$. This again allows the iteration of:

\begin{equation}
    \epsilon_t = z_t - \mu - \theta_1\epsilon_{t-1} - \theta_2\epsilon_{t-2} - \cdots - \theta_q\epsilon_{t-q}
\end{equation}

The conditional log likelihood function, much like before can be seen in equation (\ref{MAq}), which also requires numerical optimization:

\begin{equation}
    \log\mathcal{L}(\Theta) = - \frac{T}{2}\log(2\pi) - \frac{T}{2}\log(\sigma^2) - \sum_{t=1}^T\frac{\epsilon_t^2}{2\sigma^2}
\label{MAq}
\end{equation}

\subsubsection*{Autoregressive Integrated Moving Average (ARIMA)}

Developed in the 1970s, ARIMA models combine autoregressive, integration, and moving average components to capture trends, seasonality, and other patterns in time series data. ARIMA and its variants, such as ARIMAX (ARIMA with explanatory variables), have been widely used for univariate forecasting tasks and are particularly useful for modeling non-stationary time series data by applying differencing to induce stationarity.

\textbf{Motivation for Integration term in AR/MA models}

One of the assumptions for AR and MA models to work is that the model is stationary. Often, time series will not be stationary, however, differencing (i.e, \textit{integrating}) the time series can be a tool to potentially introduce stationarity.

Differencing a time series means taking the difference between the time series of $t$ and that of lag 1, i.e., $t-1$. Introducing the \textit{backwards difference} operator $\nabla$, we can now define it with $k>0$ as:

\begin{align}
    \nabla^1z_t &= z_t - z_{t-1} \\
    \nabla^kz_t &= \nabla^{k-1}z_t - \nabla^{k-1}z_{t-1}
\end{align}

Backwards differencing a time series once will then yield the following:

\begin{equation*}
    \nabla^1z_t = z_t - z_{t-1}
\end{equation*}

adding yet another difference term would yield:

\begin{align*}
    \nabla^2z_t &= \nabla^1z_t - \nabla^1z_{t-1} \\ 
                &= (z_t - z_{t-1}) - (z_{t-1} - z_{t-2}) \\
                &= z_t - 2z_{t-1} + z_{t-2} \\
\end{align*}

This can also be written using the backshift operator giving us the differencing operator $\nabla^d = (1-\bm{B})^d$:
\begin{equation*}
 \nabla^1 z_t = (1 - \bm{B})z_t = z_t - z_{t-1}
\end{equation*}

and for another term:

\begin{equation*}
    \nabla^2 z_t = (1 - \bm{B})^2 z_t = z_t - 2z_{t-1} + z_{t-2}
\end{equation*}

By continuing to add difference terms, we can further transform the series until stationarity is achieved. Each level of differencing removes a trend or seasonality in the data, simplifying the underlying structure and making it suitable for AR and MA modeling. For instance, applying a second level of differencing as shown transforms the series by removing linear trends.

In general, if a time series requires $d$ levels of differencing to become stationary, it is said to be integrated of order $d$, denoted as $I(d)$, indicating the number of differencing steps required for stationarity. 

Recall that if the roots to the characteristic polynomial of the autoregressive equation $\phi(\bm{B})=1-\sum_{i}^p \phi_i B^i=0$, are greater than 1, the time series was considered stationary. In case one, or more, of the roots are not, the time-series is considered non-stationary. In such cases, differentiating can be a tool to introduce it.

\textbf{ARIMA model}

An ARIMA model of order $(p, d, q)$ can be written as:
$$
\nabla^d z_t = \phi_1 \nabla^d z_{t-1} + \cdots + \phi_p \nabla^d z_{t-p} + \epsilon_t - \theta_1 \epsilon_{t-1} - \cdots - \theta_q \epsilon_{t-q}
$$
Where $p$ is the order of the autoregressive part, $d$ is the degree of differencing, $q$ is the order of the moving average part and $\nabla^d$ is the differencing operator applied $d$ times.

Using the backshift operator $\bm{B}$ and the differencing operator $\nabla^d$, the ARIMA model can be compactly represented as:

$$
\phi(\bm{B})\nabla^d z_t = \theta(\bm{B})\epsilon_t
$$

where $\phi(\bm{B})$ is the autoregressive part, $(1 - \bm{B})^d$ is the integration part, and $\theta(\bm{B})\epsilon_t$ is the moving average part.

An ARIMA(0,0,0) is equivalent to $z_t = Z_t$ and can be considered white noise, ARIMA(0,1,0) is equivalent to $z_t=Z_t-Z_{t-1}$ and is a random walk

\begin{figure}[H]
    \centering
    \includegraphics[width=\linewidth]{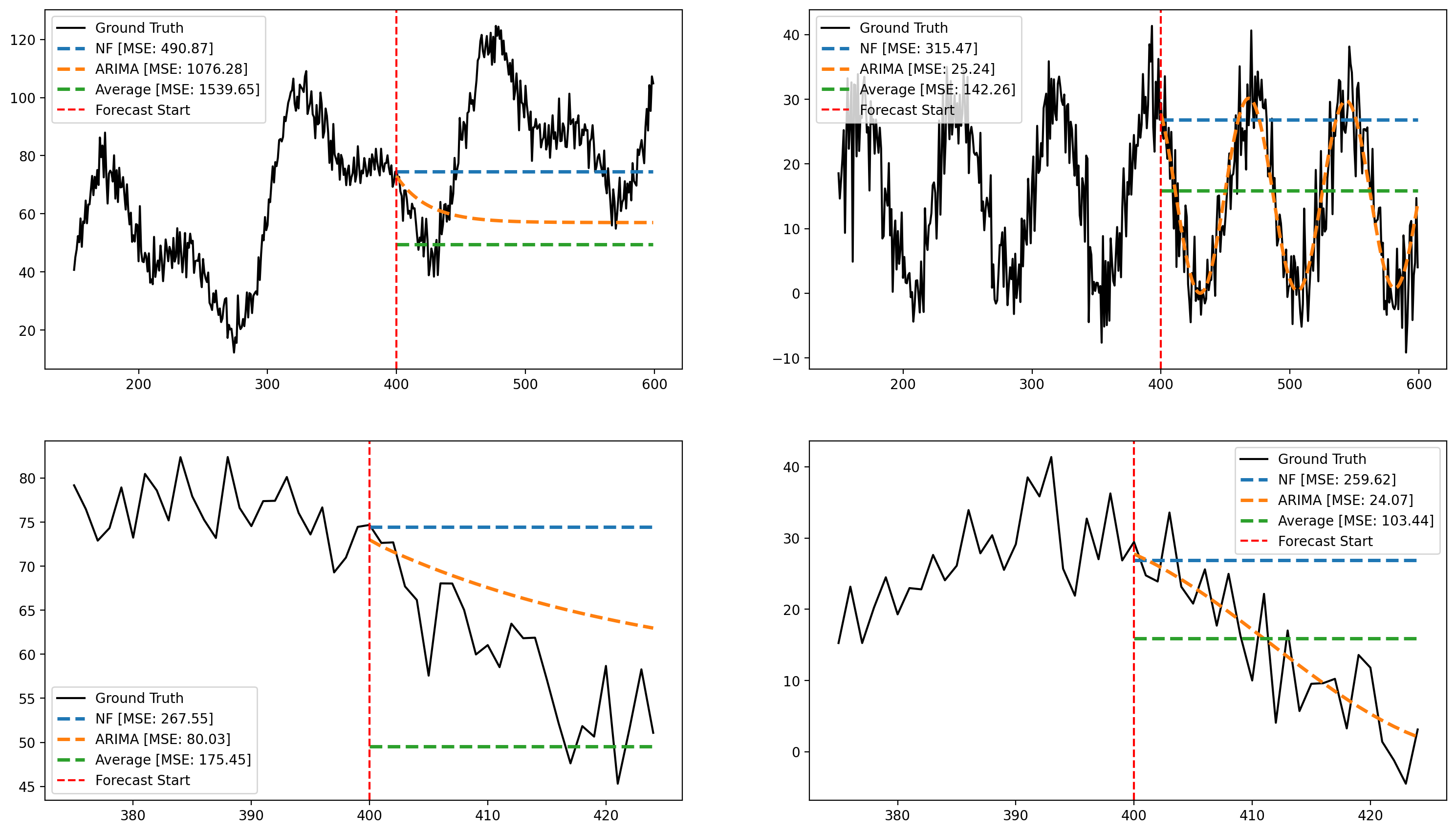}
    \caption{Examples of the baselines and ARIMA on two types of synthetic data. \\ 
    \indent \textbf{Left: } Synthetic data consists of 3 sinusoidal waves with normally distributed noise $\mathcal{N}(\mu=15, \sigma^2=25)$ and a drift of $d=0.15$. Here the Naïve forecast performs better on longer horizons, but ARIMA does best on short horizons. \\ 
    \indent \textbf{Right:} A simpler sinusoidal wave with noise $\mathcal{N}(\mu=15, \sigma^2=25)$ and no drift. Here ARIMA almost perfectly captures the underlying signal, achieving superior performance, whereas the Naïve forecast does much worse.}
    \label{fig:baselines}
\end{figure}

\textbf{Estimating parameters}

Estimating the parameters of any of the previous models can be done using different methods including \textit{Least Squares Estimation} (LSE), \textit{Maximum Likelihood Estimation} (MLE), \textit{Method of Moments} (MoM) etc,. We mention these, but we will only inspect the method \textit{Maximum Likelihood Estimation} (MLE), more specifically the \textit{conditional} version, and how it can be used to estimate the parameters $\hat{\mu}, \hat{\phi_1}, \hat{\theta_1}$ etc.

\subsubsection*{Conditional Maximum Likelihood Estimation}

Maximum likelihood estimation is a method to estimate parameters in a probabilistic framework given observed data. The idéa is a two-step process: first is to define a likelihood function $\mathcal{L}$ based on the assumed or determined distribution of the observed data and the second step is maximize it with regards to the parameters and determine the values. This is usually done by differencing with regards to the parameters and setting it to 0.

\begin{equation*}
    \frac{\partial\mathcal{L}}{\partial x} = 0,\quad x \in \{\mu, \sigma, \theta_i, \phi_j\}
\end{equation*}

For most time series data, the error is assumed to be i.i.d., following a gaussian distribution around a zero mean $\epsilon_t \sim \mathcal{N}(0, \sigma^2)$.

The likelihood function can turn out to be computationally tough to do, due to non-linearity. However, if one models it conditioned on a full stochastic process. If one instead assumes some deterministic behaviour, it simplifies the calculations without sacrificing too much precision in results, under assumptions of large sample size and stationarity. This is the approach we explore here. \cite{Hamilton1994-cw}

\textbf{AR(1)}

Recall the AR(1) model\footnote{Recall we assume \textit{stationarity}, as such the $\mu$ is not time dependent.}:

\begin{equation}
z_t = \mu + \phi_1z_{t-1} + \epsilon_t
\label{AR1}
\end{equation}

The process for a more exact maximum likelihood function can be found for an AR(1) process, however, it will require numerical/iterative algorithms to find a result.

Instead we here condition on the fact that the first observation, $z_1$ is \textit{deterministic}. This simplification will make it possible to derive a cleaner expression. The likelihood function $\mathcal{L}(\Theta) = (\mu, \sigma, \phi_1)$ will be conditioned on the first observation:

\begin{equation}
    \mathcal{L}(\Theta) = f_{z_T,z_{T-1},\cdots,z_2|z_1}(z_T, z_{T-1},\cdots,z_2|z_1;\Theta) = \prod_{t=2}^{T}f_{z_t|z_{t-1}}(z_t|z_{t-1},\Theta)
\end{equation}

We assumed the noise follows a Gaussian distribution, it is fair to assume the same for the time series observations.\footnote{The maximum likelihood for non-gaussian time series requires some other approaches and methods, but can in some cases be modelled as if gaussian. This is known as \textit{quasi-maximum likelihood estimation}. This will not be explored in this report.} With the assumption that the first observation is deterministic, from equation (\ref{AR1}), the expected value and variance is trivially seen to be $(c + \phi_1\cdot z_{t-1}, \sigma^2)$, and we can now model the conditional distribution as $P(z_t|z_{t-1}) \sim \mathcal{N}((\mu+\phi_1z_t),\sigma^2)$:

\begin{equation}
   f_{z_t|z_{t-1}}(z_t|z_{t-1},\Theta) = \frac{1}{\sqrt{2\pi\sigma^2}}\exp{\left(\frac{-(z_t - \mu - \phi_1z_{t-1})^2}{2\sigma^2}\right)}
\end{equation}

Our likelihood functions is thus:

\begin{equation}
    \mathcal{L}(\Theta) = \prod_{t=2}^{T}\frac{1}{\sqrt{2\pi\sigma^2}}\exp{\left(\frac{-(z_t - \mu - \phi_1z_{t-1})^2}{2\sigma^2}\right)}
\end{equation}

It is customary to take the logarithm of ones likelihood function. As the logarithm functions is monotone and increasing, the values that would maximize a likelihood function would also maximize a log likelihood function. When using computer software, it is also provides more correct results due to floating point arithmetic, as a product of many value is instead a sum of the same values.

\begin{align}
    \log\mathcal{L}(\Theta) &= \sum_{t=2}^{T}\log\left(\frac{1}{\sqrt{2\pi\sigma^2}}\exp{\left(\frac{-(z_t - \mu - \phi_1z_{t-1})^2}{2\sigma^2}\right)}\right) \\
    &= \sum_{t=2}^{T}\log\left(\frac{1}{\sqrt{2\pi\sigma^2}}\right) + \log\left(\exp{\left(\frac{-(z_t - \mu - \phi_1z_{t-1})^2}{2\sigma^2}\right)}\right) \\
    &= \sum_{t=2}^{T}\log\left(\frac{1}{\sqrt{2\pi\sigma^2}}\right) - \frac{(z_t - \mu - \phi_1z_{t-1})^2}{2\sigma^2} \\
    &= \sum_{t=2}^{T} -\frac{\log(2\pi)}{2} - \frac{\log(\sigma^2)}{2} - \frac{(z_t - \mu - \phi_1z_{t-1})^2}{2\sigma^2} \\
    &= -\frac{(T-1)}{2}\cdot\log(2\pi) - \frac{(T-1)}{2}\cdot\log(\sigma^2) - \sum_{t=2}^{T}\frac{(z_t - \mu - \phi_1z_{t-1})^2}{2\sigma^2}
\label{LogLikelihoodAr1}
\end{align}

From equation (\ref{LogLikelihoodAr1}) a sharp eye can see that a maximization of the log likelihood with regards to $\phi_1$ and $\mu$ would simply be the minimization of $\sum_{t=2}^T(z_t - \mu - \phi_1z_{t-1})^2$, which can be achieved using Ordinary Least Squares (OLS) Regression. After writing out the normal equations, the computation simplifies to the following. \cite{Hamilton1994-cw}

\begin{equation*}
    \left[
    \begin{array}{c}
         \hat{\mu} \\
         \widehat{\phi_1}
    \end{array}
    \right] = \left[
    \begin{array}{cc}
        T-1 & \sum z_{t-1} \\
        \sum z_{t-1} & \sum z^2_{t-1}
    \end{array}
    \right]^{-1}\left[
    \begin{array}{c}
         \sum z_t \\
         \sum z_{t-1}z_t
    \end{array}
    \right]
\end{equation*}

The conditional maximum likelihood estimate of the variance, $\hat{\sigma}^2$ is done by differentiating equation (\ref{LogLikelihoodAr1}) w.r.t, $\sigma^2$ and setting it equal to 0.

\begin{align}
    \frac{\partial\log\mathcal{L}(\Theta}{\partial\sigma^2} = -\frac{T-1}{2\sigma^2} + \sum_{t=2}^T\left(\frac{(z_t - \mu - \phi_1z_{t-1})^2}{2\sigma^4}\right) &= 0 \\
    \hat{\sigma}^2 &= \frac{1}{T-1}\sum_{t=2}^T(z_t - \hat{\mu} - \hat{\phi_1}z_{t-1})^2
\end{align}

\textbf{MA(1)}

Unfortunately where AR(1)'s conditional maximum likelihood equations could be solved, MA does not impose such elegance - the exact likelihood function is also not trivial. Nonetheless, we'll derive the conditional likelihood function, and then refer to Numerical Optimization algorithms to solve them.

Recall the MA(1) model.

\begin{equation}
    z_t = \mu + \epsilon_t + \theta\epsilon_{t-1}
\end{equation}

Assume that we know for certain $\epsilon_{t-1}$ and that $\epsilon_0 = 0$\footnote{This imposes a condition on $\theta$ and it is that $|\theta| < 1$. The reason is, if $\epsilon_0$ is \textit{not} 0, but we model as if it is, the error will consequently accumulate over time, and the result must be discarded, and the reciprocal value of $\theta$ should be investigated as a starting point instead.\cite{Hamilton1994-cw}}, then the probability of the first observation is simply $(z_1|\epsilon_0) \sim \mathcal{N}(\mu, \sigma^2)$

Consequently, one now knows $\epsilon_1 = z_1 - \mu$ allowing one to form the probability density function for $z_2$ given $z_1$:

\begin{align*}
    f_{z_2|z_1,\epsilon_0=0}(z_2|z_1,\epsilon_0=0;\Theta) &= \frac{1}{\sqrt{2\pi\sigma^2}}\exp\left({\frac{-(z_2-\mu-\theta_1\epsilon_1)^2}{2\sigma^2}}\right) \\
    &= \frac{1}{\sqrt{2\pi\sigma^2}}\exp\left({\frac{-\epsilon_2^2}{2\sigma^2}}\right)
\end{align*}

Similar can be written for the proceeding $z_t$'s! This allows us to form the conditional log-likelihood function, omitting some of the intermediary steps:

\begin{align}
    \log\mathcal{L}(\Theta) &= \log f_{z_T, z_{t-1},\cdots,z_1|\epsilon_0=0}(z_T,z_{t-1},\cdots,z_{1}|\epsilon_0=0;\Theta)
    &= -\frac{T}{2}\log(2\pi)-\frac{T}{2}\log(\sigma^2)-\sum_{t=1}^T\frac{\epsilon_t^2}{2\sigma^2}
\end{align}

The numerical values of $\Theta$ allows one to compute the sequence of $\epsilon$ s from the data, for which the log likelihood is a function of the sum of squares of these. It is therefore then a complicated non-linear function, which in practice requires numerical optimizations to calculate.

\textbf{ARMA(p, q)}

Assume now the general ARMA model:
\begin{equation}
    z_t = \mu + \phi_1z_{t-1} + \phi_2z_{t-2} + \cdots + \phi_pz_{t-p} + \epsilon_t - \theta_1\epsilon_{t-1} - \theta_2\epsilon_{t-2} - \cdots - \theta_q\epsilon_{t-q}
\end{equation}

A common approximation of a likelihood function will require conditions on both $z$'s and $\epsilon$'s, just like the individual AR and MA processes.

As such, assume the the values of $(z_0, z_{-1}, \cdots, z_{-p+1})$ as given, likewise with $(\epsilon_0, \epsilon_{-1},\cdots, \epsilon{-q+1})$. This will again allow the sequence of noise following $t=0$ to be calculated:

\begin{equation}
    \epsilon_t = z_t - \mu - \phi_1z_{t-1} - \phi_2z_{t-2} - \cdots - \phi_pz_{t-p} - \theta_1\epsilon_{t-1} - \theta_2\epsilon_{t-2} - \cdots - \theta_q\epsilon_{t-q}
\label{ARMAiteration}
\end{equation}

And with that the conditional log likelihood function is as seen before, albeit a bit larger expression:

\begin{equation}
    \log\mathcal{L}(\Theta) = -\frac{T}{2}\log(2\pi) - \frac{T}{2}\log(\sigma^2)-\sum_{t=1}^T\frac{\epsilon_t^2}{2\sigma^2}
\end{equation}

Like the the MA process, one must assume the values of the given noise and $z$'s. Commonly one assumes the $\epsilon$ takes on the value 0, and $z$'s take the observed values, and iteration in equation (\ref{ARMAiteration}) starts from $t=p+1$, and the noise:

\begin{equation*}
    \epsilon_p = \epsilon_{p-1} = \cdots = \epsilon_{p-q+1} = 0
\end{equation*}

Which now allows the conditional likelihood to be of the form:

\begin{equation}
    \log\mathcal{L}(\Theta) = -\frac{T-p}{2}\log(2\pi) - \frac{T-p}{2}\log(\sigma^2)-\sum_{t=p+1}^T\frac{\epsilon_t^2}{2\sigma^2}
\end{equation}

\textbf{MLE of ARIMA}

When dealing with the parameters of ARIMA, one now deals with a differentiated time series, this has a direct impact on the Maximum Likelihood Estimation, and depends on the amount of differentiating. Nonetheless, the general approach is the same as in ARMA.